\documentclass[pdflatex, sn-mathphys-ay]{sn-jnl}

\usepackage{graphicx}%
\usepackage{multirow}%
\usepackage{amsmath,amssymb,amsfonts}%
\usepackage{amsthm}%
\usepackage{mathrsfs}%
\usepackage[title]{appendix}%
\usepackage{xcolor}%
\usepackage{textcomp}%
\usepackage{manyfoot}%
\usepackage{booktabs}%
\usepackage{algorithm}%
\usepackage{algorithmicx}%
\usepackage{algpseudocode}%
\usepackage{listings}%

\usepackage{colortbl}
\usepackage{booktabs}
\usepackage{array}
\usepackage{subfig}
\usepackage{booktabs}
\usepackage{siunitx}

\definecolor{myorg}{rgb}{.976,.820,.710}
\theoremstyle{thmstyleone}%

\theoremstyle{thmstyletwo}%

\theoremstyle{thmstylethree}%

\begin{document}
	
\title[Article Title]{Flexible Camera Calibration using a Collimator System}

\author[1]{\fnm{Shunkun} \sur{Liang}}\email{liangshunkun@nudt.edu.cn}
\author*[1]{\fnm{Banglei} \sur{Guan}}\email{guanbanglei12@nudt.edu.cn}
\author[1]{\fnm{Zhenbao} \sur{Yu}}\email{zhenbaoyu@whu.edu.cn}
\author[1]{\fnm{Dongcai} \sur{Tan}}\email{tandongcai23@nudt.edu.cn}
\author[1]{\fnm{Pengju} \sur{Sun}}\email{sunpengju23@nudt.edu.cn}
\author[1]{\fnm{Zibin} \sur{Liu}}\email{liuzibin19@nudt.edu.cn}
\author[1]{\fnm{Qifeng} \sur{Yu}}\email{yuqifeng@nudt.edu.cn}
\author[1]{\fnm{Yang} \sur{Shang}}\email{shangyang1977@nudt.edu.cn}

\affil[1]{\orgdiv{College of Aerospace Science and Engineering}, \orgname{National University of Defense Technology}, \city{Changsha}, \postcode{410073}, \country{China}}
	
\abstract{Camera calibration is a crucial step in photogrammetry and 3D vision applications. This paper introduces a novel camera calibration method using a designed collimator system. Our collimator system provides a reliable and controllable calibration environment for the camera. Exploiting the unique optical geometry property of our collimator system, we introduce an angle invariance constraint and further prove that the relative motion between the calibration target and camera conforms to a spherical motion model. This constraint reduces the original 6DOF relative motion between target and camera to a 3DOF pure rotation motion. Using spherical motion constraint, a closed-form linear solver for multiple images and a minimal solver for two images are proposed for camera calibration. Furthermore, we propose a single collimator image calibration algorithm based on the angle invariance constraint. This algorithm eliminates the requirement for camera motion, providing a novel solution for flexible and fast calibration. The performance of our method is evaluated in both synthetic and real-world experiments, which verify the feasibility of calibration using the collimator system and demonstrate that our method is superior to existing baseline methods. Demo code is available at \href{https://github.com/LiangSK98/CollimatorCalibration}{https://github.com/LiangSK98/CollimatorCalibration}}
\maketitle
	
\section{Introduction}\label{sec:intro}
Geometric camera calibration determines the mapping between 3D rays in space and 2D image pixels, upgrading cameras from recording devices to accurate measurement instruments. Various vision applications, such as simultaneous localization and mapping(SLAM) \citep{Campos2021}, structure from motion(SfM) \citep{Colmap2016}, and pose estimation \citep{GuanTCYB2021, Guan2023}, rely on camera parameters as inputs. As a fundamental step in 3D vision applications, the flexibility and accuracy of camera calibration have always been core topics. The close-range calibration methods using a planar target represent the prevailing techniques in 3D vision applications \citep{Brown1971,Sturm1999,Zhang2000,Lochman2021,Thomas2020}. However, in some special outdoor scenarios, such as long-range depth estimation \citep{Zhang2020} and wind turbine blades monitoring \citep{Guan2022}, {where cameras operate at long working distance and have a large field of view}, traditional target-based calibration methods \citep{Tsai1987,Zhang2000,Bouguet2004} face several challenges. These include the requirement for large-sized targets that match the field of view, the logistical difficulties of moving or arranging targets over a wide area, and the sensitivity to varying and unstable lighting conditions. 

In these application scenarios, providing effective calibration data and a stable calibration environment is crucial. {Self-calibration methods \citep{Hartley1997, Pollefeys1999, Herrera2016} are often ineffective in these scenarios due to their sensitivity to sparse features and lighting variations, which can lead to unstable and inaccurate estimations.} Although constructing a 3D calibration field offers precise calibration data \citep{Wang2015,Xiao2010,Shang2013}, its operational complexity and limited adaptability present significant challenges. Collimators have a long-standing history in providing calibration targets for cameras \citep{Hothmer1958,Hallert1963,Clarke1998}. Their design aims to produce parallel rays. {The collimator contains an independent light source}, which can minimize the impact of environmental factors such as ambient light changes and reflections. However, traditional collimator-based methods typically rely on measuring the direction of collimated rays using high-precision goniometers or theodolites. This requirement makes the calibration process challenging, and measurement errors directly impact the accuracy \citep{Yuan2019,Yuan2021,Wu2007,Hieronymus2012}.

\begin{figure}[tbp]
	\centering
	\includegraphics[width=0.7\linewidth]{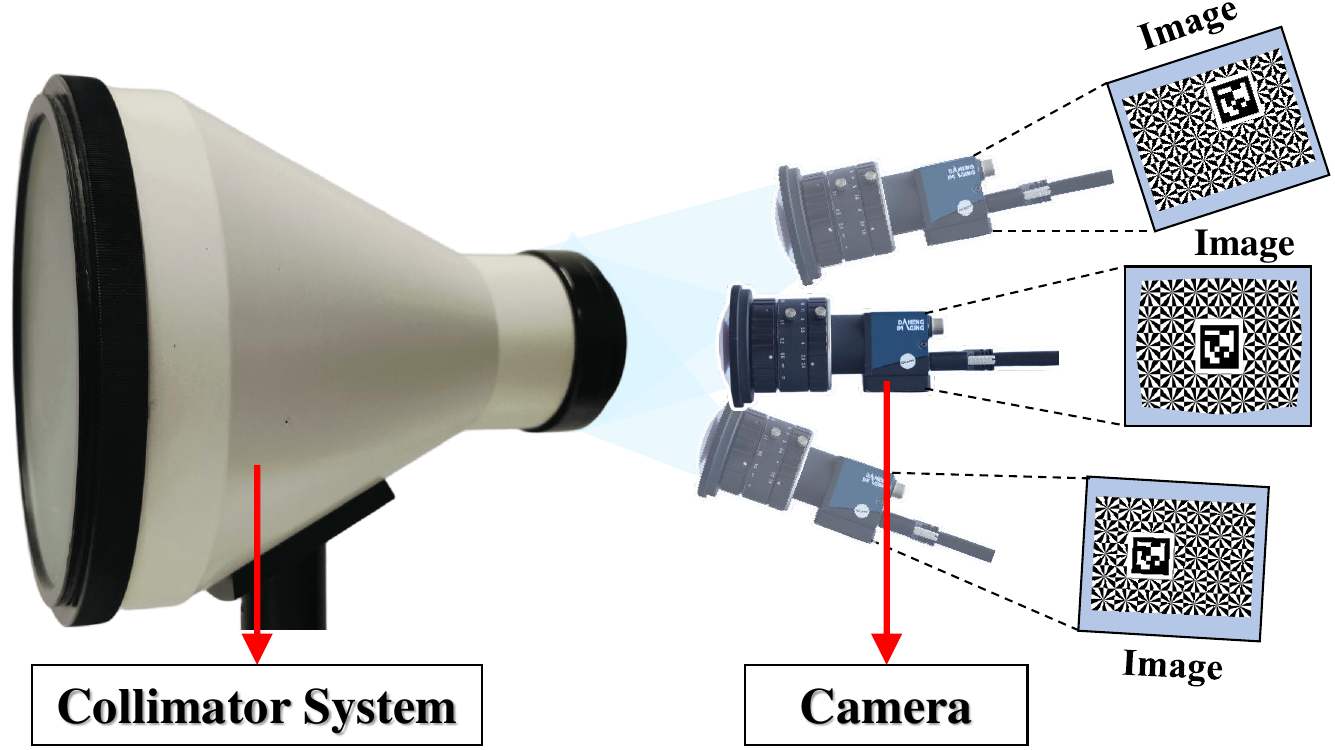}
	\caption{{Schematic diagram of our collimator-based calibration. The camera achieves precise calibration by capturing one or more images of the collimator system.}}
	\label{fig:coll_calib}
\end{figure}

In this paper, we design a collimator system for camera calibration. A known structured pattern is attached to the reticle of the collimator to provide calibration target. Based on the collimator image, a new calibration method is proposed. Our method only requires the camera to observe the pattern in the collimator system from various orientations, without the need for goniometer or theodolite, thereby offering greater flexibility. {Figure~\ref{fig:coll_calib} 
presents the real collimator system along with a diagram of the calibration method. The camera to be calibrated captures one or more collimator images from our system to achieve accurate calibration.} Unlike previous methods, the proposed considers the optical geometric properties of the collimator for the first time. Specifically, we demonstrate the angle invariance of any point pairs on the reticle, a direct consequence of the collimator's ability to produce parallel rays. By leveraging angle invariance, we further prove that the relative motion between the calibration target and the camera conforms to a spherical motion model. This constraint effectively reduces the original 6DOF general motion to the 3DOF pure rotation motion. 

Furthermore, we propose two calibration algorithms using collimator images. For multiple images, our algorithm combines spherical motion constraint with planar homography constraint to fully estimate the camera parameters. Compared with the target-based methods \citep{Zhang2000,Bouguet2004}, more geometric constraints are derived from the collimator, leading to higher accuracy. Compared to traditional collimator-based methods \citep{Hallert1963,Hieronymus2012}, {our method is more flexible because its structure is simple and eliminates the need for direction measurement devices.} When only a single collimator image is available, the proposed algorithm utilizes angle invariance to estimate the camera parameters. Calibration with a single collimator image allows us to quickly and batch determine camera parameters, significantly enhancing the flexibility and practicality of the calibration process.

Our primary contributions can be summarized as follows:
\begin{itemize}
	\item We present a collimator system specifically designed for camera calibration. Its advantageous optical properties provide a reliable and controllable calibration environment, suitable for calibrating lenses of various focal lengths.
	\item By exploring the collimator's property to produce parallel rays, we introduce an angle invariance constraint and prove that the relative motion between the calibration target and the camera conforms to the spherical motion model. This constraint effectively reduces the number of motion parameters to be estimated.
	\item For multiple calibration images, we propose two calibration solvers using the spherical motion constraint. One is a closed-form solver suitable for multiple images, and the other is a minimal solver designed for two images. Moreover, we analyze and prove two possible degenerate configurations.
	\item {We propose an algorithm for fast and flexible camera calibration using only a single collimator image.} This algorithm leverages angle invariance as a constraint to formulate a cost function that solely involves the intrinsic, thereby eliminating the need for camera motion and enabling rapid, batch calibration.
\end{itemize}

This paper is an extension of our conference paper published at ECCV 2024 \citep{Liang2024}. In the new version of the paper: 1) We offer a novel proof for the spherical motion model, providing enhanced clarity and simplicity. This motion is validated through both theoretical analysis and experiments using real collimator images. 2) We present two new degenerate configurations for multiple collimator image calibration and provide detailed proof of these degeneracy cases. 3) We propose a novel camera calibration algorithm using a single collimator image based on angle invariance. The proposed algorithm provides a new solution for rapid and batch camera calibration. 4) Additional real-data experiments have been supplemented, including validation of spherical motion, results of structure from motion, calibration results from a single collimator image, and more detailed discussions.

\section{Related Work}\label{sec:Related}
Camera calibration is a crucial subject in 3D vision and continues to be actively researched. Camera calibration comprises three essential components: camera model, calibration algorithm, and calibration target. The camera model is a mathematical representation of the camera's imaging process, with different models tailored for various types of cameras. Common cameras are typically modeled using a pinhole model with distortion \citep{Brown1971,Zhang2000}, whereas fisheye cameras are better described by models such as the Kannala-Brandt model \citep{Kannala2006}, the dual-sphere model \citep{Usenko2018Double}, or the omnidirectional model \citep{Scaramuzza2006Omnidirectional}. More generally, the generic camera model proposed by \cite{Grossberg2001general}, establishes individual mappings between image pixels and rays in 3D space. The literature on camera calibration introduces a wide range of algorithms and toolboxes \citep{Tsai1987,Sturm1999,Zhang2000,Kannala2006,Scaramuzza2006Omnidirectional,Thomas2020,Lochman2021}. Zhang's method \citep{Zhang2000} is a widely adopted technique that has been incorporated into various calibration toolboxes \citep{Opencv2000,Bouguet2004}. To achieve high calibration accuracy, targets with known structures are usually used for camera calibration. Various calibration targets have been proposed, including 3D targets \citep{Tsai1987}, 2D targets \citep{Heikkila2000,Ha2017} and 1D targets \citep{Zhang2004}. In scenarios where calibration targets are unavailable, self-calibration techniques can be used, which establish correspondences between images captured during camera motion \citep{Berlin1992, Pollefeys1999, Nister2005, Hartley1994, Herrera2016, Gustavo2017}. Among these options, the target-based close-range calibration method is widely used. 

In outdoor long-range measurement applications \citep{Zhang2020,Guan2022,lei2025event}, cameras operate at significant distances to measure distant targets. Providing effective calibration targets for the camera in these scenarios presents a challenge. {The close-range calibration methods based on planar targets \citep{Zhang2000,Thomas2020} may be impractical due to the site limitation and challenging lighting condition}. A common approach involves constructing a calibration field, which serves as a known structure for the calibration \citep{Liu2015,Hu2022,Wang2015,Shang2013,Xiao2010,Oniga2018}. For example, \cite{Oniga2018} placed markers on buildings to calibrate cameras mounted on UAVs. \cite{Shang2013} proposed a high-precision calibration method by placing several markers near the ground. Additionally, \cite{Klaus2004} proposed a method for camera calibration using stars in the night sky. This method utilizes the known precise angular positions of fixed stars to construct a calibration field. {\cite{Sagawa2008} regarded distant markers as pairs of parallel light rays and proposed a method for calibrating camera parameters.} Although these methods can effectively obtain camera parameters, they are complex to operate and lack flexibility.

The collimator, with its ability to generate parallel rays, can serve as an effective calibration target for long-distance cameras. Additionally, the collimator's built-in light source provides a stable and controllable calibration environment for the camera. In fact, as early as the mid-20th century, collimators have been used for camera calibration \citep{Hothmer1958,Hallert1963}. {Early collimator-based camera calibration methods primarily employ multi-collimator array structures \citep{Hallert1963,karren1968camera,tayman1978analytical}. In these configurations, each collimator provides only a single independent control point and precise pre-measurement of the entire array structure is essential.} The review by Clarke and Fryer offers a detailed discussion on the calibration of cameras using collimator \citep{Clarke1998}. Recently, collimators are also commonly used to provide targets for camera calibration in laboratory settings \citep{Wu2007,Bauer2008,Hieronymus2012,Yuan2019,Li2020Networks,Yuan2021,liu2025collimator}. {With the advancement of collimator technology and the expansion of its field of view, researchers have begun customizing multiple pinhole structures \citep{zhang2016multiple} or designing diffractive optical elements (DOE) \citep{Bauer2008} on a single collimator.} A comparative study by \cite{Hieronymus2012} compared three calibration methods (calibration field methods, collimator methods, and diffractive optical elements methods) and concluded the collimator method can also yield results of comparable accuracy. Previous collimator-based methods often relied on goniometer or theodolite to obtain the direction of collimated rays \citep{Yuan2021,Wu2007}. While offering both accuracy and flexibility, these methods require expensive, high-precision goniometers or theodolites. To address this, \cite{Yuan2019} designed a multiple pinhole mask for the collimator to generate the collimated rays with known directions. However, the accuracy of calibration is sensitive to machining errors. \cite{Bauer2008} proposed a calibration method using a diffractive optical element (DOE). The angle between the laser beam and the camera can then be estimated to determine the camera parameters. 


\section{Collimator System}\label{sec:Collimator}
This section presents the designed collimator system. {We analyze its optical geometry and demonstrate the properties of angle invariance and the spherical motion model.}

\subsection{Scheme of Collimator System} \label{sec:Scheme}
The collimator is an optical instrument designed to produce collimated (parallel) rays, playing an important role in the adjustment of other optical instruments. As illustrated in the left of Fig.~\ref{fig:Collimator_Camera}, a collimator consists of three essential components: a light source, a reticle, and an optical lens. The reticle is a thin glass element positioned precisely at the focal plane of the collimator's lens. Our collimator utilizes a planar array light source to ensure uniform illumination of the reticle patterns. This light source helps establish a consistent and controllable calibration environment, even in poor lighting conditions. The generation of collimated rays can be regarded as the inverse process of imaging a target at infinity onto the focal plane of the lens. Specifically, a light beam emanating from a point on the lens's focal plane will become a collimated beam after passing through the lens. When a camera observes this point, the point can be regarded as being located at infinity. {Therefore, cameras with varying parameters can consistently observe the target through the collimator at close range.} 

We design a collimator system for camera calibration and its core function is to provide an observable target. To this end, we attach a pattern with known structure to the reticle of the collimator as the calibration target. We select the star-based pattern, proposed by \cite{Thomas2020}, as it offers richer gradient information for improving feature extraction accuracy. The pattern's center is an AprilTag \citep{Wang2016} to ensure unambiguous localization. Similar to the plane-based calibration method \citep{Sturm1999,Zhang2000}, the star-based pattern provides 2D-3D correspondences for camera calibration. Our collimator system is portable with a size of about $200mm \times 170mm \times 300mm$. {Compared to multi-collimator arrays \citep{karren1968camera,tayman1978analytical}, our system utilizing a single collimator significantly simplifies the overall structural design. The multiple pinhole structures \citep{zhang2016multiple} or DOE \citep{Bauer2008} face challenges in effectively distinguishing feature and typically require manual feature extraction and matching. In contrast, the used star-based pattern supports fully automated feature extraction and matching, significantly improving calibration efficiency. Furthermore, the system does not rely on angular measurement devices, thereby simplifying the hardware configuration. As a result, it can be flexibly used in complex environments such as outdoor settings, and offers portability and applicability comparable to conventional 2D calibration targets.} The system is designed to provide a reliable and controllable calibration environment for some special scenarios like inappropriate illumination, limited space, and long-range cameras. However, printed targets face challenges in these scenarios. When calibrating a camera, we only need to capture the target from multiple orientations through the lens of the collimator, {which is as simple as using printed targets.}

\subsection{Spherical Motion Model} \label{sec:Motion Model}
\begin{figure*}[tbp]
	\centering
	\begin{minipage}[t]{0.45\linewidth}
		\centering
		\includegraphics[width=0.9\linewidth]{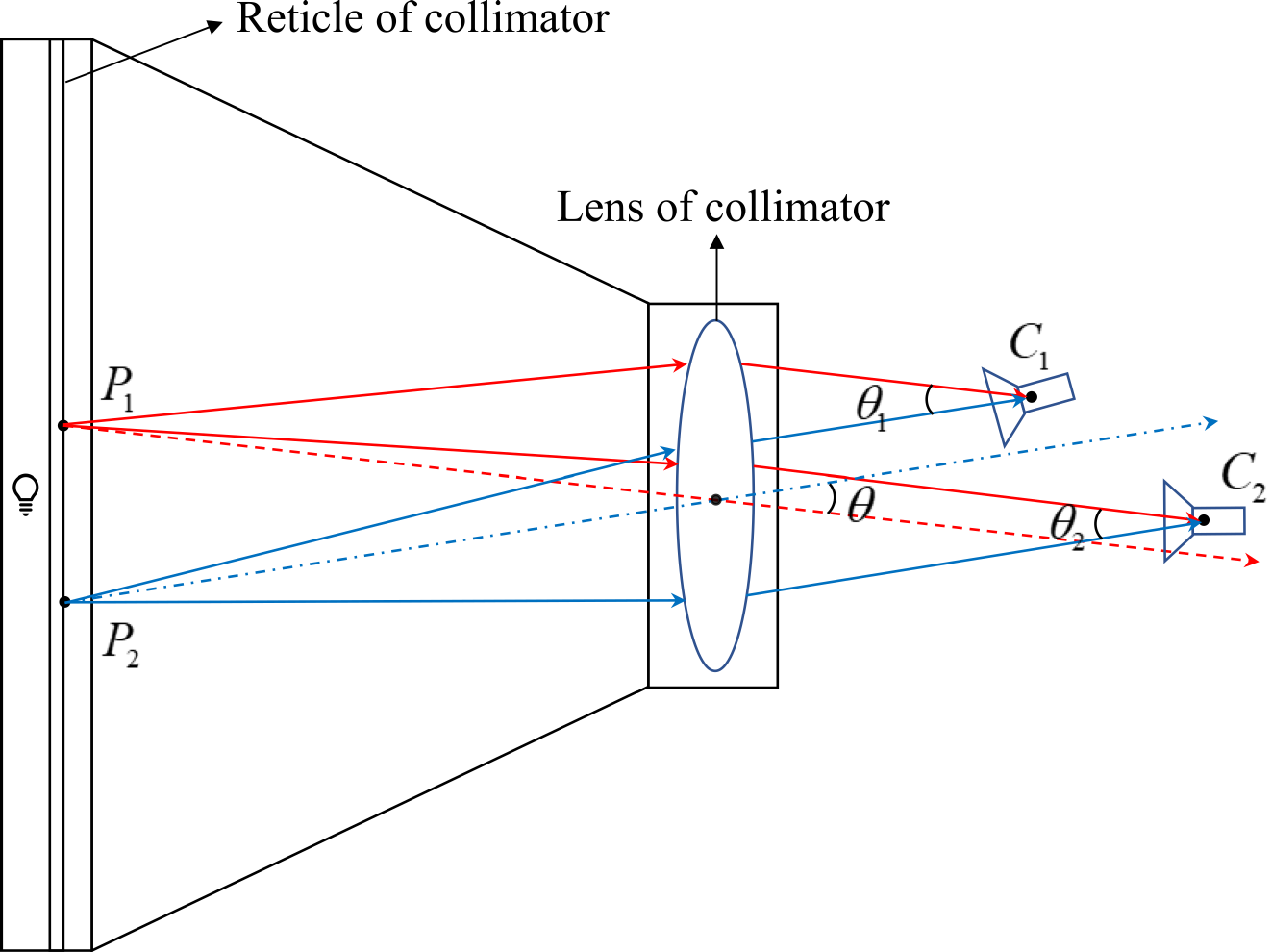}
	\end{minipage}
	\qquad
	\begin{minipage}[t]{0.45\linewidth}
		\centering
		\includegraphics[width=0.9\linewidth]{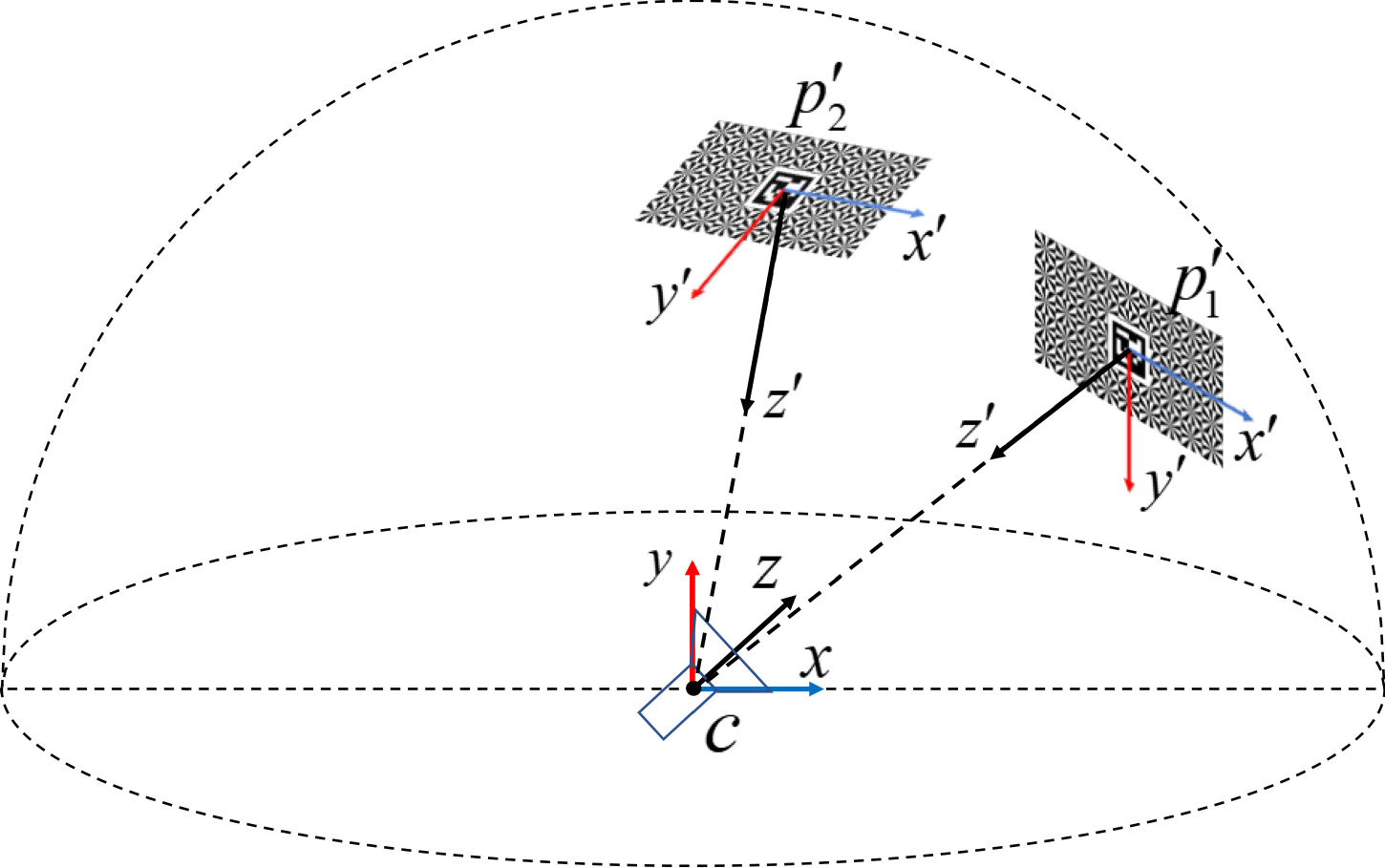}
	\end{minipage}
	\caption{\textbf{Left}: Schematic diagram of our collimator system. The geometric property of the collimator system leads to angle invariance of any point pair on the reticle; \textbf{Right}: Schematic diagram of spherical motion model. The angle invariance forces the relative motion between the calibration target and the camera to satisfy the spherical motion model.}
	\label{fig:Collimator_Camera}
\end{figure*}

Firstly, we briefly analyze the geometric property of the collimator and introduce the angle invariance. {As shown in the left of Fig.~\ref{fig:Collimator_Camera}}, let us note two arbitrary points $P_1$ and $P_2$ on the reticle of collimator. Rays emitted from $P_1$ and $P_2$ pass through the collimator lens, generating two parallel beams observed by two cameras ($C_1$ and $C_2$) with arbitrary poses. The propagation routes of the two beams are visually distinguished by red and blue lines. Obviously, the angle between two lines observed by the two cameras is equal, that is, $\theta_1 \equiv \theta_2 \equiv \theta$. Based on the optical geometry of collimators, we can obtain a property that the imaging angle between any given pair of points on the collimator reticle is invariant, regardless of the poses of observing cameras. {We refer to this fundamental property as \textbf{angle invariance constraint}.} 

Next, we show that this setup, where a camera moves around a collimator system and captures images, is theoretically equivalent to the capture setting where the camera is fixed, and the images move spherically around the camera. This equivalence is derived directly from the principle of angle invariance. For convenience in introducing our derivation process, we treat the camera as a fixed reference coordinate and assume that the calibration target moves relative to the camera in space. The angle invariance allows us to infer that when the target performs a specific motion in space, the angular distance between any point pair on the target always remains invariant. The angular distance means the angle between two vectors from the origin (camera optical center) to two points. By arbitrarily selecting a pair of points $\mathbf{P}_i$ and $\mathbf{P}_j$ on the target and transforming them, we can establish the following equation
\begin{equation}
	\angle\left(\mathbf{P}_i , \mathbf{P}_j \right) = \angle\left(\mathbf{R}\left(\mathbf{P}_i+\mathbf{t}\right), \mathbf{R}\left(\mathbf{P}_j +\mathbf{t}\right)\right),
	\label{eq:angPiPj}
\end{equation}
where $\mathbf{R} \in SO(3)$ is rotation matrix and $\mathbf{t} \in \mathbb{R}^3$ is translation vector. From the rotation invariance \citep{Wang2022} that a pair of 3D points are jointly rotated about the origin, their angular distance remains constant, we have $\angle\left(\mathbf{R}\left(\mathbf{P}_i+\mathbf{t}\right), \mathbf{R}\left(\mathbf{P}_j +\mathbf{t}\right)\right) = \angle\left(\mathbf{P}_i+\mathbf{t}, \mathbf{P}_j +\mathbf{t}\right)$. Thus, the angle invariance is held when the calibration target is arbitrarily rotated about the origin. Then, we get a new equation:
\begin{equation}
	\angle(\mathbf{P}_i, \mathbf{P}_j)) = 
	\angle(\mathbf{P}_i+\mathbf{t}, \mathbf{P}_j +\mathbf{t}).
	\label{eq:angPiPj+t}
\end{equation}
{Below, we present and prove a key proposition.}

\noindent{{\textbf{Proposition 1.} \textit{For any pair of points $\mathbf{P}_i$ and $\mathbf{P}_j$ on the target, the necessary and sufficient condition for Eq.~\eqref{eq:angPiPj+t} to hold is that the translation vector $\mathbf{t}$ is a $\mathbf{0}$ vector.}}}

{\textit{Proof.} Obviously, when $\mathbf{t} = [0,0,0]^T$, Eq.~\eqref{eq:angPiPj+t} holds, thereby proving the sufficiency. Next, we focus on demonstrating the necessity.}

Angle invariance holds for any point pair on the target, so it must hold for specific pairs. We select several special point pairs to support our proof. Without loss of generality, we establish a coordinate system with camera optical center as the origin. In this system, the $z$-axis is perpendicular to the plane of the target, and the $xy$-plane is parallel to the target. The distance from the origin to the target is $r(r>0)$. We select three mutually orthogonal points on the target: $\mathbf{A} = [\sqrt{2}r, 0, r]^T$, $\mathbf{B} = [-\frac{\sqrt{2}}{2}r, \frac{\sqrt{6}}{2}r, r]^T$ and $\mathbf{C} = [-\frac{\sqrt{2}}{2}r, -\frac{\sqrt{6}}{2}r, r]^T$. The angle distance of the pairs $(\mathbf{A},\mathbf{B})$, $(\mathbf{A},\mathbf{C})$ and $(\mathbf{B},\mathbf{C})$ are all $90^\circ$, that is, $\cos(\angle(\mathbf{A},\mathbf{B})) = \cos(\angle(\mathbf{A},\mathbf{C})) = \cos(\angle(\mathbf{B},\mathbf{C})) = 0$. 

Let us define $\mathbf{t}=(x,y,z)^T$ to represent the relative translation between the target and the origin. From Eq.~\eqref{eq:angPiPj+t}, we have:
\begin{equation}
	\begin{aligned}
		\cos(\angle(\mathbf{A},\mathbf{B})) &= \cos(\angle(\mathbf{A}+\mathbf{t},\mathbf{B}+\mathbf{t})) = \frac{(\mathbf{A}+\mathbf{t})^T(\mathbf{B}+\mathbf{t})}{\Vert \mathbf{A}+\mathbf{t} \Vert \Vert \mathbf{B}+\mathbf{t} \Vert } = 0, \\ 
		\cos(\angle(\mathbf{A},\mathbf{C})) &= \cos(\angle(\mathbf{A}+\mathbf{t},\mathbf{C}+\mathbf{t})) = \frac{(\mathbf{A}+\mathbf{t})^T(\mathbf{C}+\mathbf{t})}{\Vert \mathbf{A}+\mathbf{t} \Vert \Vert \mathbf{C}+\mathbf{t} \Vert } = 0, \\ 
		\cos(\angle(\mathbf{B},\mathbf{C})) &= \cos(\angle(\mathbf{B}+\mathbf{t},\mathbf{C}+\mathbf{t})) = \frac{(\mathbf{B}+\mathbf{t})^T(\mathbf{C}+\mathbf{t})}{\Vert \mathbf{B}+\mathbf{t} \Vert \Vert \mathbf{C}+\mathbf{t} \Vert } = 0.
	\end{aligned}
	\label{eq:angABC}
\end{equation}
By simplifying Eq.~\eqref{eq:angABC}, we obtain the following system of equations:
\begin{equation}
	\begin{aligned}
		(\mathbf{A} + \mathbf{B})^T \mathbf{t} + \mathbf{t}^T\mathbf{t} &= 0, \\
		(\mathbf{A} + \mathbf{C})^T \mathbf{t} + \mathbf{t}^T\mathbf{t} &= 0, \\
		(\mathbf{B} + \mathbf{C})^T \mathbf{t} + \mathbf{t}^T\mathbf{t} &= 0,
	\end{aligned}
	\label{eq:angABCt}
\end{equation}
Obviously, each equation in Eq.~\eqref{eq:angABCt} represents a sphere in space. The solution to this system of equations corresponds to the intersection point of the three spheres. By substituting the specific values of points $\mathbf{A}$, $\mathbf{B}$, and $\mathbf{C}$ into Eq.~\eqref{eq:angABCt} and solving the resulting system, we obtain two solutions: $\mathbf{t} = [0, 0, 0]^T$ and $\mathbf{t} = [0, 0, -2r]^T$. But the latter solution should be discarded, as practical scenarios dictate that the target cannot be positioned behind the camera. Up to now, we have proven that the relative translation between the target and camera must be a $\mathbf{0}$ vector. {Thus, the necessity is also established. \qed}

{Therefore, the calibration target undergoes pure rotation motion with respect to the origin (camera optical center).} Such pure rotational motion is a typical \textbf{spherical motion model}. In summary, the spherical motion model is established based on the inherent optical geometric properties of the collimator. Note that both the camera and collimator can move freely, without the need for moving in spherical motion. {The spherical motion is illustrated on the right of Fig.~\ref{fig:Collimator_Camera}.} In the spherical motion model, the calibration target exhibits only rotational motion with a fixed distance relative to the camera. We establish a local coordinate on the target, where the $z^{\prime}$-axis aligns with the ray between the camera coordinate system $c$ and calibration target coordinate system $p^{\prime}$. Inspired by \citep{Ventura2016}, the relative pose between the calibration target and camera can be mathematically represented as:
\begin{equation}
	\mathbf{T}_{cp^\prime} = \left[{\begin{array}{*{20}{c}}
			{\mathbf{R}_{cp^\prime}}&\mathbf{t}_{cp^\prime}\\
			{\mathbf{0}_{1 \times 3}}&1
	\end{array}} \right] ,
	\label{eq:Tcpp}
\end{equation}
where $\mathbf{R}_{cp^\prime} \in SO(3) $, $\mathbf{t}_{cp^\prime} = {(0,0,-r)^T}$ and $r$ is the radius of spherical motion. We usually establish the origin of the coordinate system $p$ on the upper left corner of the calibration pattern. Therefore, there is a relative translation $\mathbf{t}_{p^{\prime} p} = (x,y,0)$ between the coordinate systems $p^{\prime}$ and $p$. Furthermore, the transformation matrix between camera $c$ and calibration target $p$ is expressed as follows:
\begin{equation}
	\mathbf{T}_{pc} = \left[ \left[ {\begin{array}{*{20}{c}}
			{\mathbf{I}_{3\times3}}&\mathbf{t}_{p^\prime p}\\
			{\mathbf{0}_{1 \times 3}}&1
	\end{array}} \right] \left[ {\begin{array}{*{20}{c}}
			{\mathbf{R}_{cp^\prime}}&\mathbf{t}_{cp^\prime}\\
			{\mathbf{0}_{1 \times 3}}&1
	\end{array}} \right] \right]^{-1} =
	\left[ {\begin{array}{*{20}{c}}
			{{\mathbf{R}_{pc}}}&{ - {\mathbf{R}_{pc}}{\mathbf{t}_{cp}}}\\
			{\mathbf{0}_{1 \times 3}}&1
	\end{array}} \right],
	\label{eq:Tpc}
\end{equation}
where $\mathbf{R}_{pc} = \mathbf{R}_{cp^\prime} ^ {T}$ and  ${\mathbf{t}_{cp}}\!=\!\mathbf{t}_{p^{\prime} p}\!+\!\mathbf{t}_{cp^\prime}\!=\!(x,y,-r)^T$ determines the position of the camera in the calibration target coordinate system. Note that $\mathbf{t}_{cp}$ is a fixed value and has nothing to do with the pose of the calibration target. Given the constraint from the spherical motion, the original 6DOF general motion is limited to a 3DOF pure rotation motion. 
\section{Calibration Method for Multiple Images} \label{sec:Calibration}
The spherical motion constraint effectively reduces the number of motion parameters to be estimated. In this section, {we present methods for calibrating camera parameters using a set of collimator images.} A closed-form solver for $N (N>2)$ images and a minimal solver for two images are proposed.

\subsection{Geometric Constraints}
This section introduces the geometric constraints used in calibration. {We use the well-known pinhole camera model.} The mapping from a 3D point $\mathbf{P} = (X,Y,Z)^T$ to an image point $\mathbf{p} = (u,v)^T$ is computed as:
\begin{equation}
	s \widetilde{\mathbf{p}}=\mathbf{K}[\mathbf{R} \mid \mathbf{t}] \widetilde{\mathbf{P}} \text {, with } \mathbf{K}=\left[\begin{array}{ccc}
		f_{x} & \gamma & c_{x} \\
		0 & f_{y} & c_{y} \\
		0 & 0 & 1
	\end{array}\right],
	\label{eq:KRtP}
\end{equation}
where $s$ is an arbitrary scale factor, $\widetilde{\mathbf{p}} = [u,v,1]^T$ and $\widetilde{\mathbf{P}} = [X,Y,Z,1]^T$ denote the homogeneous coordinate forms of points $\mathbf{p}$ and $\mathbf{P}$, respectively. The matrix $[\mathbf{R} \mid \mathbf{t}]$ is the camera pose with respect to the calibration target coordinate. The intrinsic matrix $\mathbf{K}$ consists of five parameters: the focal length $(f_x,f_y)$, the principal point $(c_x,c_y)$ and skew factor $\gamma$. These parameters need to be estimated during camera calibration.

Our method requires the camera to observe the calibration target through the collimator system from multiple orientations, and use point correspondences between the target and the images to estimate the camera parameters. Without loss of generality, we assume the planar target is on the plane $Z = 0$ of the world coordinate system. The transformation between the image plane and calibration target plane can be constrained by a $3 \times 3$ homography matrix, denoted as $\mathbf{H}$. For perspective projection, the homography matrix $\mathbf{H}$ depends on the image pose $[\mathbf{R} \mid \mathbf{t}]$ and intrinsic matrix $\mathbf{K}$:
\begin{equation}
	\mathbf{H} = \left[ {\begin{array}{ccc}
			\mathbf{h}_1 & \mathbf{h}_2 & \mathbf{h}_3
	\end{array}} \right] = \lambda \mathbf{K} \left[ {\begin{array}{ccc}
			\mathbf{r}_1&\mathbf{r}_2&\mathbf{t}
	\end{array}} \right],
	\label{eq:H}
\end{equation}
where $\lambda$ is an arbitrary scalar, {$\mathbf{r}_i$ is the $i$-th column of $\mathbf{R}$.} Given an image of the calibration target, the homography matrix $\mathbf{H}$ can be estimated using the plane point correspondences \citep{Hartley2004}. Combining spherical motion constraint (Eq.~\eqref{eq:Tpc}) and planar homography constraint (Eq.~\eqref{eq:H}), we get:
\begin{equation}
\left[ {\begin{array}{*{20}{c}}
			\mathbf{{r_1}}&\mathbf{{r_2}}& -\mathbf{R} \mathbf{t}_{cp}
	\end{array}} \right] = \frac{1}{\lambda} \mathbf{K}^{-1} \mathbf{H}.
\label{eq:r1r2t}
\end{equation}
{By multiplying the matrices on both sides of Eq.~\eqref{eq:r1r2t} by their transposes and using the orthogonality of the rotation matrix $\mathbf{R}$, we can obtain:}
\begin{align}
	{\mathbf{H}^T}{\mathbf{K}^{ - T}}{\mathbf{K}^{ - 1}} \mathbf{H} = \lambda^2 \left[ {\begin{array}{ccc}
			1&0&{ - x}\\
			0&1&{ - y}\\
			{ - x}&{ - y}&{{{\left\| {{\mathbf{t}_{cp}}} \right\|}^2}}
	\end{array}} \right].
	\label{eq:HKKH}
\end{align}
There are 8 parameters to be estimated, 5 for intrinsic matrix and 3 for $\mathbf{t}_{cp} = (x,y,-r)^T$. For each given image, Eq.~\eqref{eq:HKKH} provides 5 independent constraints. Theoretically, a minimum of two images can completely solve all unknown parameters. In the next two subsections, we give a closed-form solver for $N(N>2)$ collimator images and a minimal solver for two collimator images.

\subsection{Closed-form Solver for $N$ Images}
When there are more than two collimator images for calibration, we model the problem as a linear system and propose a closed-form solution. Suppose that the homography matrices $\mathbf{H}_i (i = 1,...,N)$ have been estimated for all images. Let us define $\mathbf{M} = \left[ {\begin{array}{*{20}{c}}
		\mathbf{{r_1}}&\mathbf{{r_2}}& -\mathbf{R} \mathbf{t}_{cp}
\end{array}} \right]$, the transformation between images $i$ and $j$ is expressed as:
\begin{equation}
	\mathbf{H}_j^{-1} \mathbf{H}_i = \frac{\lambda_i}{\lambda_j} \mathbf{M}_j^{-1} \mathbf{M}_i.
	\label{eq:HjHi}
\end{equation}
For the matrix $\mathbf{M}$ of each image, we can calculate the determinant as:
\begin{equation}
	\begin{aligned}
		\left|\mathbf{M} \right| &= \left| \begin{array}{ccc} \mathbf{r}_1&\mathbf{r}_2&-\mathbf{R}\mathbf{t}_{cp} \end{array} \right| \\
		&= \left| \begin{array}{ccc} \mathbf{r}_1 & \mathbf{r}_2 & -x\mathbf{r}_1-y\mathbf{r}_2+r\mathbf{r}_3 \end{array} \right| \\
		&= -x\left| \begin{array}{ccc} \mathbf{r}_1 & \mathbf{r}_2 & \mathbf{r}_1 \end{array} \right| 
		- y\left| \begin{array}{ccc} \mathbf{r}_1 & \mathbf{r}_2 & \mathbf{r}_2 \end{array} \right| 
		+ r\left| \begin{array}{ccc}
			\mathbf{r}_1 & \mathbf{r}_2 & \mathbf{r}_3 \end{array} \right| \\ &= r.
	\end{aligned}
	\label{eq:detM}
\end{equation}
{The determinant of matrix $\mathbf{M}$ is a non-zero constant, which which ensures its nonsingularity and invertibility. Furthermore, we can obtain the determinant constraint: $\left|\mathbf{M}_j^{-1} \mathbf{M}_i \right| = 1$.} Combined with Eq.~\eqref{eq:HjHi}, we have $\lambda_{ij} = \lambda_i / \lambda_j = \sqrt[3]{|\mathbf{H}_j^{-1} \mathbf{H}_i|}$. Then, we choose the image with the most feature points observed as the base image. For convenience, here the first image is selected as the base and we get $\lambda_{i1} = \sqrt[3]{|\mathbf{H}_i^{-1} \mathbf{H}_1|}$. 
By inverting Eq.~\eqref{eq:HKKH} of image $i$, the constraint is re-written as:
\begin{equation}
	\mathbf{H}_i^{-1} \mathbf{K} \mathbf{K}^T \mathbf{H}_i^{-T} = 
	\frac{1}{\lambda_i^2 r^2} \left[ {\begin{array}{*{20}{c}}
			{{r^2} + {x^2}}&{xy}&x\\
			{xy}&{{r^2} + {y^2}}&y\\
			x&y&1
	\end{array}} \right].
	\label{eq:HKKH2}
\end{equation}
Substitute $\lambda_i = \lambda_{i1} \lambda_1$ into Eq.~\eqref{eq:HKKH2}, we get
\begin{equation}
	\mathbf{H}_i^{-1} \mathbf{K} \mathbf{K}^T \mathbf{H}_i^{-T} =
	\frac{1}{\lambda_{i1}^2} \mathbf{A}, \text{ with }	\mathbf{A} = \left[ {\begin{array}{{ccc}}
			\frac{r^2+x^2}{(\lambda_1 r)^2}&\frac{xy}{(\lambda_1 r)^2}& \frac{x}{(\lambda_1 r)^2}\\
			\frac{xy}{(\lambda_1 r)^2}&\frac{r^2+y^2}{(\lambda_1 r)^2}& \frac{y}{(\lambda_1 r)^2}\\
			\frac{x}{(\lambda_1 r)^2}&\frac{y}{(\lambda_1 r)^2}& \frac{1}{(\lambda_1 r)^2}
	\end{array}} \right].
	\label{eq:HKKH3}
\end{equation}
Let us define a 6-vector $\mathbf{a}$ to represent matrix $\mathbf{A}$:
\begin{equation}
	\mathbf{a} = (A_{11}, A_{12}, A_{13}, A_{22}, A_{23}, A_{33})^T,
\end{equation}
where $A_{mn}$ is the value of $m$-th row and $n$-th column of matrix $\mathbf{A}$. {Let us define:}
\begin{equation}
	\begin{aligned}
		\mathbf{W} = \mathbf{K} \mathbf{K}^T = \left[\begin{array}{ccc}
			{c_x}^2 + {f_x}^2 + \gamma^2 & c_x c_y + f_y \gamma & c_x \\
			c_x c_y + f_y \gamma & {c_y}^2 + {f_y}^2 & c_y \\
			c_x & c_y & 1
		\end{array}\right].
	\end{aligned}
	\label{eq:W}
\end{equation}
{Note that $\mathbf{W}$ is a symmetric matrix and can be represented by a 5-vector:}
\begin{equation*}
	\mathbf{w} = (W_{11}, W_{12}, W_{13}, W_{22}, W_{23})^T, 
\end{equation*}
where $W_{mn}$ is the value of $m$-th row and $n$-th column of matrix $\mathbf{W}$. Obviously, Eq.~\eqref{eq:HKKH3} represents a linear system of equations involving the unknowns $[\mathbf{w}^T, \mathbf{a}^T]^T$. Combining Eq.~\eqref{eq:HKKH3} and Eq.~\eqref{eq:W}, we can get the following equations for each image:
\begin{equation}
	\left[ {\begin{array}{*{20}{c}}
			{\mathbf{V}}&{ - {\lambda _{i1}^2} {\mathbf{I}_{5 \times 5}}}
	\end{array}} \right]\left[ {\begin{array}{*{20}{c}}
			\mathbf{w}\\ \mathbf{a}
	\end{array}} \right] = \mathbf{b},
	\label{eq:VI}
\end{equation}
with
\begin{equation*}
	\begin{aligned}
		&\mathbf{V} = [\mathbf{v}_{11}, \mathbf{v}_{12}, \mathbf{v}_{13}, 
		\mathbf{v}_{22}, \mathbf{v}_{23}]^T, \\
		&\mathbf{v}_{mn} =  {\left[h_{m1} h_{n1}, h_{m1} h_{n2}+h_{m2} h_{n1}, h_{m1} h_{n3}+h_{m3} h_{n1},\right.} \\
		&\left. \qquad \quad \ h_{m2} h_{n2}, h_{m2} h_{n3} + h_{m3} h_{n2}\right], \\
		&\mathbf{b} = - [h_{13}h_{13}, h_{13}h_{23}, h_{13}h_{33}, h_{23}h_{23}, h_{23}h_{33}]^T, 
	\end{aligned}
\end{equation*}
where $h_{mn}$ denotes the value of $m$-th row and $n$-th column of matrix $\mathbf{H}^{-1}$. Given $N(N>2)$ collimator images, we can stack $N$ such equations as Eq.~\eqref{eq:VI} together in matrix form:
\begin{equation}
	\mathbf{D} \left[ {\begin{array}{*{20}{c}}
			\mathbf{w}\\ \mathbf{a}
	\end{array}} \right] = \mathbf{b},
	\label{eq:Dwa}
\end{equation}
where $\mathbf{D}$ is a $5N\times11$ matrix. The closed-form solution of Eq.~\eqref{eq:Dwa} is given by
\begin{equation}
	\left[ {\begin{array}{*{20}{c}}
			\mathbf{w}\\ \mathbf{a}
	\end{array}} \right] =
	(\mathbf{D}^T \mathbf{D})^{-1}\mathbf{D}^T \mathbf{b}.
\end{equation}
Once $\mathbf{w}$ and $\mathbf{a}$ are estimated, we can calculate all camera intrinsic parameters and camera optical center $\mathbf{t}_{cp}$ by decomposing $\mathbf{W}$ and $\mathbf{A}$ respectively:
\begin{equation*}
\begin{aligned}
	c_x &= W_{13}, \quad c_y = W_{23}, \quad f_y = \sqrt{W_{22} - {c_y}^2}, \\
	\gamma &= (W_{12} - c_x c_y)/f_y, \quad	f_x = \sqrt{W_{11} - {c_x}^2 - {\gamma}^2}, \\
	x &= A_{13} / A_{33}, \quad y = A_{23} / A_{33}, \\	
	r &= -\sqrt{(A_{11}/A_{33}) - (A_{13}/A_{33})^2}.
\end{aligned}
\end{equation*}
{Subsequently,} the rotation matrix of each image can be calculated with Eq.~\eqref{eq:H}:
\begin{equation}
	\mathbf{r}_1 = (1/\lambda ){\mathbf{K}^{ - 1}} \mathbf{h}_1, \quad
	\mathbf{r}_2 = (1/\lambda ){\mathbf{K}^{ - 1}}\mathbf{h}_2, \quad
	\mathbf{r}_3 = \mathbf{r}_1 \times \mathbf{r}_2,
\end{equation}
where $\lambda = (\left \| {\mathbf{K}^{ - 1}} \mathbf{h}_1 \right \| + \left \| {\mathbf{K}^{ - 1}} \mathbf{h}_2 \right \|)/2$. Finally, a singular value decomposition for re-orthogonalizing the rotation matrix  $\mathbf{R} = [\mathbf{r}_1, \mathbf{r}_2, \mathbf{r}_3]$ is required.

\subsection{Minimal Solver for Two Images}
When only two collimator images are used for calibration, {the rank of the coefficient matrix in Eq.~\eqref{eq:Dwa} is at most 10. Directly solving this rank-deficient linear system of equations will yield a solution set. The true solution is a linear combination of this solution set, with coefficients that need to be computed through the nonlinear constraints imposed by the parameters $\mathbf{w}$ and $\mathbf{a}$. To avoid the complex discussion of solution set, we construct a nonlinear system and propose a novel solution approach.} This method utilizes the hidden variable technique \citep{Hartley2012} to estimate the unknowns.

Here, {we define $\mathbf{Q} = \mathbf{K}^{-T} \mathbf{K}^{-1}$}, which is known as the image of the absolute conic \citep{Hartley2004}. {Estimation of $\mathbf{Q}$ is the core of calibration. Matrix $\mathbf{Q}$ is symmetric and can be represented by a 6-vector $\mathbf{q} = (Q_{11}, Q_{12}, Q_{13}, Q_{22}, Q_{23}, Q_{33})^T$.} From Eq.~\eqref{eq:HKKH}, {we have the following 5 constraints for each image:} 
\begin{align}
\mathbf{u}_{12}^T \mathbf{q} &= 0, \label{eq:v12} \\
(\mathbf{u}_{11}^T - \mathbf{u}_{22}^T) \mathbf{q} &= 0, \label{eq:v11}\\
\mathbf{u}_{13}^T \mathbf{q} + x \mathbf{u}_{11}^T \mathbf{q} &=0, \label{eq:v13} \\
\mathbf{u}_{23}^T \mathbf{q} + y \mathbf{u}_{11}^T \mathbf{q} &=0, \label{eq:v23} \\
\mathbf{u}_{33}^T \mathbf{q} -  {\left\| {{\mathbf{t}_{cp}}} \right\|^2} \mathbf{u}_{11}^T \mathbf{q} &=0 \label{eq:v33}.
\end{align}
{The vector $\mathbf{u}_{mn}$ can be calculated as:}
\begin{equation*}
	\begin{aligned}
		\mathbf{u}_{mn}= & {\left[h_{1m} h_{1n}, h_{1m} h_{2n}+h_{1n} h_{2m}, h_{1m} h_{3n} + h_{1n} h_{3m}\right.} \\
		& \left.h_{2m} h_{2n}, h_{2m} h_{3n}+h_{2n} h_{3m}, h_{3m}h_{3n}\right]^{T},
	\end{aligned}
\end{equation*}
where $h_{mn}$ denotes the value of $m$-th row and $n$-th column of matrix $\mathbf{H}$. Given 2 collimator images, we can obtain 10 independent equations for 8 unknowns. To efficiently solve this over-determined nonlinear system, the hidden variable technique \citep{Hartley2012} is employed to facilitate the solution of the problem. Specifically, combining Eq.~\eqref{eq:v13} , Eq.~\eqref{eq:v23} and Eq.~\eqref{eq:v33}, we obtain a new equation:
\begin{equation}
	(\mathbf{u}_{13}^T + \mathbf{u}_{23}^T + \mathbf{u}_{33}^T)\mathbf{q} + c \mathbf{u}_{11}^T \mathbf{q} =0,
	\label{eq:v1233}
\end{equation}
where $c = (x+y-{\left\| {{\mathbf{t}_{cp}}} \right\|^2})$. We first select constraints Eq.~\eqref{eq:v12},Eq.~\eqref{eq:v11} and Eq.~\eqref{eq:v1233} to establish a new polynomial equation system. Next, we treat $c$ as a hidden variable, and the complete set of equations can be expressed in matrix form as:
\begin{equation}
	\mathbf{C}(c) \mathbf{q}=0, \text{ with } \mathbf{C}(c) = \left[ {\begin{array}{*{20}{c}}
			{\mathbf{u}_{12}^T}\\
			{(\mathbf{u}_{11}^T - \mathbf{u}_{22}^T)}\\
			{	(\mathbf{u}_{13}^T + \mathbf{u}_{23}^T + \mathbf{u}_{33}^T) + c\mathbf{u}_{11}^T}
	\end{array}} \right].
\end{equation}
The coefficient matrix $\mathbf{C}(c)$ contains the hidden variable $c$. Two images provide 6 equations for solving $c$, and $\mathbf{C}(c)$ is a $6 \times 6$ matrix. {Since the equation system $\mathbf{C}(c) \mathbf{q}=0$ has at least one nontrivial solution for $\mathbf{q}$}, the determinant of the coefficient matrix $\mathbf{C}(c)$ must be equal to zero, that is, $\det (\mathbf{C}(c)) =0 $. This determinant equation is a quadratic polynomial in terms of $c$, which can be efficiently solved using a straightforward root-finding algorithm. Substituting the known value of $c$ into the constraint equations Eq.~\eqref{eq:v12}, Eq.~\eqref{eq:v11} and Eq.~\eqref{eq:v1233}, {we obtain six constraint equations about $\mathbf{q}$. The least squares method is then employed to solve these linear equations and compute the optimal $\mathbf{q}$ up to scale. Once $\mathbf{q}$ is estimated,} the intrinsic matrix $\mathbf{K}$ can be calculated using the Cholesky factorization.

\subsection{Bundle Adjustment}
Up to now, we have obtained an initial guess of camera parameters by minimizing algebraic error. The physical refinement step is responsible for the final calibration accuracy. Minimizing re-projection error is a standard cost function in bundle adjustment \citep{Triggs2000}, widely used in various application's refinement step.

In our calibration method, bundle adjustment uses the minimization of re-projection error as the cost function to jointly refines the camera intrinsic $\mathbf{K}$, distortion coefficients $\mathbf{d}$, rotation of the $i$-th image $\mathbf{R}_i$ and camera optical center $\mathbf{t}_{cp}$. In cases of moderate lens distortion, {it is typical to initialize the distortion coefficients to $\mathbf{0}$ and estimate them during the bundle adjustment step \citep{Herrera2016}.} Given a set of $N$ images, each with $M$ corresponding 2D image points, the cost function is expressed as follows:
\begin{equation}
	\mathop {\min }\limits_\mathbf{X} \sum\limits_{i=1}^N {\sum\limits_{j=1}^M {\rho \left( {{{\left\| {\pi \left( {\mathbf{K}, \mathbf{d}, {\mathbf{R}_i},{\mathbf{t}_{cp}},{\mathbf{P}_{ij}}} \right) - {\mathbf{p}_{ij}}} \right\|}^2}} \right)} },
	\label{eq:BA}
\end{equation}
where $\mathbf{X} = (\mathbf{K},\mathbf{d},{\mathbf{R}_i},{\mathbf{t}_{cp}})$ is the parameter to be refined. This step iteratively refines the parameters $\mathbf{X}$, to minimize the difference between the 2D image points obtained by projecting 3D pattern points $\mathbf{P}_{ij}$ using the projection function $\pi(\cdot)$ and the real observed image points $\mathbf{p}_{ij}$. To enhance robustness against outliers, we employ the Cauchy robust function $\rho(\cdot)$. The optimization process is commonly performed using the Levenberg-Marquardt method \citep{LM}, {which is efficiently implemented in well-known optimization frameworks such as Ceres Solver \citep{Ceres}.}

Note that the camera optical center is fixed at $\mathbf{t}_{cp}$ due to the spherical motion constraint. However, for general motion, the camera optical center is free. Therefore, when calibrating the camera using collimator images, the number of motion parameters to be refined is reduced from $6N$ to $3N+3$. {Fewer parameters reduce the parameter complexity of the optimization problem. The synthetic data experiments demonstrate that fewer parameters contribute to achieving better calibration results.}

\subsection{Degenerated Configuration}
In this section, {we discuss potential degenerate configurations that may arise during camera calibration using our collimator system.} We identify two specific motions that can lead to degeneracy when camera captures images from the collimator. 

The first degenerate motion: \textit{Pure translational motion of the camera does not generate new observed image points.} 

In Sect.~\ref{sec:Scheme}, when the camera observes points on the focal plane of the collimator, these points can be considered at infinity. Equivalently, when observing points at infinity, pure translation will not change the observed directions. The pixel position of these points remains constant within the image plane. Therefore, {this motion does not yield any new observed image points and is considered degenerate.}

\begin{figure}[htbp]
	\centering
	\includegraphics[width=0.5\linewidth]{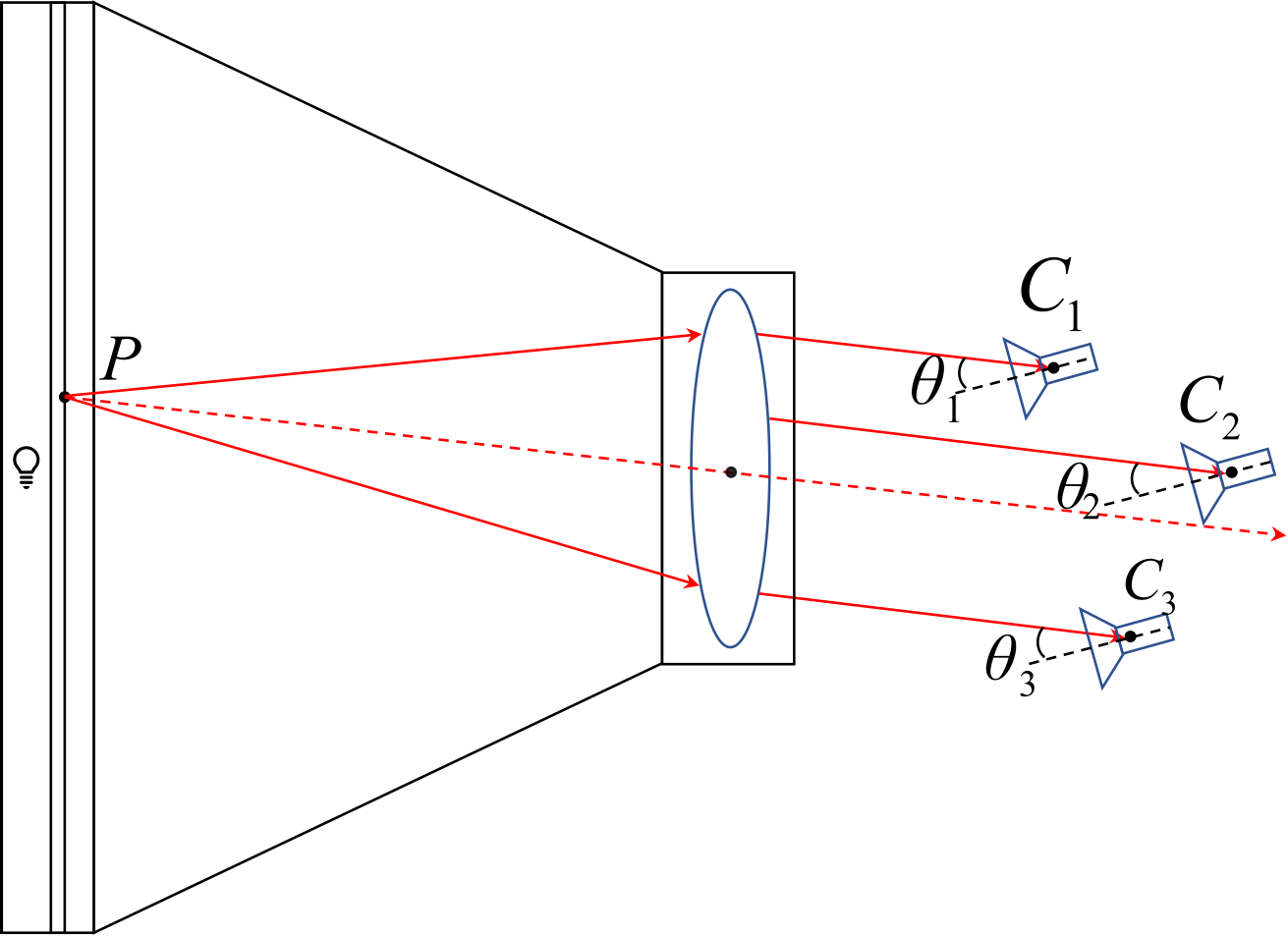}
	\caption{Schematic diagram of the camera capturing images with pure translation. Since the observation rays of $P$ are parallel, the angle between the optical axis and the observation ray is unchanged after the camera's pure translation ($\theta_1 = \theta_2 = \theta_3$). This motion does not generate new image points, resulting in degeneracy.}
	\label{fig:pure_translation}
\end{figure}

Naturally, a more readily comprehensible schematic representation could be used to illustrate this degenerate case. As illustrated in Fig.~\ref{fig:pure_translation}, three cameras ($C_1$, $C_2$, and $C_3$), exhibiting only relative translational displacement, observe a point $P$ on the reticle via the collimator system. The angles between the optical axes of three cameras (marked with black dashed lines) and the three observation rays to point $P$ are $\theta_1$, $\theta_2$, and $\theta_3$, respectively. Since the three observation rays (marked with red lines) are parallel, three angles are equal when the cameras have no relative rotation ($\theta_1 = \theta_2 = \theta_3$). This means that the image point of $P$ is invariant across three cameras. Consequently, pure translational motion of the cameras yields no new observations, {leading to a degenerate configuration.}

The second degenerate motion: \textit{Pure rotation of the camera around the $z$-axis does not provide new constraints for camera parameters.} 

This degeneracy differs from the first one; {it provides new image points but does not provide additional constraints on the camera parameters.} Following \citep{Zhang2000}, a newly added image is regarded as degraded when it cannot provide more constraints for solving the camera parameters. In the spherical motion model, the calibration target exhibits only 3DOF rotational motion with respect to the camera. The second degenerate motion, manifested in spherical motion, is equivalent to the following degenerate configuration.

{\textbf{Proposition 2.}} \textit{If the calibration target rotates only around an axis that is parallel to the $z$ axis and intersects the target plane at $(x,y,0)^T$, {the newly added image does not provide additional constraints on the camera intrinsic parameters}.}

\textit{Proof.} We will use the superscript ($^\prime$) to distinguish the matrix related to the newly added and original images. Let $\mathbf{T}_{pc}$ and $\mathbf{T}_{pc}^{\prime}$ represent the poses of the calibration target in the original and newly added images, respectively. In our degenerate configuration, pose $\mathbf{T}_{pc}^{\prime}$ can be expressed as:
\begin{equation}
	\mathbf{T}_{pc}^{\prime} = \begin{bmatrix}
		\mathbf{R}_{pc}\mathbf{R}_z & - \mathbf{R}_{pc} \mathbf{R}_z\mathbf{t}_{cp} \\
		\mathbf{0} & 1
	\end{bmatrix},
	\quad \text{with} \quad
	\mathbf{R}_z = \begin{bmatrix}
		\cos \theta & -\sin \theta & 0 \\
		\sin \theta & \cos \theta & 0 \\
		0 & 0 & 1
	\end{bmatrix},
	\label{eq:RbzRa}
\end{equation}
{where $\theta$ is the rotation angle.} Then, the components of pose $\mathbf{T}_{pc}^{\prime}$ can be expressed as functions of the components of $\mathbf{T}_{pc}$. Following the expression of Eq.~\eqref{eq:H}, we get:
\begin{equation}
	\begin{aligned}
		\mathbf{r}_1^\prime &= \mathbf{r}_{1} \cos \theta+\mathbf{r}_{2} \sin \theta, \\
		\mathbf{r}_2^\prime &= -\mathbf{r}_{1} \sin \theta+\mathbf{r}_{2} \cos \theta, \\
		\mathbf{t}^\prime &= r \mathbf{r}_3 - x \mathbf{r}_1^\prime - y \mathbf{r}_2^\prime.
	\end{aligned}
	\label{eq:rrt}
\end{equation}
Substituting Eq.~\eqref{eq:rrt} into Eq.~\eqref{eq:H}, each component of the homography matrix $\mathbf{H}^\prime$ is expressed as:
\begin{equation}
	\begin{aligned}
		\mathbf{h}_{1}^{\prime}&=\frac{\lambda^{\prime}}{\lambda}\left(\mathbf{h}_{1} \cos \theta+\mathbf{h}_{2} \sin \theta\right), \\
		\mathbf{h}_{2}^{\prime}&=\frac{\lambda^{\prime}}{\lambda}\left(-\mathbf{h}_{1} \sin \theta+\mathbf{h}_{2} \cos \theta\right),\\
		\mathbf{h}_{3}^{\prime}& = \frac{\lambda^{\prime}}{\lambda}\left[\mathbf{h}_{3}  +  (x  +  y \sin \theta  -  x \cos \theta) \mathbf{h}_{1}  \right. \\
		 & \left. \qquad \qquad +(y  -  x \sin \theta  -  y \cos \theta) \mathbf{h}_{2}\right].
	\end{aligned}
	\label{eq:ha123}
\end{equation}
Substituting Eq.~\eqref{eq:ha123} into constraint Eq.~\eqref{eq:HKKH}, the constraints on camera parameter from newly added $\mathbf{H}^{\prime} = [\mathbf{h}_{1}^{\prime} \quad \mathbf{h}_{2}^{\prime} \quad \mathbf{h}_{3}^{\prime}]$ can be expressed as a function of the constraints from $\mathbf{H} = [\mathbf{h}_{1} \quad \mathbf{h}_{2} \quad \mathbf{h}_{3}]$:
\begin{equation}
	\mathbf{H}^{\prime T} \mathbf{K}^{-T} \mathbf{K}^{-1} \mathbf{H}^\prime = \mathcal{F} \left( \mathbf{H}^{T} \mathbf{K}^{-T} \mathbf{K}^{-1} \mathbf{H} \right).
\end{equation}
We can enumerate that these constraints obtained from $\mathbf{H}^{\prime}$ are linear combinations of the constraints from $\mathbf{H}$. For example, the constraint of the 1-st row and 2-nd column can be calculated as:
\begin{equation}
	\begin{split}
		\mathbf{h}_{1}^{\prime T} \mathbf{K}^{-T} \mathbf{K}^{-1} \mathbf{h}_{2}^{\prime} &= \left(\frac{\lambda'}{\lambda}\right)^2 \left[ \cos 2\theta \left(\mathbf{h}_{1}^{T} \mathbf{K}^{-T} \mathbf{K}^{-1} \mathbf{h}_{2}\right) \right. \\
		&\quad \left. - \cos \theta \sin \theta \left(\mathbf{h}_{1}^{T} \mathbf{K}^{-T} \mathbf{K}^{-1} \mathbf{h}_{1} - \mathbf{h}_{2}^{T} \mathbf{K}^{-T} \mathbf{K}^{-1} \mathbf{h}_{2}\right) \right].
	\end{split}
	\label{eq:haKKha}
\end{equation}
Obviously, Eq.~\eqref{eq:haKKha} is a linear combination of the three constraints from $\mathbf{H}$. Likewise, the remaining constraints from $\mathbf{H}^\prime$ can also be expressed as linear combinations of constraints provided by $\mathbf{H}$. {This implies that the newly added $\mathbf{H}^\prime$ does not provide additional constraints for solving the camera parameters.} \qed

\section{Calibration Method for Single Image}\label{sec:single_image}
In this section, we present a algorithm for fast and flexible camera calibration using only a single collimator image. 

\subsection{Single-Image Calibration Algorithm}
\begin{figure}[htbp]
	\centering
	\includegraphics[width=\linewidth]{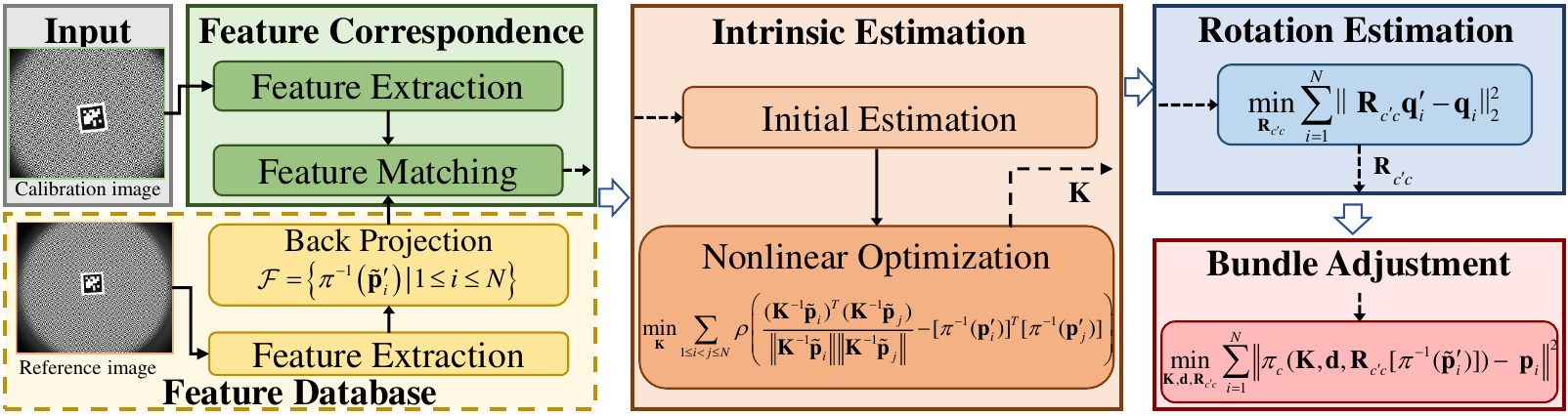}
	\caption{Flowchart of the single-image calibration algorithm. {The input includes only a single calibration image. A pre-constructed feature database is built using reference images, which captures the inherent angular information of our collimator system. }The core constraint of our algorithm stems from the angle invariance of collimator system. The entire process involves consecutively solving three optimization problems.}
	\label{fig:flowchart}
\end{figure}

Since this algorithm utilizes only a single collimator image for calibration, the spherical motion constraint cannot be applied. Our method employs angle invariance as the core constraint for estimating camera parameters. Specifically, {we begin by creating a feature database for the collimator, which is traditionally achieved using goniometric instruments \citep{Sagawa2008}}. However, the complexity of using goniometers increases due to the high density of points in the calibration pattern. Instead, we propose a more efficient method: {constructing the feature database using a collimator image with known parameters.} This image, known as the reference image, can be obtained by capturing a collimator image using a camera with known parameters. {The process of constructing the feature database is shown in the dashed box in Fig.~\ref{fig:flowchart}. For the reference image, feature extraction is performed first, followed by back projection using the known parameters to obtain the feature set $\mathcal{F} = \left\{ {{\pi ^{ - 1}}\left( {{{{\bf{\tilde p'}}}_i}} \right)|1 \le i \le N} \right\}$.} Subsequently, the camera to be calibrated captures its own collimator image, known as the calibration image. The parameters of the target camera are then determined based on the angle invariance constraint. Importantly, the reference image needs to be captured only once, allowing for the calibration of all subsequent cameras, even those with differing parameters. This reusability stems from the primary purpose of the reference image in providing inherent and invariant prior angular information of the collimator system. Therefore, we can define this algorithm as a single-image calibration approach.

{The algorithm flowchart is illustrated in Fig.~\ref{fig:flowchart}. The input of our algorithm includes only the calibration image, and the output is the camera parameters.} Our algorithm consists of four key steps: feature extraction and matching, independent estimation of the intrinsic matrix based on angle invariance constraint, optimal estimation of the rotation, and final optimization using bundle adjustment. {Note that the steps for constructing the feature database are pre-processing steps that only need to be executed once, and are therefore represented by dashed box. In the first stage, \cite{Thomas2020} offers a feature extraction algorithm for star-based pattern and assign a unique identifier for each feature. Then, we can establish a one-to-one correspondence between the features of the calibration image and those in the database.} Given the feature correspondences, our algorithm solves three optimization problems to determine the camera parameters of the calibration image. The formulation and solution for these optimization problems are detailed below.

\subsubsection{Intrinsic Estimation}
In Sect.~\ref{sec:Motion Model}, we show the angle invariance constraint of the collimator system. Specifically, the angle between any pair of points on the reticle of the collimator system remains invariant when imaged by cameras with arbitrary poses. This pose-independent property allows us to derive constraints only related to the camera's intrinsic matrix. Let us select two corresponding pairs of points on the calibration image and the reference image, denoted as $(\mathbf{p}_i, \mathbf{p}_j)$ and $(\mathbf{p}_i^{\prime}, \mathbf{p}_j^{\prime})$, respectively. The angle invariance allows us to construct the following equation:
\begin{equation}
	\frac{(\mathbf{K}^{-1}  \widetilde{\mathbf{p}}_i)^T (\mathbf{K}^{-1}  \widetilde{\mathbf{p}}_j)}{\left\| \mathbf{K}^{-1}  \widetilde{\mathbf{p}}_i \right\| \left\| \mathbf{K}^{-1}  \widetilde{\mathbf{p}}_j \right\|} = [\pi^{-1}( \widetilde{\mathbf{p}}_i^{\prime})]^T[\pi^{-1}( \widetilde{\mathbf{p}}_j^{\prime})],
	\label{eq:KpKp}
\end{equation}
where $\widetilde{\mathbf{p}}$ is the homogeneous form of $\mathbf{p}$, $\pi^{-1}(\cdot)$ denotes the camera's back-projection function and $\pi^{-1}( \widetilde{\mathbf{p}}_i^{\prime})$ calculates the unit direction vector of the ray corresponding to the reference image point $\widetilde{\mathbf{p}}_i^{\prime}$. The right of Eq.~\eqref{eq:KpKp} is an angular reference constructed using the reference image. Given $N$ corresponding points between the calibration and reference image, we define a cost function for $\mathbf{K}$ as:
\begin{equation}
	E_{\mathbf{K}} = \sum\limits_{1\leq i < j \leq N } {\frac{(\mathbf{K}^{-1}  \widetilde{\mathbf{p}}_i)^T (\mathbf{K}^{-1}  \widetilde{\mathbf{p}}_j)}{\left\| \mathbf{K}^{-1}  \widetilde{\mathbf{p}}_i \right\| \left\| \mathbf{K}^{-1}  \widetilde{\mathbf{p}}_j \right\|} - [\pi^{-1}( \widetilde{\mathbf{p}}_i^{\prime})]^T[\pi^{-1}( \widetilde{\mathbf{p}}_j^{\prime})]}.
	\label{eq:EK}
\end{equation}
Notably, the unknown in Eq.~\eqref{eq:EK} is only the intrinsic matrix $\mathbf{K}$, while the camera pose is omitted. This independent estimation of $\mathbf{K}$ eliminates the influence of extrinsic parameters, thereby minimizing the intrinsic parameter estimation error. 

Before performing nonlinear optimization, an initial estimate of $\mathbf{K}$ is required. We adopt the following reasonable prior assumptions: equal focal lengths in the $x$ and $y$ directions ($f_x = f_y =f$), zero skew ($\gamma = 0$), and a principal point located at the image center {$(c_x = W/2, c_y = H/2)$}, where $W$ and $H$ are the width and height of calibration image, respectively. By enforcing the constraint $E_{\mathbf{K}} = 0$, we obtain a univariate quartic equation:
\begin{equation}
	\sum\limits_{1\leq i < j \leq N } a_{ij} \left(\frac{1}{f}\right)^4 + b_{ij} \left(\frac{1}{f}\right)^2 + c_{ij} = 0,
\end{equation}
where the coefficients $a_{ij}$, $b_{ij}$ and $c_{ij}$ can be calculated from the calibration image points $(\widetilde{\mathbf{p}}_i, \widetilde{\mathbf{p}}_j)$, and the back-projected unit direction vectors $\pi^{-1}( \widetilde{\mathbf{p}}_i^{\prime})$ and $\pi^{-1}( \widetilde{\mathbf{p}}_j^{\prime})$ from the reference image. This quartic equation has a specific form; a simplified analytical solution method can be used to obtain an initial estimate of $f$. Combined with the prior assumptions, this yields an initial intrinsic parameter matrix $\mathbf{K}$ for subsequent optimization. The final optimization problem is defined as:
\begin{equation}
	\mathop {\min }\limits_\mathbf{K} \sum\limits_{1\leq i < j \leq N }{\rho \left( \frac{(\mathbf{K}^{-1}  \widetilde{\mathbf{p}}_i)^T (\mathbf{K}^{-1}  \widetilde{\mathbf{p}}_j)}{\left\| \mathbf{K}^{-1}  \widetilde{\mathbf{p}}_i \right\| \left\| \mathbf{K}^{-1}  \widetilde{\mathbf{p}}_j \right\|} - [\pi^{-1}( \widetilde{\mathbf{p}}_i^{\prime})]^T[\pi^{-1}( \widetilde{\mathbf{p}}_j^{\prime})] \right)}.
	\label{eq:min_K}
\end{equation}
The optimization process iteratively refines $\mathbf{K}$ to achieve the highest possible accuracy in satisfying angle invariance. {The Cauchy loss function $\rho(\cdot)$ is employed to suppress outliers that may arise from unmodeled distortions.} In fact, the optimization problem Eq.~\eqref{eq:min_K} does not include image distortion. Because distortion correction during the back-projection process requires iterative computation. Incorporating this step directly into the nonlinear optimization process would not only significantly increase computational time but also lead to numerical instability. Therefore, we handle distortion correction during the forward projection stage. 

\subsubsection{Rotation Estimation}
Based on the spherical motion model described in Eq.~\eqref{eq:Tpc}, we can infer that the transformation between the reference and calibration images is a pure rotation. Let $\mathbf{T}_{pc}$ and $\mathbf{T}_{pc^{\prime}}$ represent the relative transformation matrices from the target to the calibration image and from the target to the reference image, respectively. Consequently, {the relative transformation matrix between the calibration and reference images}, denoted as $\mathbf{T}_{c^{\prime}c}$, can be expressed as: 
\begin{equation}
	\mathbf{T}_{c^{\prime}c} = \left[ {\begin{array}{*{20}{c}}
			{{\mathbf{R}_{pc}}}&{ - {\mathbf{R}_{pc}}{\mathbf{t}_{cp}}}\\
			{\mathbf{0}_{1 \times 3}}&1
	\end{array}} \right] \left[ {\begin{array}{*{20}{c}}
			{{\mathbf{R}_{pc^{\prime}}}}&{ - {\mathbf{R}_{pc^{\prime}}}{\mathbf{t}_{cp}}}\\
			{\mathbf{0}_{1 \times 3}}&1
	\end{array}} \right]^{-1}
	= \left[ {\begin{array}{*{20}{c}}
			{{\mathbf{R}_{pc}}{\mathbf{R}_{pc^{\prime}}^T}}&{\mathbf{0}_{3 \times 1}}\\
			{\mathbf{0}_{1 \times 3}}&1
	\end{array}} \right].
\end{equation}

{The corresponding points $(\mathbf{p}_i, \mathbf{p}_i^{\prime})$ on the calibration and reference images can be related by a rotation matrix as:}
\begin{equation}
	\frac{\mathbf{K}^{-1}  \widetilde{\mathbf{p}}_i}{\| \mathbf{K}^{-1}  \widetilde{\mathbf{p}}_i \|} = \mathbf{R}_{c^{\prime}c} [\pi^{-1}( \widetilde{\mathbf{p}}_i^{\prime})],
	\label{eq:KpRp}
\end{equation}
where $\mathbf{R}_{c^{\prime}c} = {\mathbf{R}_{pc}}{\mathbf{R}_{pc^{\prime}}^T}$. Let $\mathbf{q}_i = \frac{\mathbf{K}^{-1}  \widetilde{\mathbf{p}}_i}{\| \mathbf{K}^{-1}  \widetilde{\mathbf{p}}_i \|}$ and $\mathbf{q}_i^{\prime} = \pi^{-1}( \widetilde{\mathbf{p}}_i^{\prime})$, an optimization problem related to rotation is defined as follows:
\begin{equation}
	\mathop {\min }\limits_\mathbf{\mathbf{R}_{c^{\prime}c}} \sum\limits_{i=1}^N  {\| \mathbf{R}_{c^{\prime}c} \mathbf{q}_i^{\prime} - \mathbf{q}_i \|_2^2}.
	\label{eq:min_Rcc}
\end{equation}
This is a typical vector alignment problem, which can be efficiently solved using the Kabsch algorithm \citep{Kabsch1978discussion}. First, the point sets $\mathbf{q}_i$ and $\mathbf{q}_i^{\prime}$ are centered by subtracting their respective centroids: $\overline{\mathbf{q}}_i = \mathbf{q}_i - \frac{1}{N}\sum_{i=1}^N \mathbf{q}_i$ and $\overline{\mathbf{q}}_i^{\prime} = \mathbf{q}_i^{\prime} - \frac{1}{N}\sum_{i=1}^N \mathbf{q}_i^{\prime}$. Next, the covariance matrix is computed as $\mathbf{A} = \sum_{i=1}^N \overline{\mathbf{q}}_i^{\prime} \overline{\mathbf{q}}_i^T$. A Singular Value Decomposition (SVD) is then applied to $\mathbf{A}$, yielding $\mathbf{A} = \mathbf{U} \Sigma \mathbf{V}^T$. To ensure that the rotation matrix $\mathbf{R}_{c^{\prime}c}$ maintains a right-handed coordinate system, a correction factor is computed as $d = \operatorname{sign}(\operatorname{det}(\mathbf{V}\mathbf{U}^T))$. Finally, the optimal rotation matrix is computed as $\mathbf{R}_{c^{\prime}c} = \mathbf{U} \operatorname{diag}(1, 1, d) \mathbf{V}^T$. 

\subsubsection{Bundle Adjustment}
There exists a well-defined mapping between the calibration image and the reference image, which is determined by their camera parameters and relative rotation. During the intrinsic parameter estimation stage, distortion coefficients are temporarily excluded from the constraint. Now, we consider distortion coefficients in the forward projection process of the mapping. The complete mapping can be mathematically represented as:
\begin{equation}
	\mathbf{p}_i  = \pi_c(\mathbf{K}, \mathbf{d}, \mathbf{R}_{c^{\prime}c} [\pi^{-1}( \widetilde{\mathbf{p}}_i^{\prime})]),
	\label{eq:piKpRp}
\end{equation}
where $\pi_c(\cdot)$ denotes the forward projection function of the calibration camera. To further refine the camera parameters, we jointly optimize the intrinsic matrix $\mathbf{K}$, the distortion coefficients $\mathbf{d}$ and the relative rotation $\mathbf{R}_{c^{\prime}c}$ by minimizing the re-projection error. {The optimization cost function is defined as:}
\begin{equation}
	\mathop {\min }\limits_{\mathbf{K}, \mathbf{d}, \mathbf{R}_{c^{\prime}c}} \sum\limits_{i=1}^N {{{{\left\| \pi_c(\mathbf{K}, \mathbf{d}, \mathbf{R}_{c^{\prime}c} [\pi^{-1}( \widetilde{\mathbf{p}}_i^{\prime})]) - \mathbf{p}_i \right\|}^2}} }.
	\label{eq:min_KdR}
\end{equation}
The equation first back-projects points from the reference image into 3D space, and then projects them onto the calibration image using the relative rotation and camera parameters. Eq.~\eqref{eq:min_KdR} can be iteratively solved using the Levenberg-Marquardt (LM) method. The initial values for the intrinsic matrix $\mathbf{K}$ and the relative rotation $\mathbf{R}_{c^{\prime}c}$ are obtained from the preceding calculations, while the initial value of the distortion coefficients $\mathbf{d}$ can be set to $\mathbf{0}$.

In summary, the calibration process for single image can be completed by solving three optimization problems (i.e. Eq.~\eqref{eq:min_K}, Eq.~\eqref{eq:min_Rcc} and Eq.~\eqref{eq:min_KdR}). The foundation of this process is a reference image, which serves as a consistent benchmark for the entire calibration. Even cameras with varying parameters can establish constraints based on a common benchmark. {A significant advantage of our method is that the reference collimator image needs to be captured only once and can be reused in subsequent calibration processes.} The single-image calibration method presents a novel scheme for batch and rapid camera calibration, particularly suitable for applications that require the calibration of a large number of cameras within a short time frame, such as automated production lines and large-scale camera deployments.
	
\section{Evaluation and Results}
\label{sec:Experiment}
In this section, a series of experiments are performed to support the proposed algorithm and evaluate the effectiveness of calibration using a collimator system. We first evaluate the robustness and accuracy of the proposed algorithm with synthetic data. Subsequently, the spherical motion model between the calibration target and the camera is validated using real collimator images. Most importantly, we calibrate the camera with real collimator images and confirm the validity of the proposed method.

\subsection{Synthetic Data Experiments}
{In the synthetic data experiments}, we evaluate the performance of the proposed algorithm. We simulate the process of calibrating predefined camera parameters using synthetic images that conform to the spherical motion model. This controlled environment allows for evaluation of the algorithm's accuracy and robustness. {The proposed algorithm is compared with the following well-established algorithms:} \\
$ \bullet$ \texttt{Zhang} \citep{Zhang2000} is a widely used plane-based calibration algorithm suitable for general 6DOF motion. \\
$ \bullet$ \texttt{Bouguet} \citep{Bouguet2004} is a mainstream calibration toolbox and also suitable for general 6DOF motion. A difference from \texttt{Zhang} is that \texttt{Bouguet} leverages the orthogonal vanishing points constraint to estimate the focal length.\\
$ \bullet$ \texttt{Hartley} \citep{Hartley1997} is a self-calibration method specifically designed for pure rotating camera, suitable for the 3DOF spherical motion model.

{The synthetic data generation process involves defining a virtual camera with the following parameters:} focal lengths $f_x = f_y = 1000$ pixels, principal point coordinates $c_x = 542$ pixels and $c_y = 478$ pixels, skew factor $\gamma = 0.01$, radial distortion coefficients $d_1 = 0.1, d_2 = -0.2$. The image size is set to $1080 \times 960$ pixels. The calibration target is a planar pattern containing $11 \times 8$ points, each square size of $30 mm \times 30 mm$. The target maintains a spherical motion with a radius of 700$mm$ relative to the camera. The optical center of the camera, expressed in the target coordinate system, is fixed at $\mathbf{t}_{cp} = [150, 105, -700]^T mm$, but the camera's orientation is arbitrary. To generate the synthetic images, {the planar target points are projected onto the image plane based on the predefined camera parameters and poses.}

\subsubsection{Evaluation of Calibration Performance.} 
\begin{figure}[tbp]
	\centering
	\includegraphics[width=0.245\linewidth]{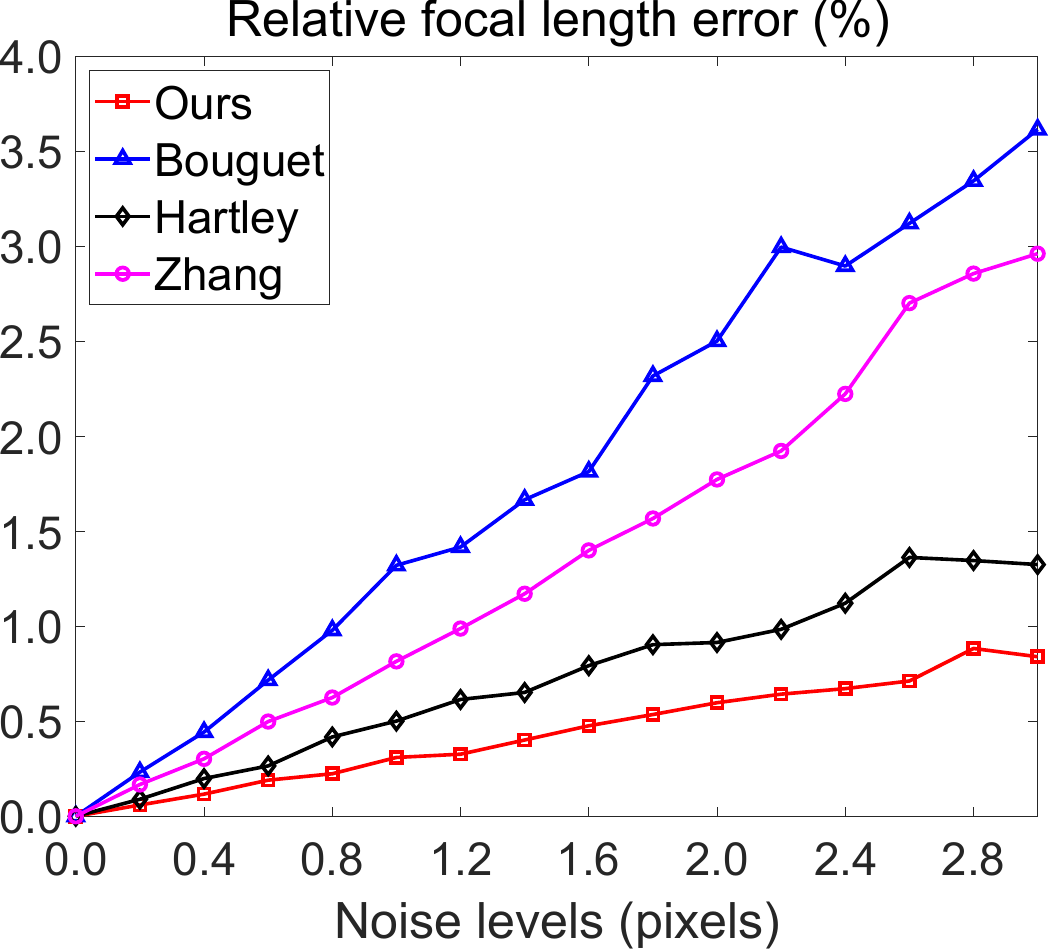} 
	\includegraphics[width=0.235\linewidth]{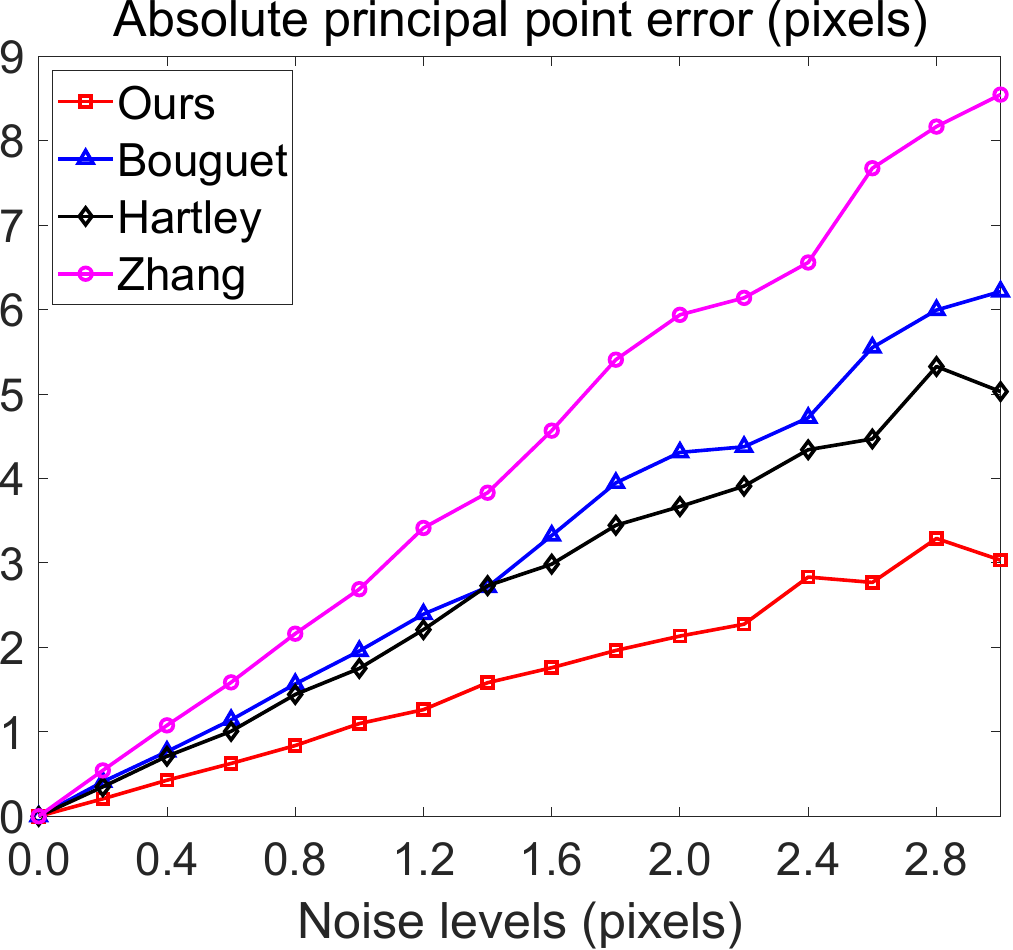} 
	\includegraphics[width=0.245\linewidth]{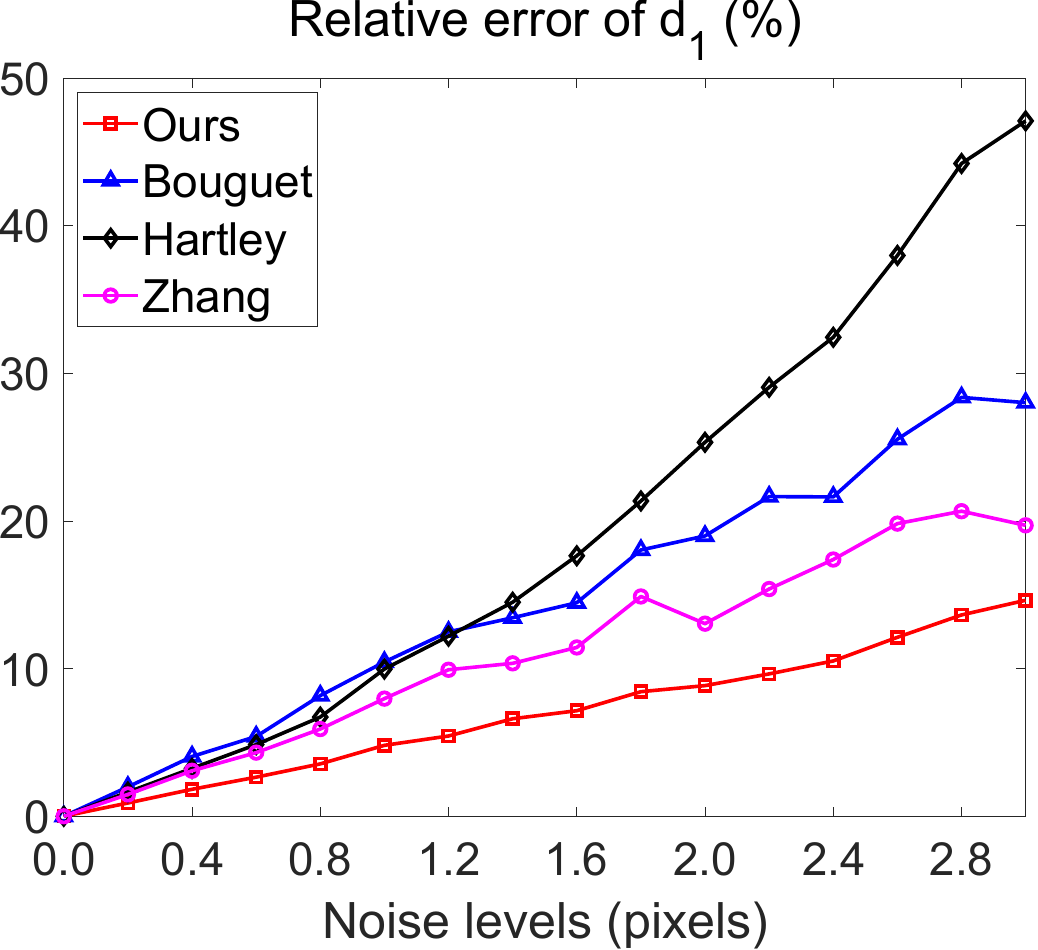}
	\includegraphics[width=0.245\linewidth]{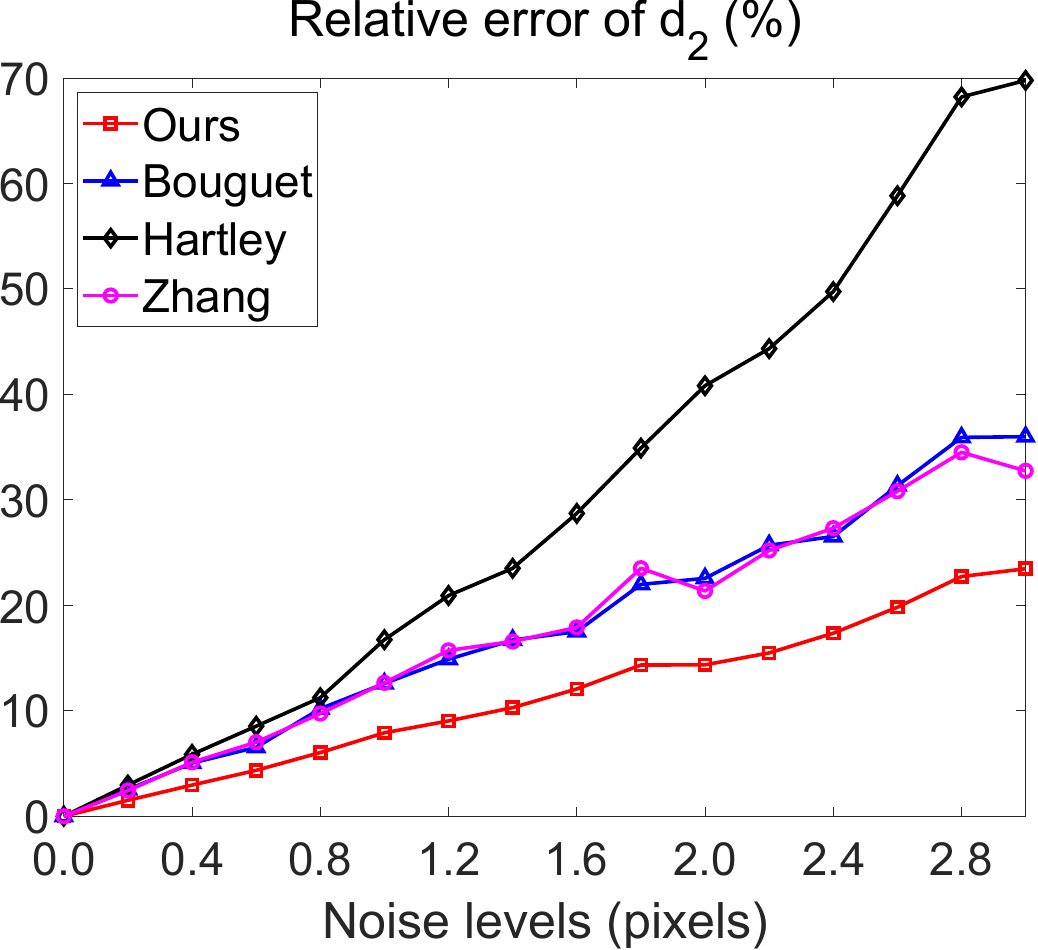}
	\caption{{Comparison of calibration errors as noise level increases. The predefined virtual camera intrinsics are: $(f_x, f_y) = (1000, 1000)$ pixels, $(c_x, c_y)= (542, 478)$ pixels, $d_1 = 0.1$ and $d_2 = -0.2$. Our method consistently outperforms other methods across all noise levels.}}
	\label{fig:Noise}
\end{figure}

The primary goal of this experiment is to evaluate the performance of the proposed algorithm in estimating camera intrinsic parameters and compare it with three widely used baseline algorithms. To ensure a fair comparison, all algorithms employ a standard two-stage process: first, initial parameter estimates are obtained using an initialization method; then, the parameters are optimized using bundle adjustment. First, the robustness of each algorithm to varying levels of image noise is evaluated. Image noise is simulated by adding zero-mean Gaussian noise with a standard deviation ranging from 0 to 3.0 pixels to the synthetic images. For each noise level, we use 15 images for calibration and perform 200 independent Monte Carlo simulations. The average estimation error of parameters is reported. The results, presented in Fig.~\ref{fig:Noise}, demonstrate that the error generally increases with the noise level. Compared with the other three algorithms, the proposed shows greater robustness to noise, particularly at high noise levels. 

\begin{figure}[tbp]
	\centering
	\includegraphics[width=0.245\linewidth]{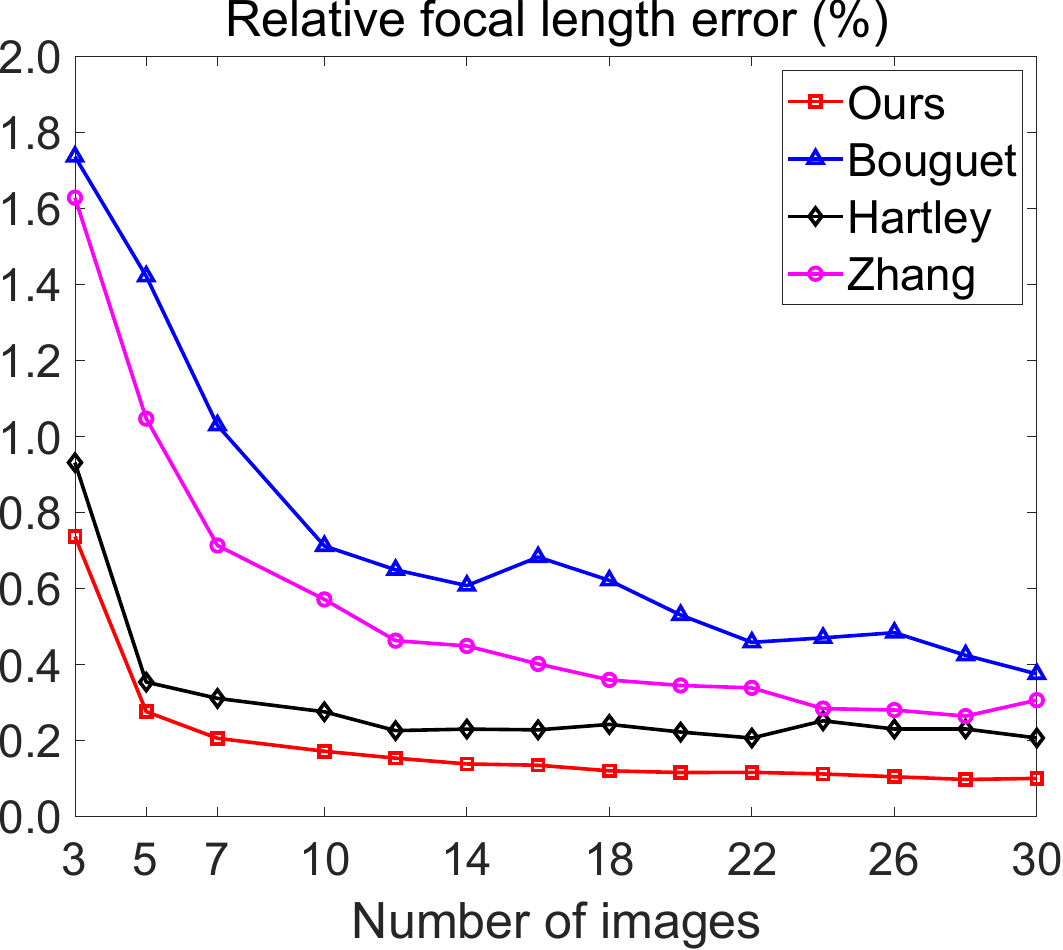}
	\includegraphics[width=0.235\linewidth]{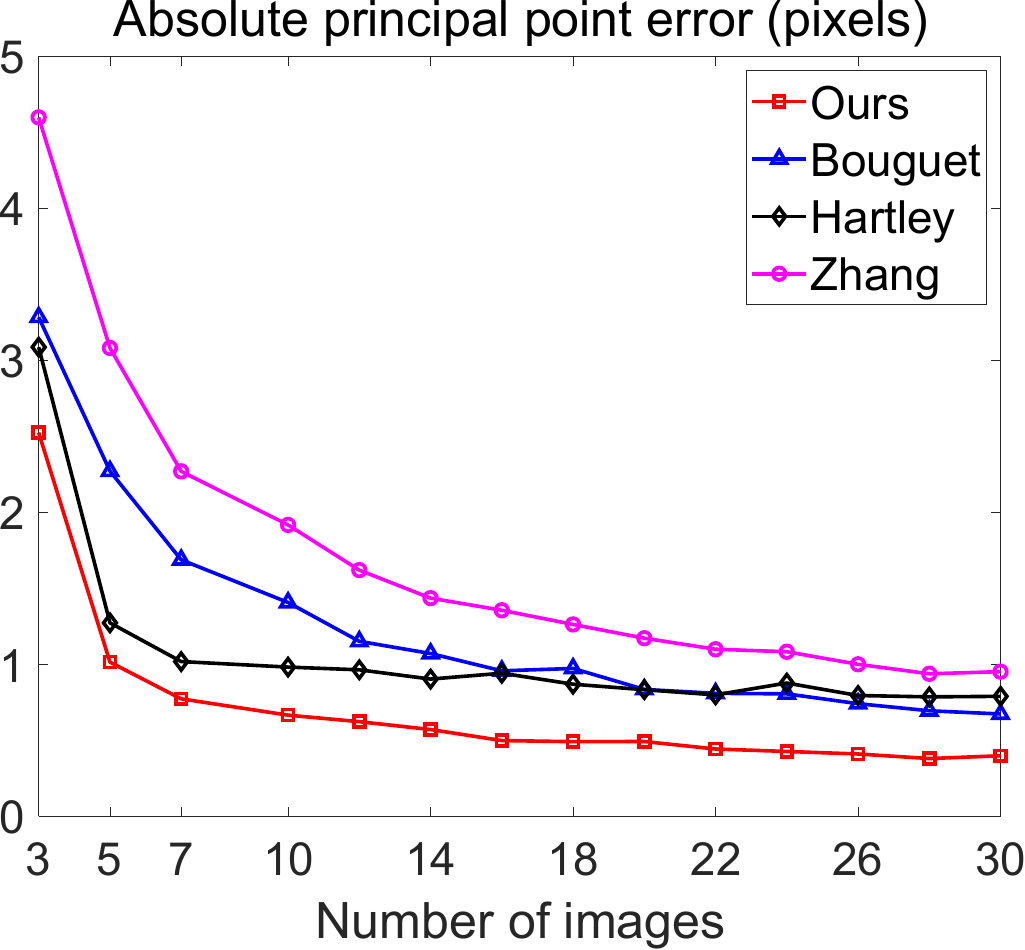} 
	\includegraphics[width=0.245\linewidth]{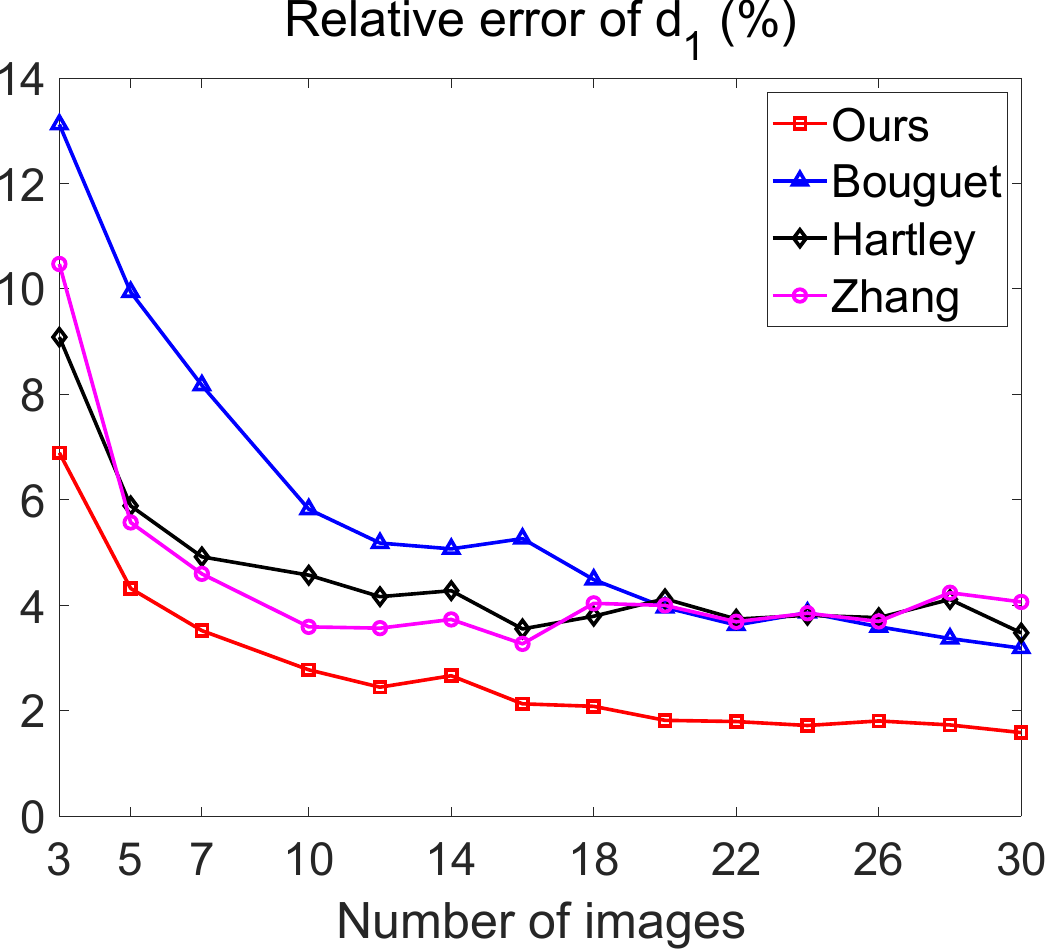}
	\includegraphics[width=0.245\linewidth]{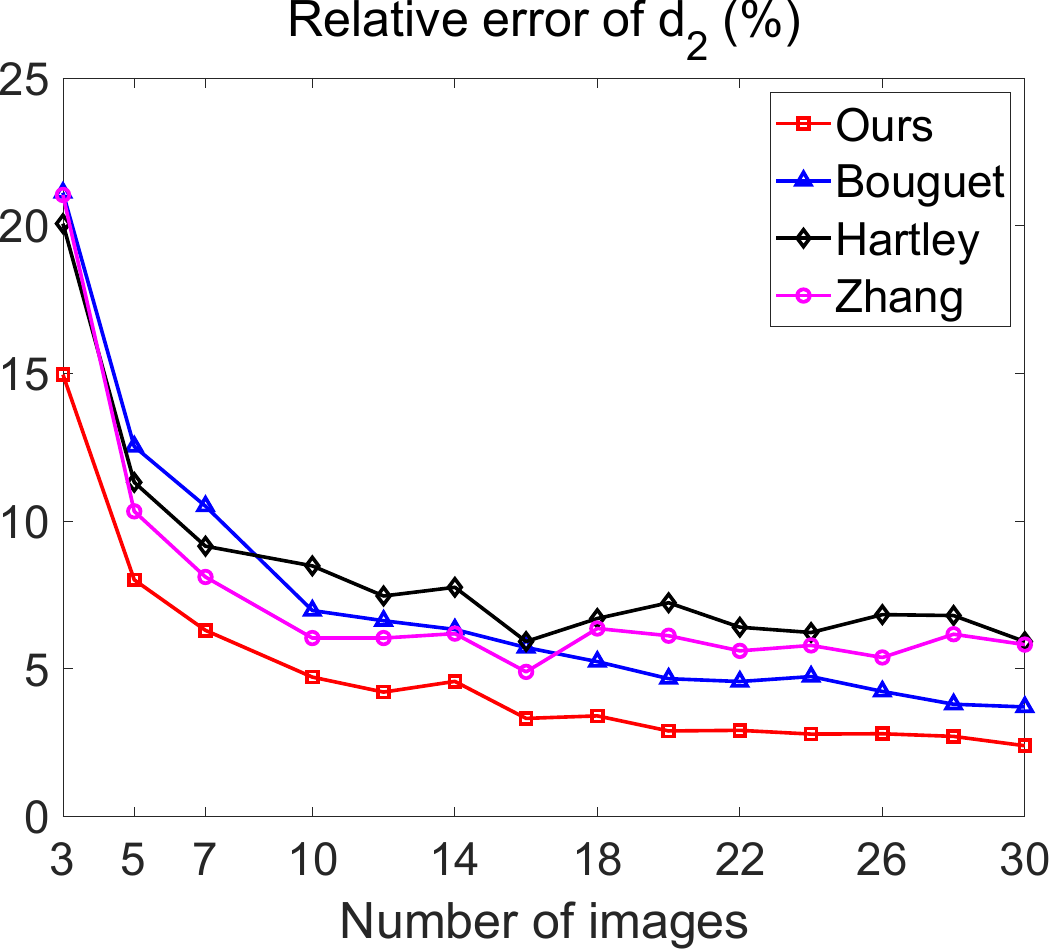}
	\caption{{Comparison of calibration errors as the number of images increases. The predefined virtual camera intrinsics are: $(f_x, f_y) = (1000, 1000)$ pixels, $(c_x, c_y)= (542, 478)$ pixels, $d_1 = 0.1$ and $d_2 = -0.2$. Our method consistently outperforms other methods across all image counts.}}
	\label{fig:nImgs}
\end{figure}

{Then, we evaluate the algorithm's accuracy with varying numbers of calibration images.} In this experiment, the noise level is fixed at 0.5 pixels, and the number of images used for calibration varies from 3 to 30. Similarly, for each number of images, we perform 200 independent Monte Carlo simulations and report the average parameter estimation error. The results are illustrated in Fig.~\ref{fig:nImgs}. A significant decrease in error is observed when increasing the number of images from 3 to 5. The results demonstrate that the proposed algorithm consistently exhibits lower error than other baseline algorithms across different numbers of images, confirming its superiority.


\begin{figure}[tbp]
	\centering
	\includegraphics[width=0.245\linewidth]{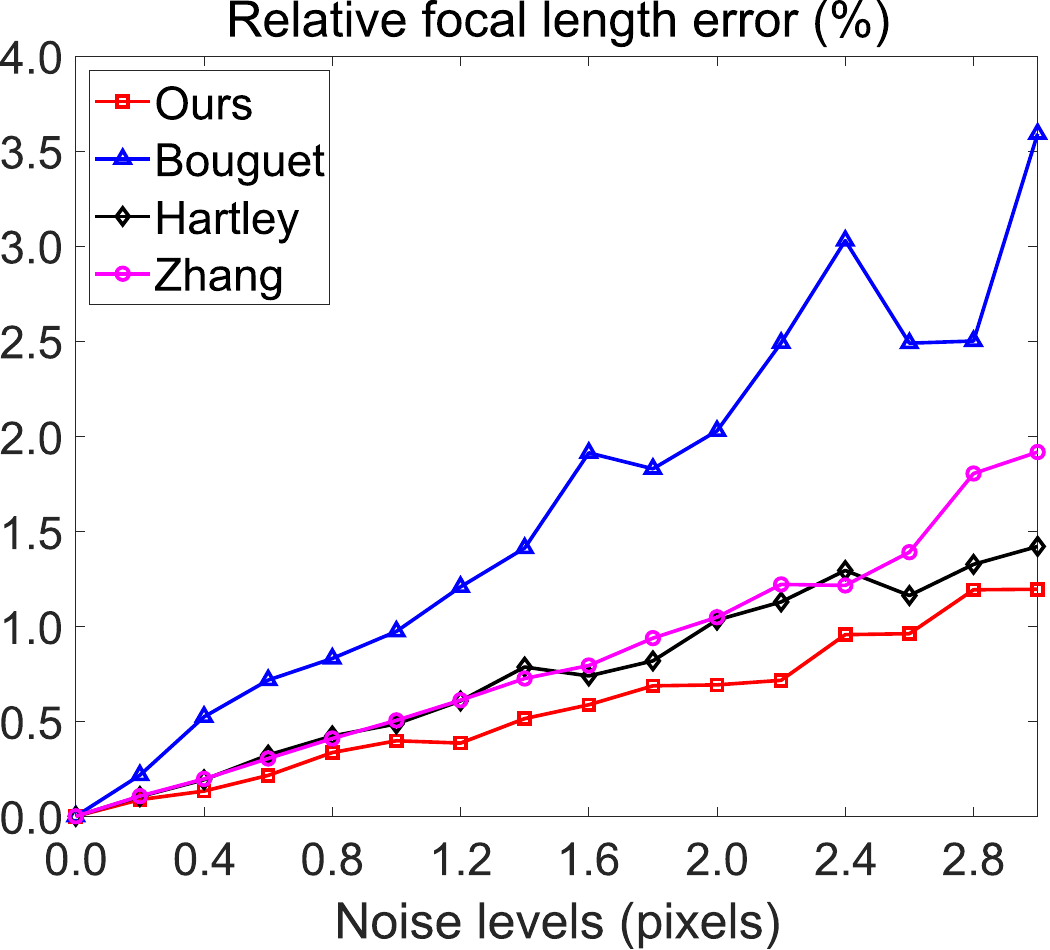} 
	\includegraphics[width=0.235\linewidth]{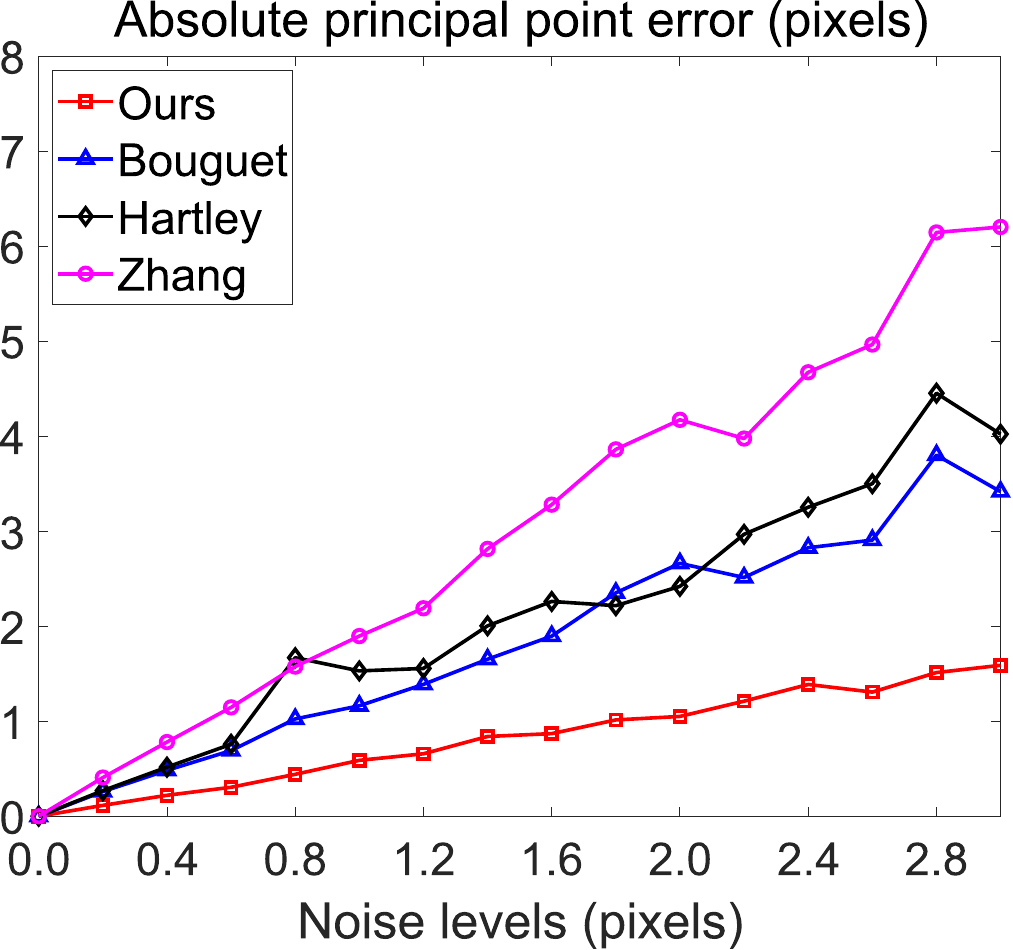} 
	\includegraphics[width=0.245\linewidth]{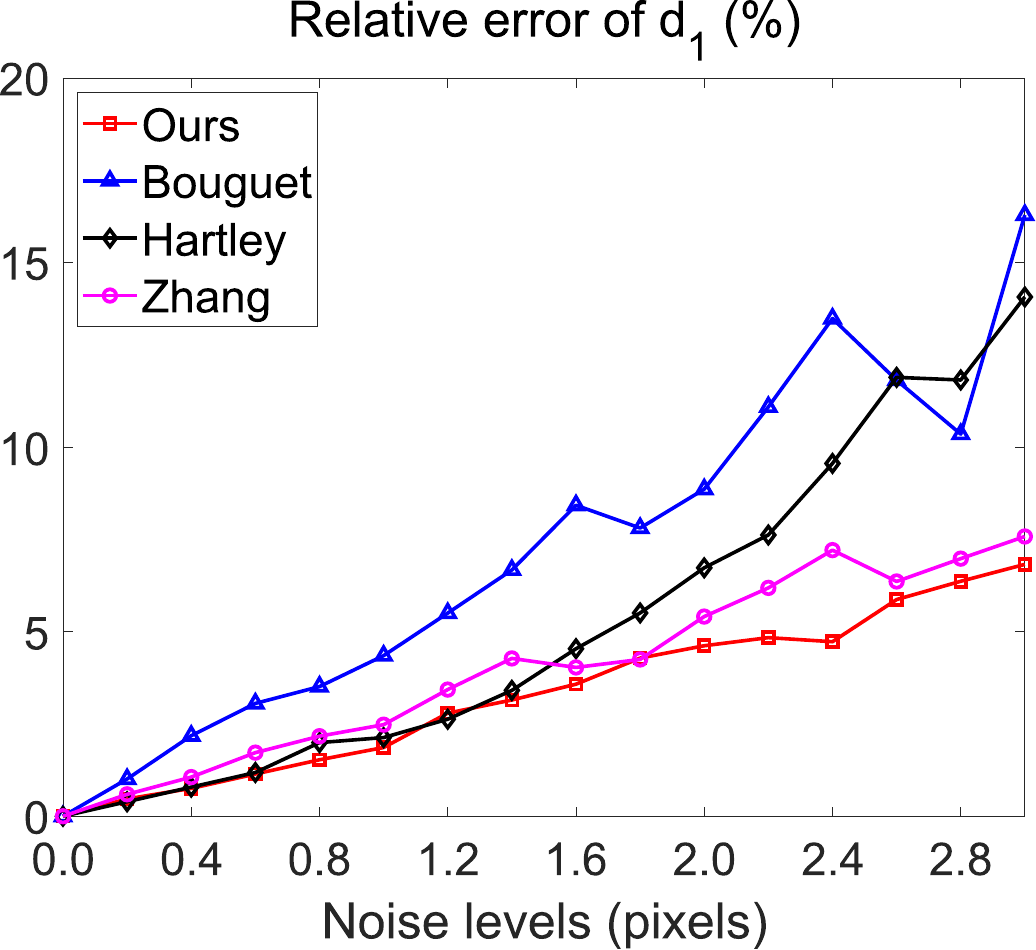}
	\includegraphics[width=0.245\linewidth]{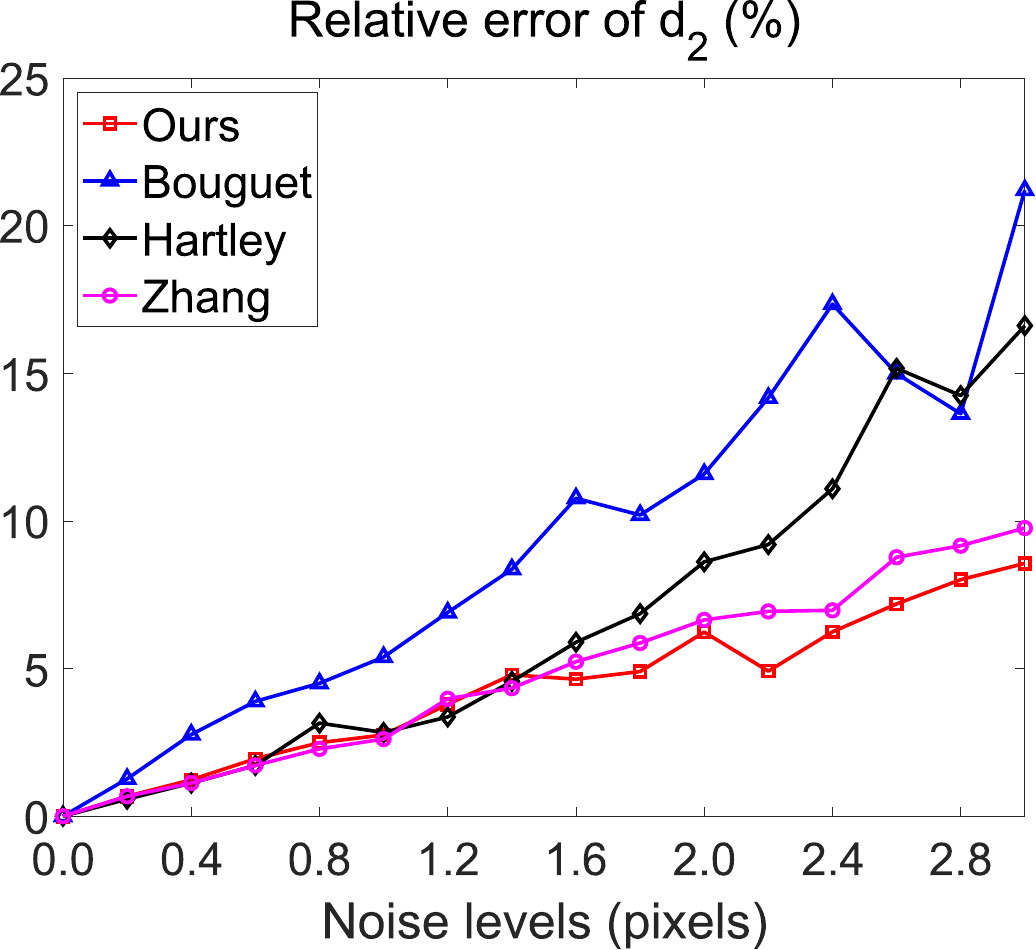}
	\caption{{Comparison of calibration errors as noise level increases under different intrinsics. The predefined virtual camera intrinsics are: $(f_x, f_y) = (800, 800)$ pixels, $(c_x, c_y)= (542, 478)$ pixels, $d_1 = 0.3$ and $d_2 = -0.5$. Compared to other methods, ours demonstrates a lower calibration error, indicating better robustness against noise.}}
	\label{fig:Noise_2}
\end{figure}

\begin{figure}[t]
	\centering
	\includegraphics[width=0.240\linewidth]{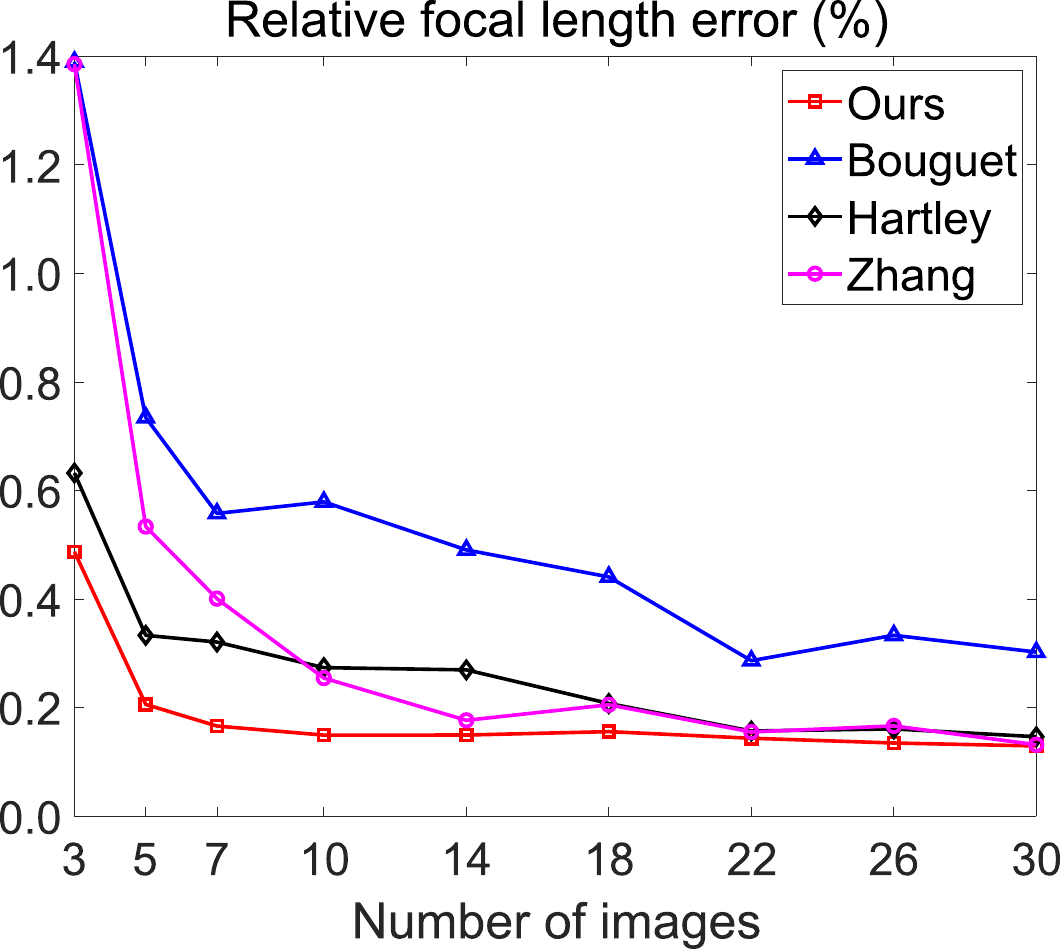}
	\includegraphics[width=0.240\linewidth]{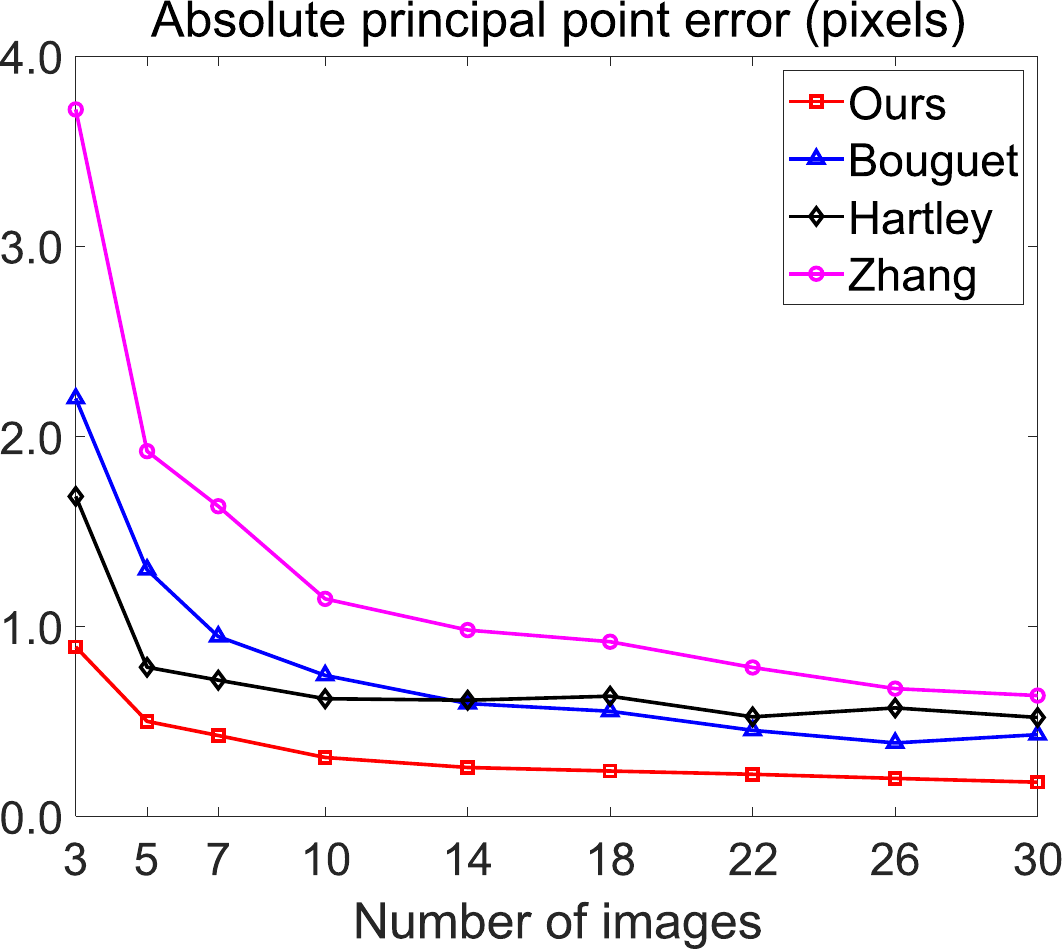} 
	\includegraphics[width=0.240\linewidth]{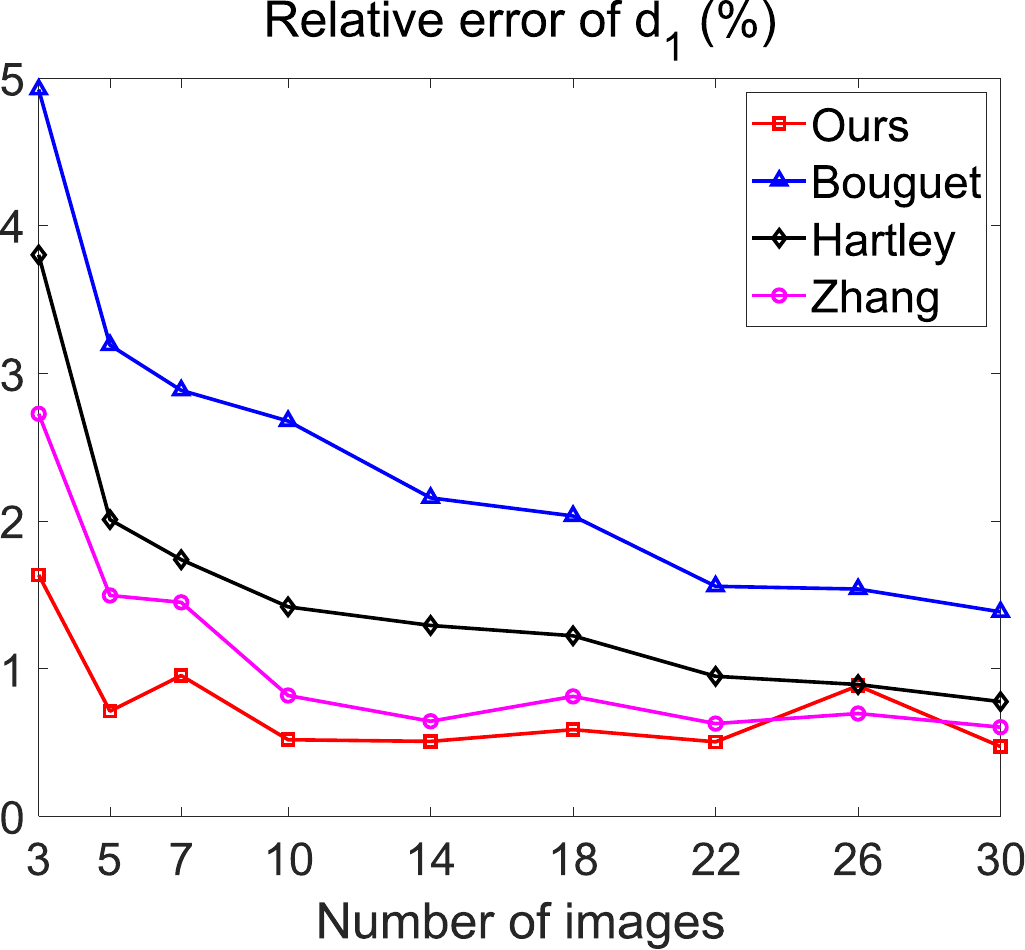}
	\includegraphics[width=0.240\linewidth]{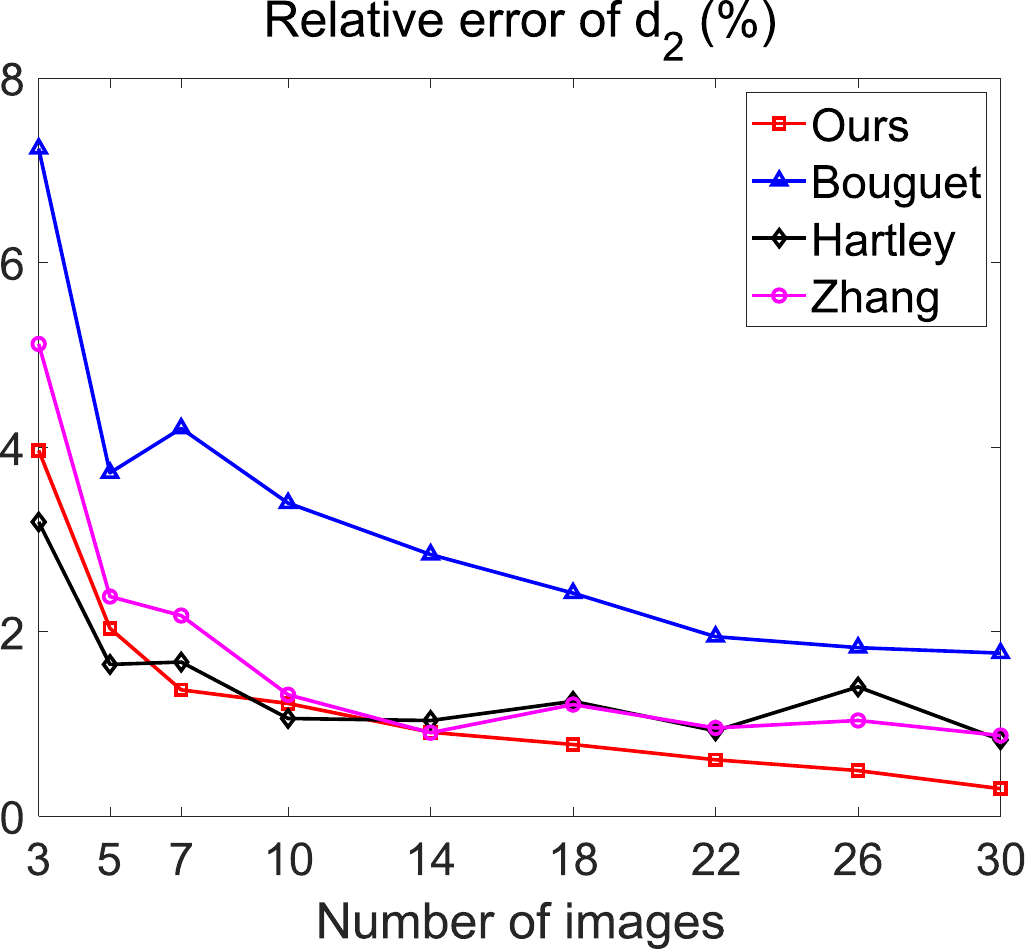}
	\caption{{Comparison of calibration errors with the number of images increases. The predefined virtual camera intrinsics are: $(f_x, f_y) = (800, 800)$ pixels, $(c_x, c_y)= (542, 478)$ pixels, $d_1 = 0.3$ and $d_2 = -0.5$. Compared to other methods, our method demonstrates a lower overall calibration error.}}
	\label{fig:nImgs_2}
\end{figure}

{Additionally, we include calibration performance evaluation results with different camera intrinsic parameters. We set the predefined intrinsics of the virtual camera as follows: $f_x = f_y = 800$ pixels, $c_x = 542$ pixels, $c_y = 478$ pixels, and increase the camera distortion to: $d_1 = 0.3$ and $d_2 = -0.5$. The other experimental settings remain unchanged. We evaluate the algorithm's robustness against different levels of image noise, with the results presented in Fig.~\ref{fig:Noise_2}; simultaneously, we evaluate the algorithm's accuracy with varying numbers of calibration images, with the results shown in Fig.~\ref{fig:nImgs_2}. The results indicate that, compared to the other algorithms, our method demonstrates stability and accuracy across different noise levels and varying numbers of images.}

Overall, {as shown in Figs.~\ref{fig:Noise}-\ref{fig:Noise_2}}, the experimental results consistently demonstrate that the proposed algorithm significantly outperforms the three baseline algorithms in robustness and accuracy. {The improved performance is attributed to \texttt{Ours} simultaneous consideration of spherical motion and planar homography constraints}. In contrast, {both \texttt{Zhang} and \texttt{Bouguet} rely solely on homography constraint}, while the \texttt{Hartley} considers only pure rotational constraint and is less sensitive to distortion. Our method combines these two geometric constraints, effectively reducing optimization variables and thereby achieving more accurate calibration.

\subsubsection{Performance of Initial Value Estimation}
In this experiment, we evaluate the performance of the algorithms in solving the initial value of camera parameters. The linear parameters of the camera include focal length and principal point. Note that the {initial linear parameters} do not include the distortion coefficients, so we set $d_1 = 0, d_2 = 0$. {These linear parameters have not undergone a nonlinear optimization step. The initial values in our proposed method are obtained by solving Eq.~\eqref{eq:Dwa} and performing its subsequent matrix decomposition}.

\begin{figure}[tbp]
	\centering
	\includegraphics[width=0.4\linewidth]{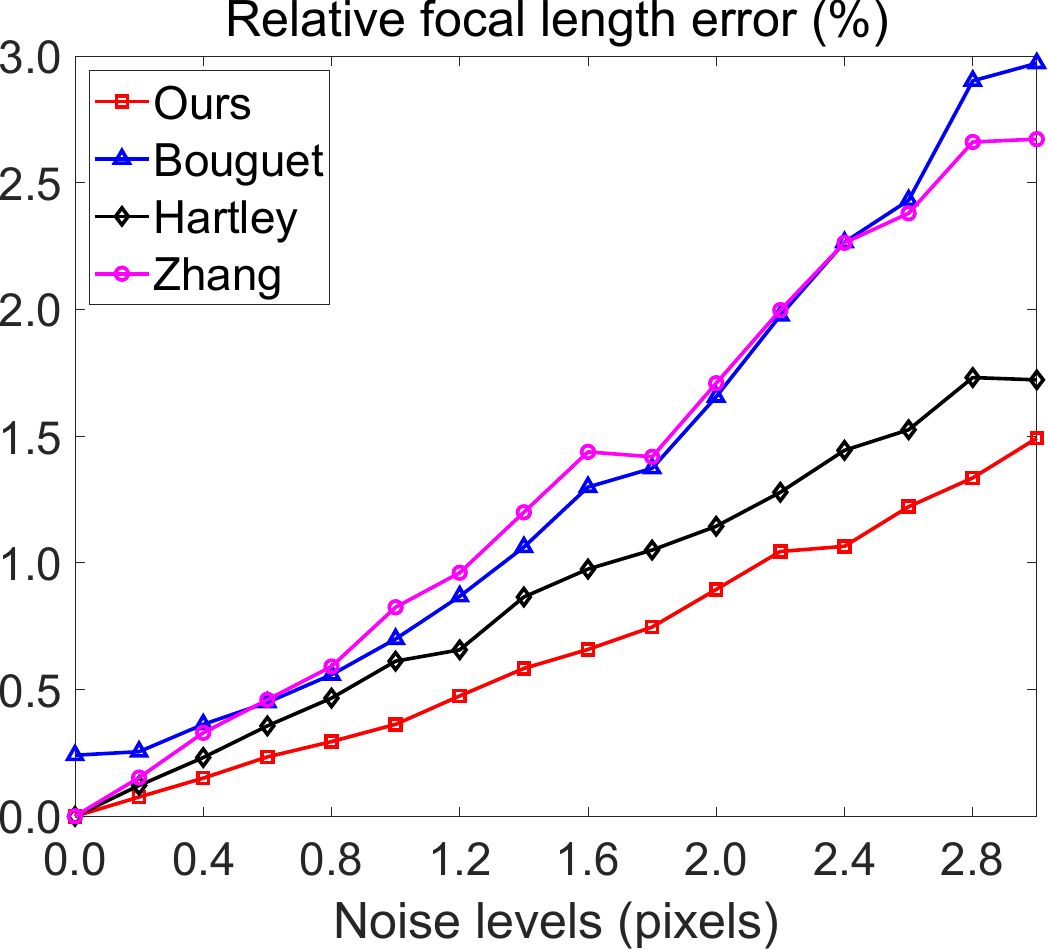}
	\quad
	\includegraphics[width=0.4\linewidth]{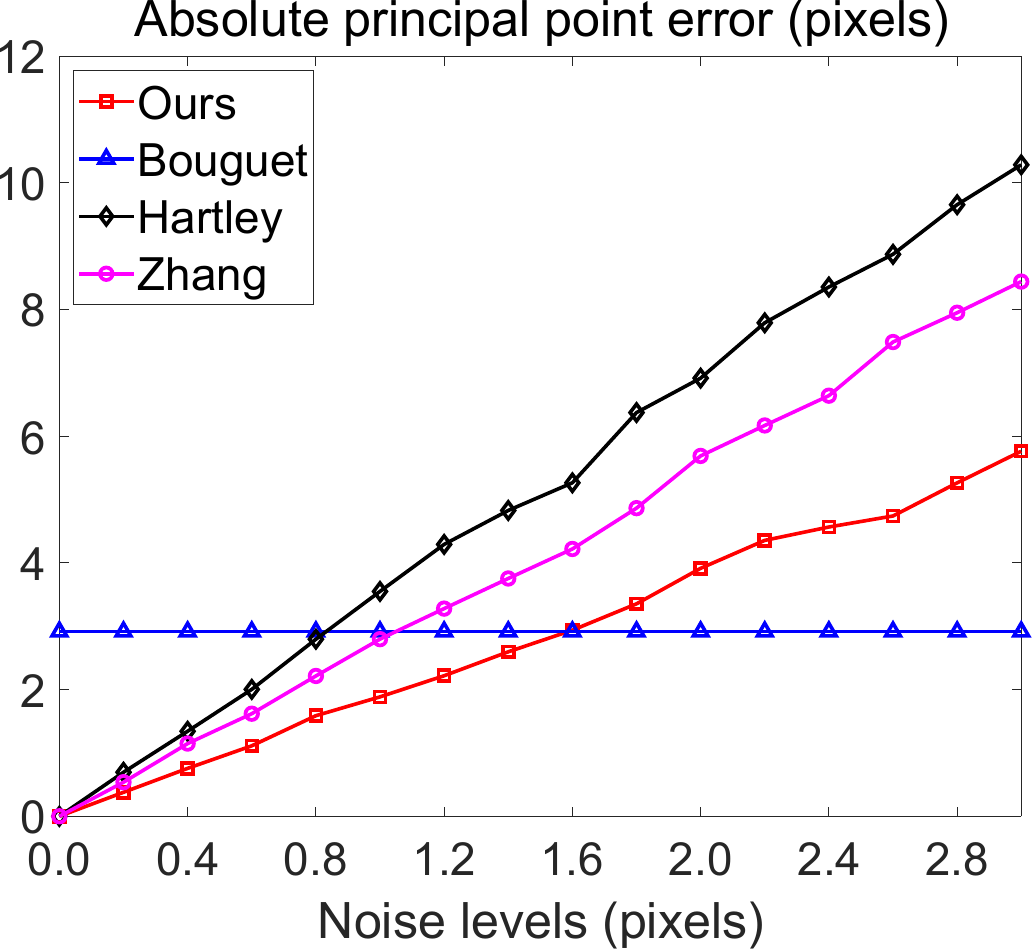}
	\caption{Robustness comparison of initial value estimation with increasing noise level. The proposed algorithm demonstrates greater robustness than other algorithms.}
	\label{fig:Init_Noise}
\end{figure}

Firstly, we evaluate the performance of algorithms with respect to the noise levels. {Gaussian noise with zero-mean and a standard deviation ranging from 0 to 3.0 pixels is added to the image points.} For each noise level, we used 15 images and performed 500 independent trials. As shown in Fig.~\ref{fig:Init_Noise}, the calibration error exhibits a linear increase with the noise level. Our proposed algorithm demonstrates superior robustness against noise compared to the other three. At a noise level of 1 pixel, the initial focal length estimation error is less than $0.5\%$, and the initial principal point estimation error is less than 2.0 pixels. The \texttt{Bouguet} toolbox assumes the principal point to be fixed at the image center during the initial value estimation, which introduces a systematic bias. 
\begin{figure}[tbp]
	\centering
	\includegraphics[width=0.413\linewidth]{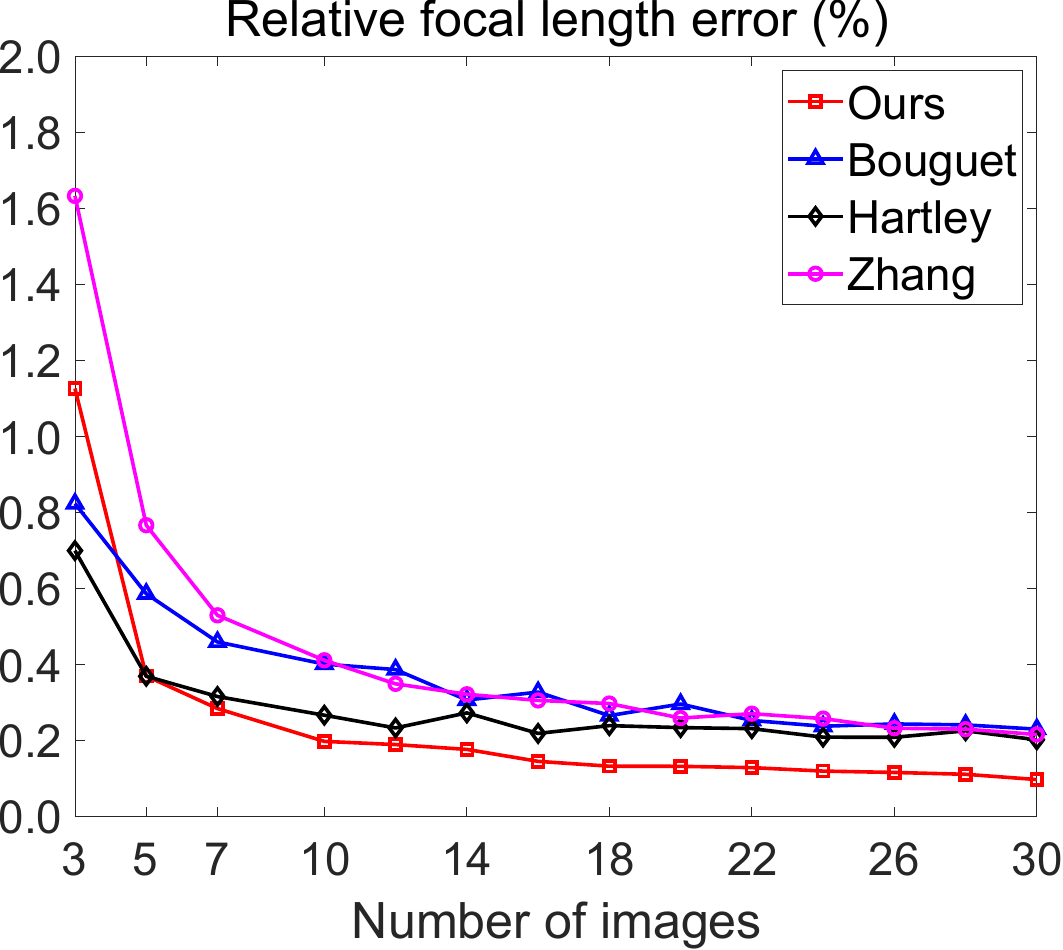}
	\quad
	\includegraphics[width=0.4\linewidth]{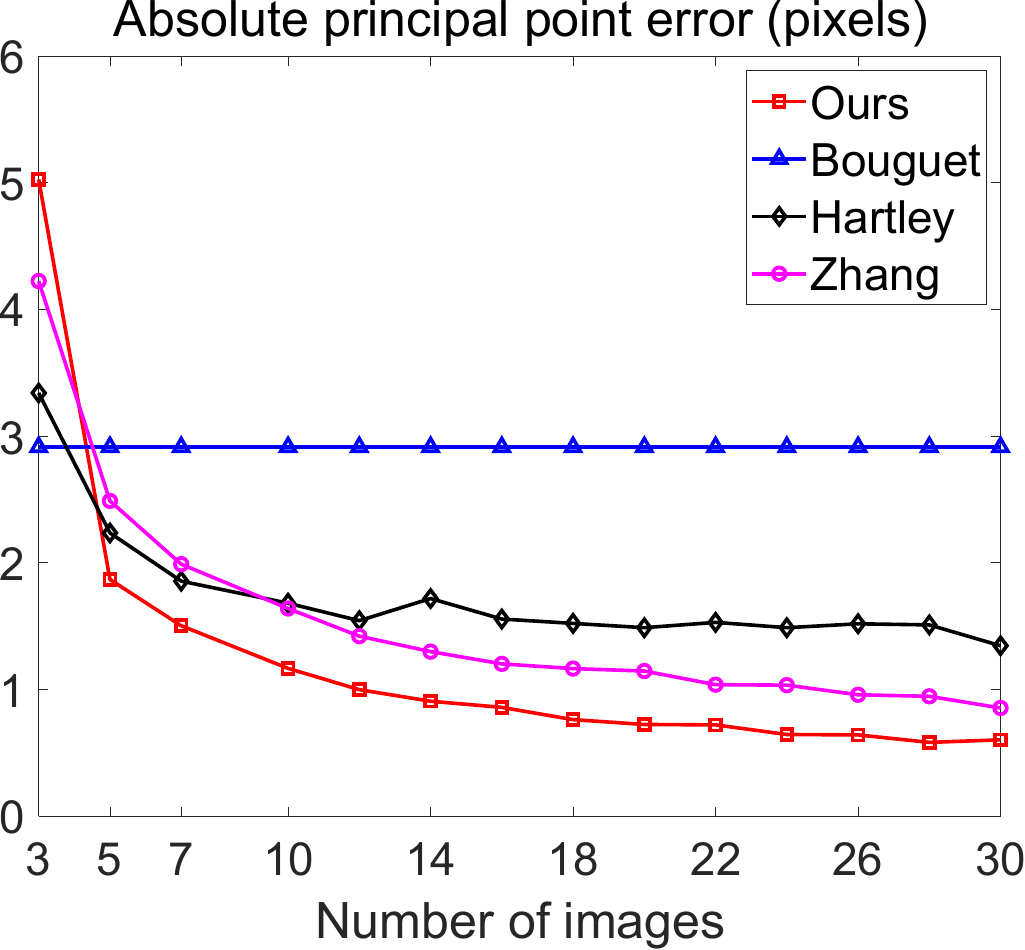}
	\caption{Accuracy comparisons of initial value estimation with the number of images increases. Our algorithm demonstrates superior accuracy compared to other algorithms.}
	\label{fig:Init_nImgs}	
\end{figure}

Secondly, we evaluate the performance of algorithms as the number of calibration images increases. The number of images used for calibration ranges from 3 to 30. Five hundred trials are conducted for each number, and the average results are shown. Fig.~\ref{fig:Init_nImgs} demonstrates that the calibration error decreases as the number of images increases. When using 10 images for calibration, our focal length error is less than $0.2\%$, and the principal point error is around 1.0 pixels. {The proposed algorithm does not demonstrate a significant improvement in accuracy when only three images are used. This is attributed to the fact that the contribution of the spherical motion constraint to accuracy improvement is relatively limited when a small number of images is used.} However, when the number of images exceeds five, \texttt{Ours} consistently outperforms the other algorithms in estimating the focal length and principal point. {The results in Fig.~\ref{fig:Init_Noise} and Fig.~\ref{fig:Init_nImgs} clearly demonstrate the benefits of incorporating spherical motion constraints on solving linear parameters.}

\subsubsection{Sensitivity of Imperfect Spherical Motion.} 
\begin{figure}[tbp]
	\centering
	\includegraphics[width=0.4\linewidth]{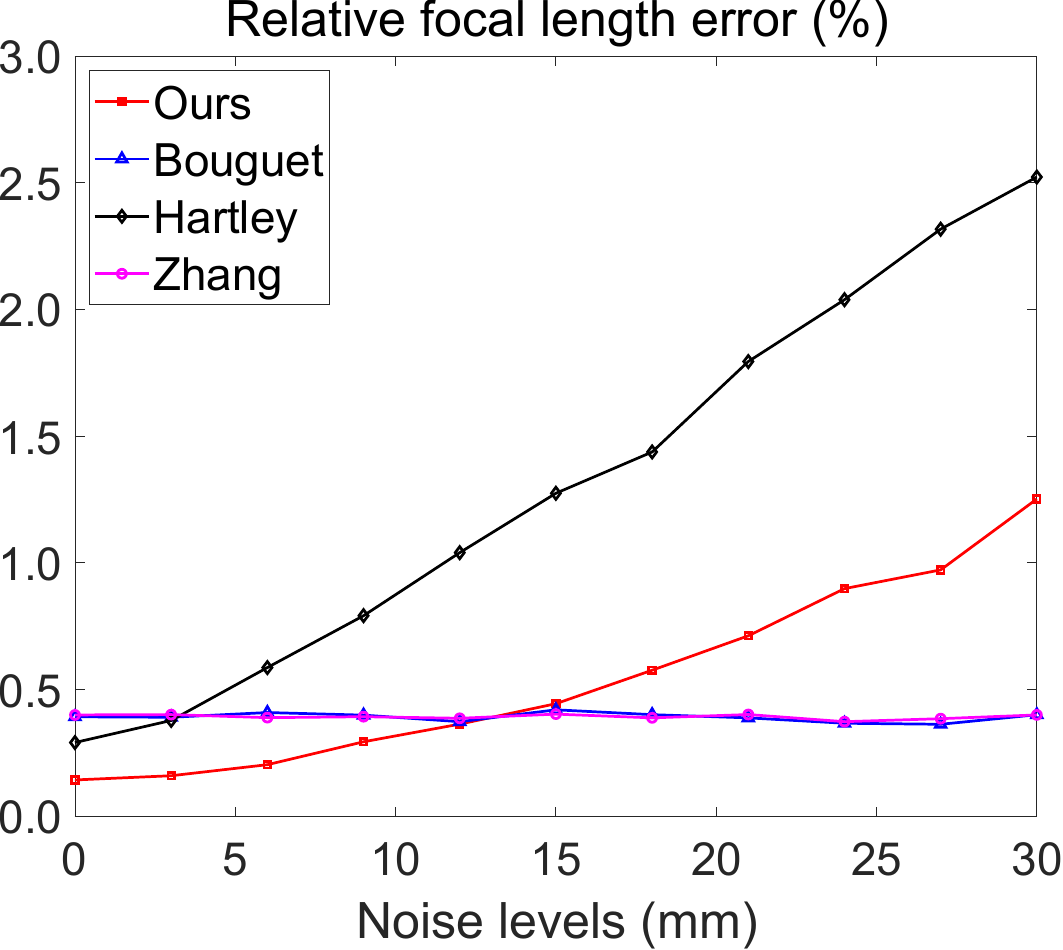} 
	\quad
	\includegraphics[width=0.4\linewidth]{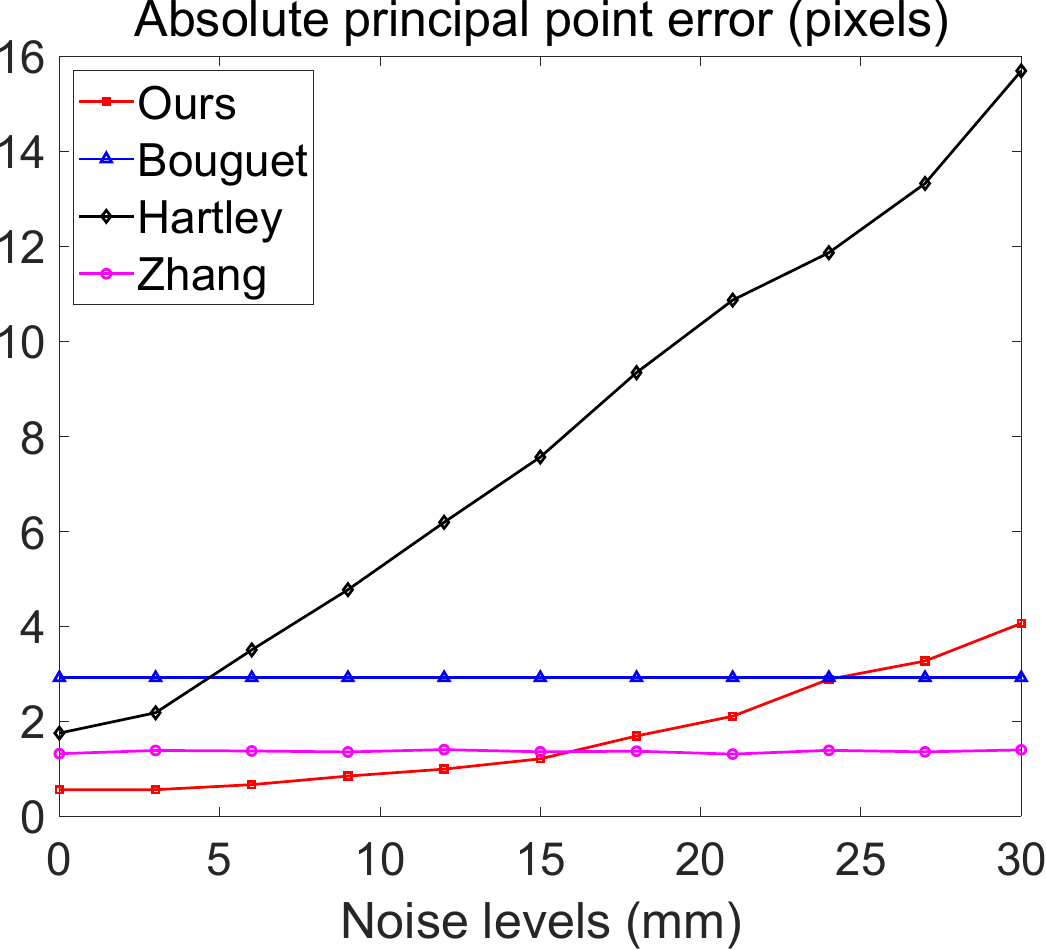}
	\caption{Sensitivity analysis of imperfect spherical motion. \texttt{Bouguet} and \texttt{Zhang} are unaffected by noise. \texttt{Ours} exhibits greater robustness to imperfect collimators compared to the \texttt{Hartley}. At low noise levels, our method continues to perform optimally.}
	\label{fig:Sensi}
\end{figure}

To further evaluate the algorithm's practicality, we test the sensitivity of the initial value estimation to imperfect spherical motion conditions. In real-world applications, low-cost collimators have limited ability to produce perfectly collimated rays and exhibit an inferior parallelism, which is reflected in the fact that the relative motion between the target and camera is not a perfect spherical motion. To replicate this imperfection, {we add zero-mean Gaussian noise with standard deviation ranging from $0mm$ to $30mm$ to $\mathbf{t}_{cp}$.} In the experiment, we use 15 synthetic images and add a noise level of 0.5 pixels to the image points. For each noise level, a total of 500 independent Monte Carlo simulations are conducted, and the average errors are computed. The sensitivity analysis results are shown in Fig.~\ref{fig:Sensi}. Noise has no impact on \texttt{Bouguet} and \texttt{Zhang}, because they do not rely on the spherical motion constraint. However, as the noise level increases, calibration errors of both \texttt{Ours} and \texttt{Hartley} increase gradually. Notably, \texttt{Hartley} shows significant sensitivity to imperfect spherical motion, while \texttt{Ours} displays superior robustness. Even at a noise level as high as $15mm$ (equivalent to $2.143\%$ of the spherical radius), our algorithm still remains superior to the alternatives. It is worth emphasizing that even low-cost collimators typically meet accuracy requirements far below this level.

\subsection{Real Images Experiments}
Using synthetic data, we validated that our algorithm outperforms other methods under spherical motion constraint. In real image experiments, we verify the effectiveness of our collimator system in calibration and evaluate the performance of the proposed algorithm. 

\subsubsection{Spherical Motion Verification}
In Section~\ref{sec:Motion Model}, we demonstrate that the motion of the calibration target relative to a fixed camera can be modeled as a spherical motion model. This experiment validates the spherical motion model using real collimator images. {A gray-scale camera with a resolution of $2448 \times 2048$ pixels and a lens with a focal length of 8$mm$ is employed to capture images from our collimator system.} The camera parameters are obtained from multiple checkerboard images and calibration toolbox \citep{Bouguet2004}. Then, a total of 1000 collimator images are used for verification. Given the calibrated camera parameters, each image's pose in the calibration target's local coordinate system can be calculated using the perspective-n-points (P$n$P) solver \citep{OpenGV}. 
\begin{table*}[tbp]
	\centering
	\caption{Statistical results of image position distribution.}
	\begin{tabular}{p{5cm} >{\centering\arraybackslash}p{2.1cm} >{\centering\arraybackslash}p{2.1cm} >{\centering\arraybackslash}p{2.1cm}}
		\toprule
		{} & x & y &z \\
		\midrule
		\rowcolor{myorg}
		{ Mean$(mm)$ }&  88.1960 &  136.1062 &  -535.9266 \\
		Standard Deviation$(mm)$& 0.4640 & 0.4693 & 0.1910 \\
		\rowcolor{myorg}
		Range$(mm)$ &  2.2163 &  2.2179 &  0.9915\\
		Relative Standard Deviation$(\%)$& 0.0866 & 0.0876 & 0.0356 \\
		\rowcolor{myorg}
		Relative Range$(\%)$ &   0.4135 &  0.4138 &  0.1850\\
		\bottomrule
	\end{tabular}
	\label{tab:CamPos}
\end{table*}
\begin{figure}[tbp]
	\centering
	\includegraphics[width=0.5\linewidth]{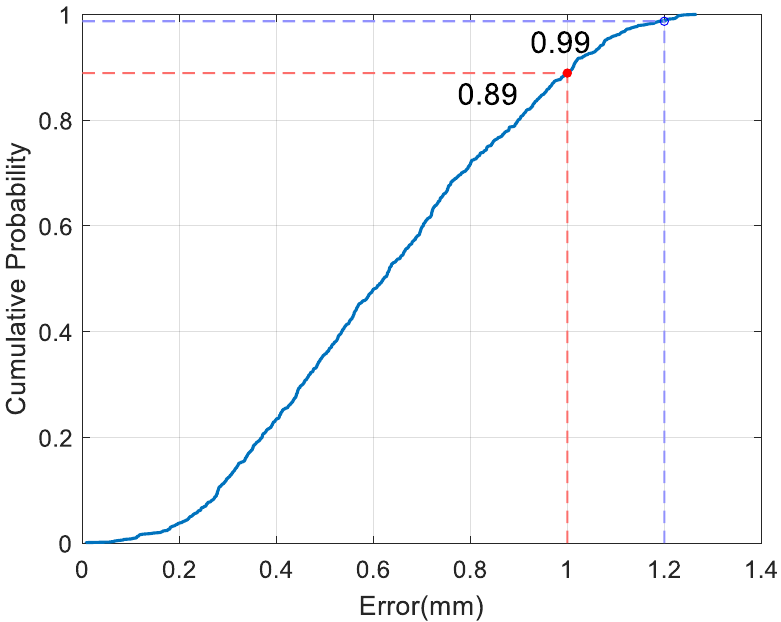} 
	\caption{Cumulative probability distribution function of image position errors. 99\% of the position errors are less than 1.2 mm.}
	\label{fig:SMV1}
\end{figure}

{The statistics on the position of 1000 images within the calibration target coordinate system are presented in Table~\ref{tab:CamPos}.} The centroid of all positions is located at $\mathbf{t}_{cp} = (88.1960, 136.1062, -535.9266)$ $mm$. In a perfect spherical motion model, the positions of images should exhibit consistency, reflecting the ideal geometric relationships. However, {these positions may experience slight offsets in practice due to real-world noise.} These offsets can arise from various factors, such as feature extraction and camera calibration. The average radius of the spherical motion is 535.9266 $mm$. Notably, the standard deviation of the image positions is only ($0.0866\%, 0.0876\%, 0.0356\%$) relative to the spherical radius, which indicates a tightly concentrated distribution. In the synthetic data experiments of the paper, we show that our algorithm is superior to other algorithms when the imperfect spherical noise level is less than $2.143\%$ of the spherical radius. The parallelism of our collimator is better than this noise level. The range represents the side length of the minimum enclosing cuboid that encompasses all the positions. The variation range of the image positions relative to the spherical radius is ($0.4135\%, 0.4138\%, 0.1850\%$). In addition, we calculated the Euclidean distance between each position and $\mathbf{t}_{cp}$ to quantify the error, {and the cumulative distribution function of the errors is presented in Fig.~\ref{fig:SMV1}}. We can see that 89\% of the position errors are less than 1.0 $mm$, and 99\% are less than 1.2 $mm$. In summary, it can be concluded that the image positions are nearly invariant. 
\begin{figure}[tbp]
	\centering
	\includegraphics[width=0.6\linewidth]{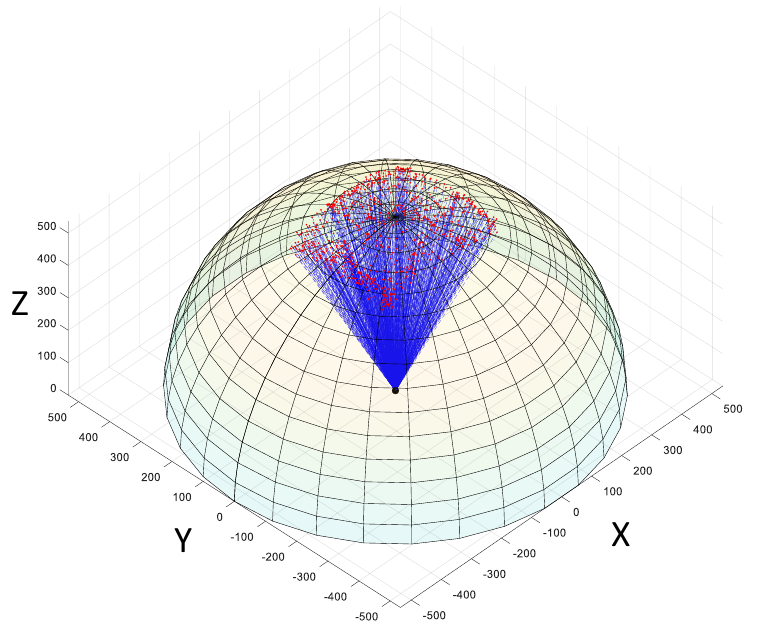} 
	\caption{Visualization of the pose of the calibration targets. The \textcolor{black}{black dots} represents the camera optical center located at the origin of the coordinate system. The {red dots} represent the positions of calibration targets, which fit nicely on the transparent sphere. The \textcolor{blue}{blue lines} depict the $z$-axis of the calibration patterns, converging towards the camera optical center, indicating that the calibration target's motion conforms to the spherical motion model.}
	\label{fig:SMV2}
\end{figure}

Moreover, we set the camera as the fixed reference coordinate system and determine the pose of all calibration targets. To visualize this, we generate a sphere with the camera's optical center as its center and the average distance as its radius ($r = 535.9266$ $mm$). Subsequently, we depict the pose of all calibration targets on this sphere. The local coordinate system of the calibration targets is translated from the upper left corner to the $(88.1960, 136.1062, 0.0)$ $mm$, but the directions remain unchanged. {Figure~\ref{fig:SMV2} shows the pose distribution of the calibration target within the camera coordinate system corresponding to $\mathbf{T}_{pc}$ in Eq.~\eqref{eq:Tpc}. As depicted in Fig.~\ref{fig:SMV2}}, the position of the calibration targets (red dots) closely fits the spherical surface, with an average distance error of 0.3 $mm$. {The $z$-axis of calibration targets (blue lines) consistently aligns with the camera optical center (Black dot).} The blue lines intersect with the $XY$-plane. The centroid of intersections is located at $(-0.0083, 0.0165, 0.0)mm$, with a standard deviation of $(0.4770, 0.4821, 0.0)mm$. {In summary, we jointly validated the spherical motion model from both the image positions and the calibration target poses.}

\subsubsection{Camera Pose Estimation Results}
\begin{figure}[tbp]
	\centering
	\includegraphics[width=0.19\linewidth]{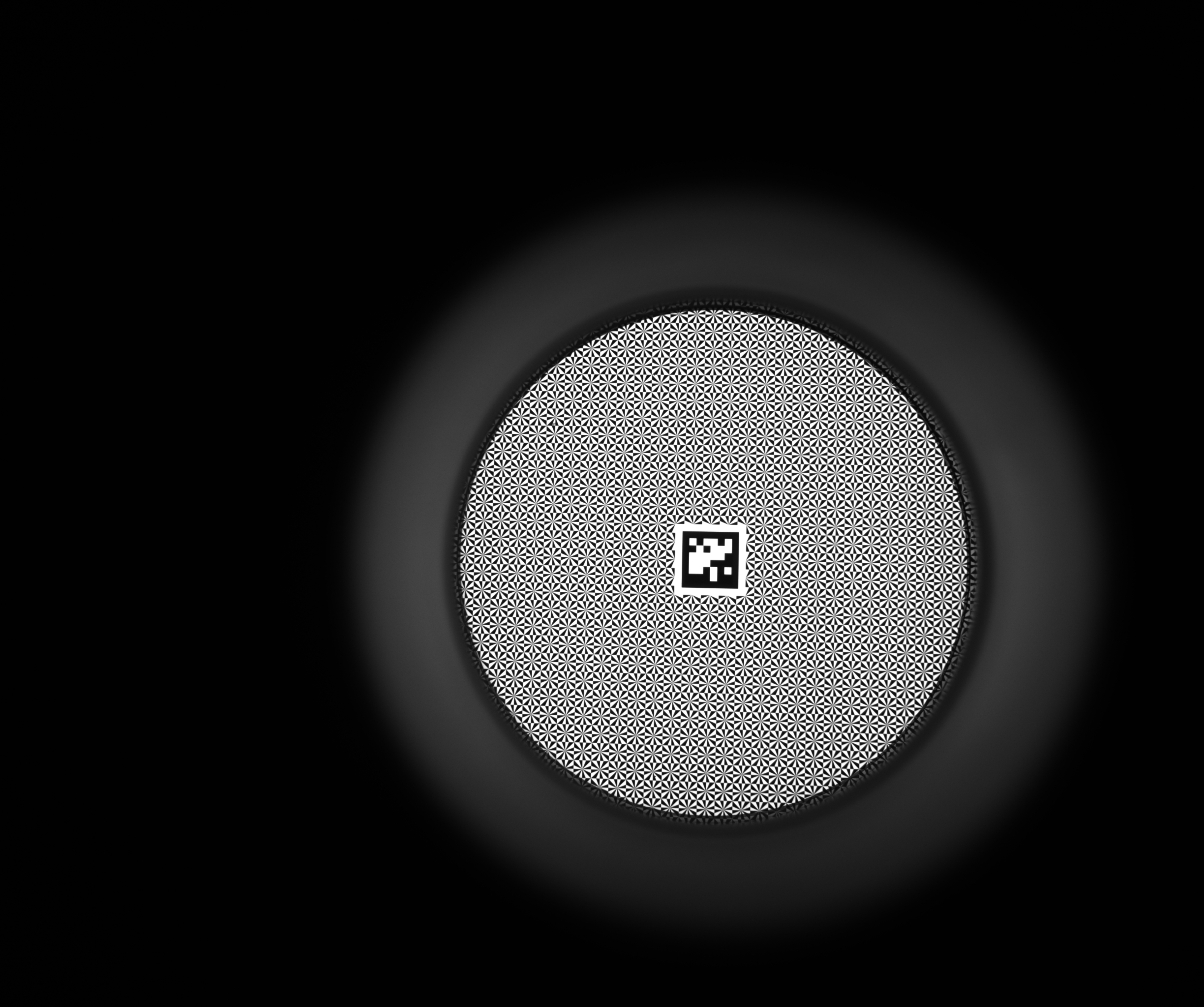} 
	\includegraphics[width=0.19\linewidth]{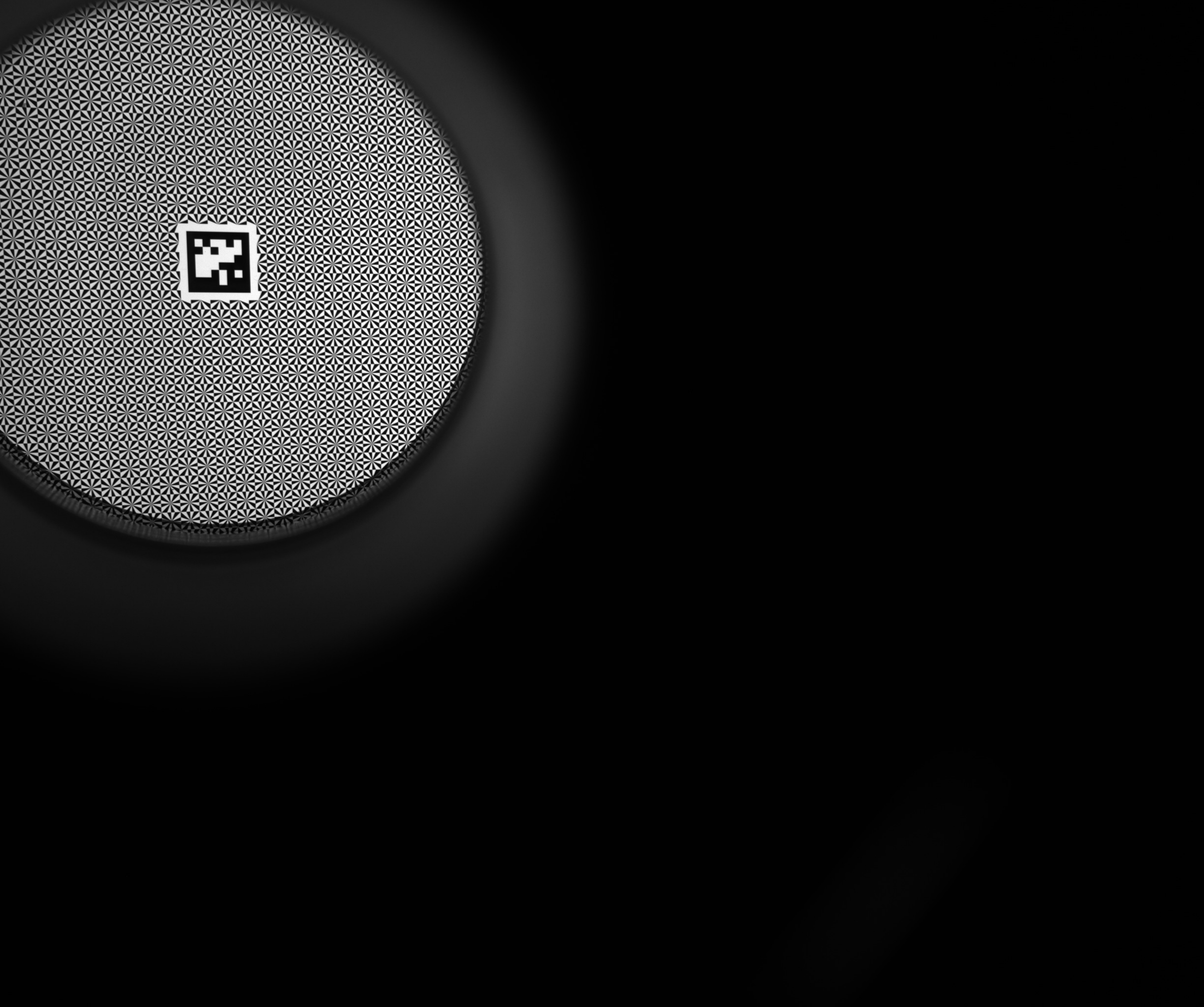} 
	\includegraphics[width=0.19\linewidth]{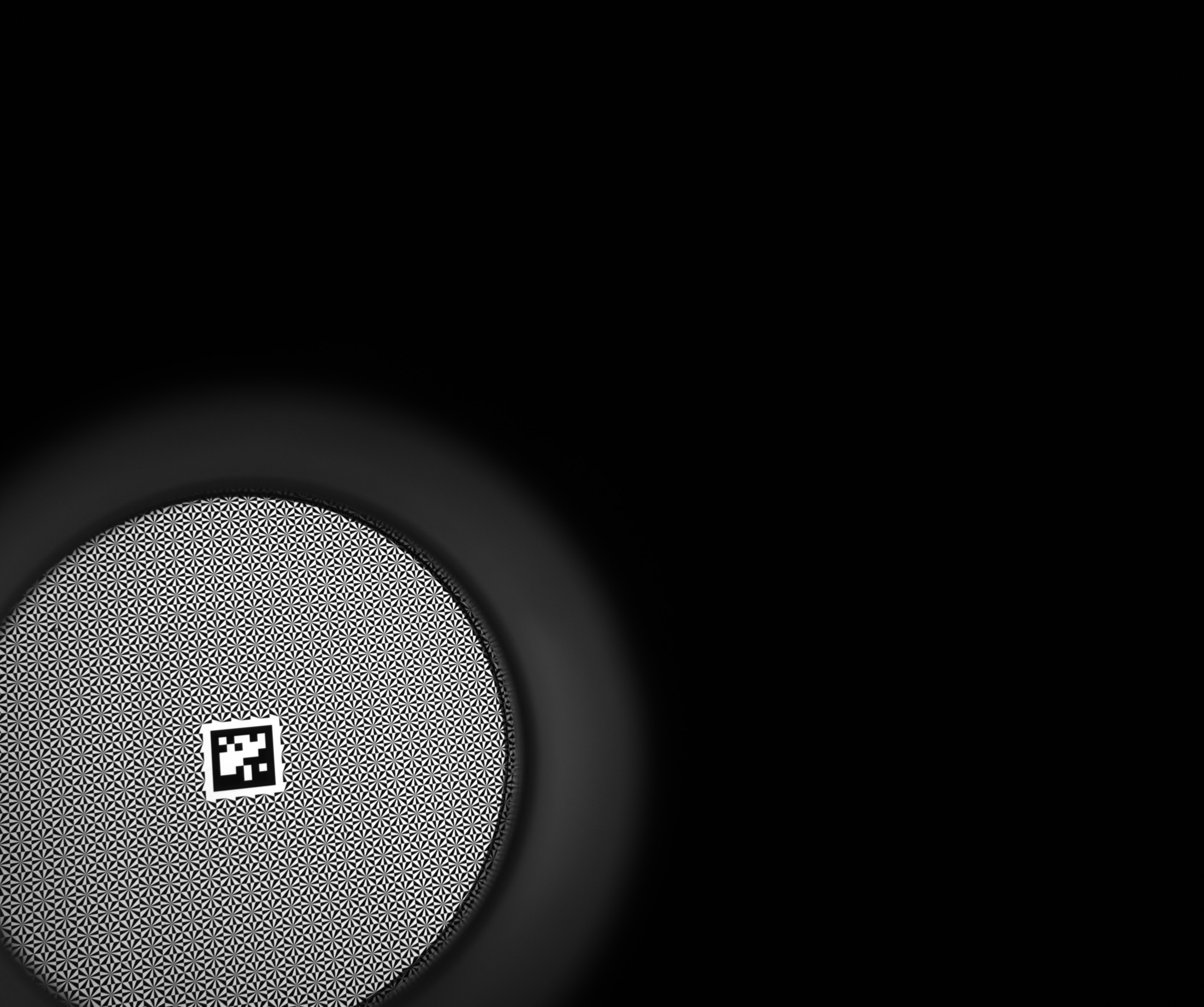}
	\includegraphics[width=0.19\linewidth]{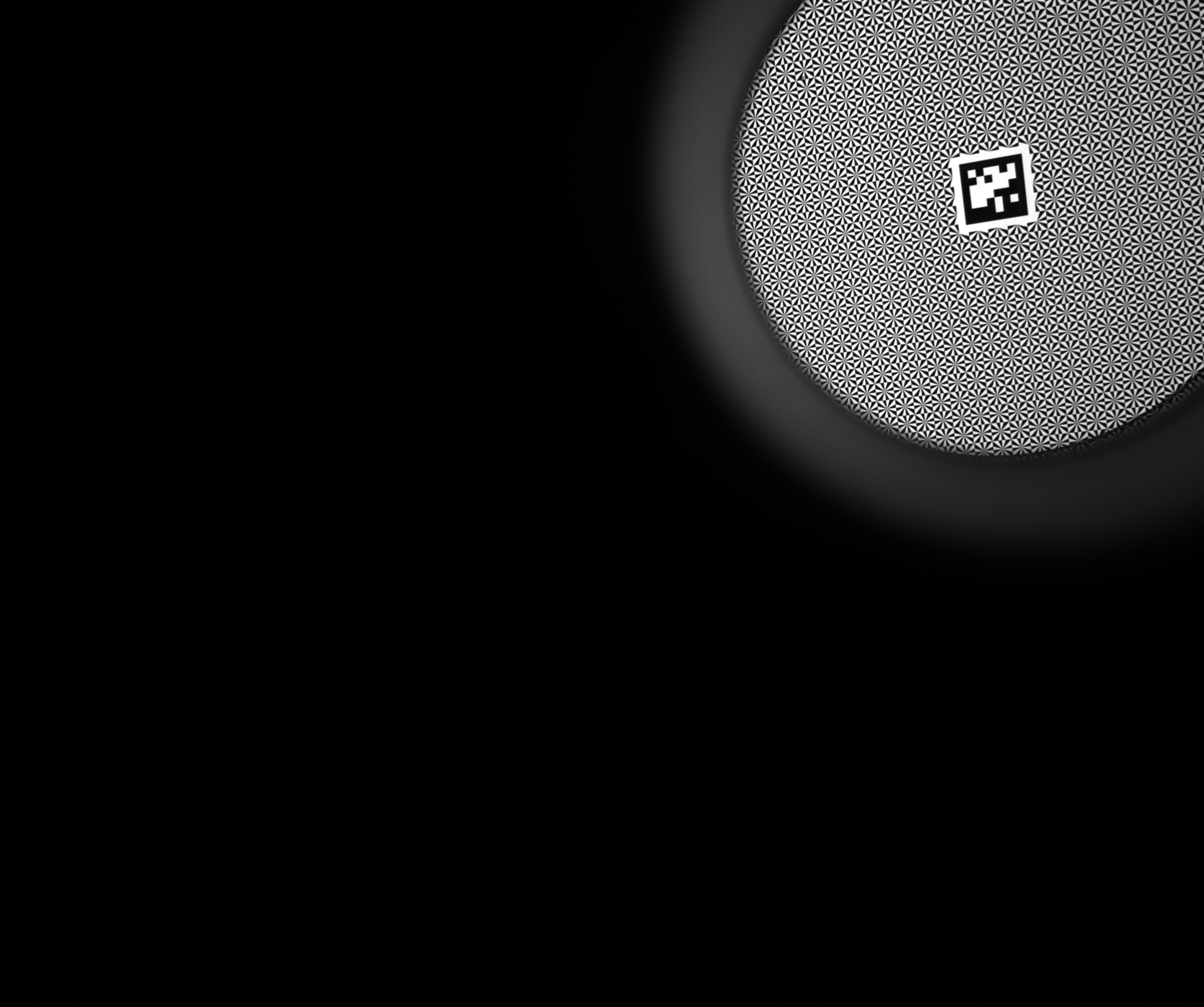}
	\includegraphics[width=0.19\linewidth]{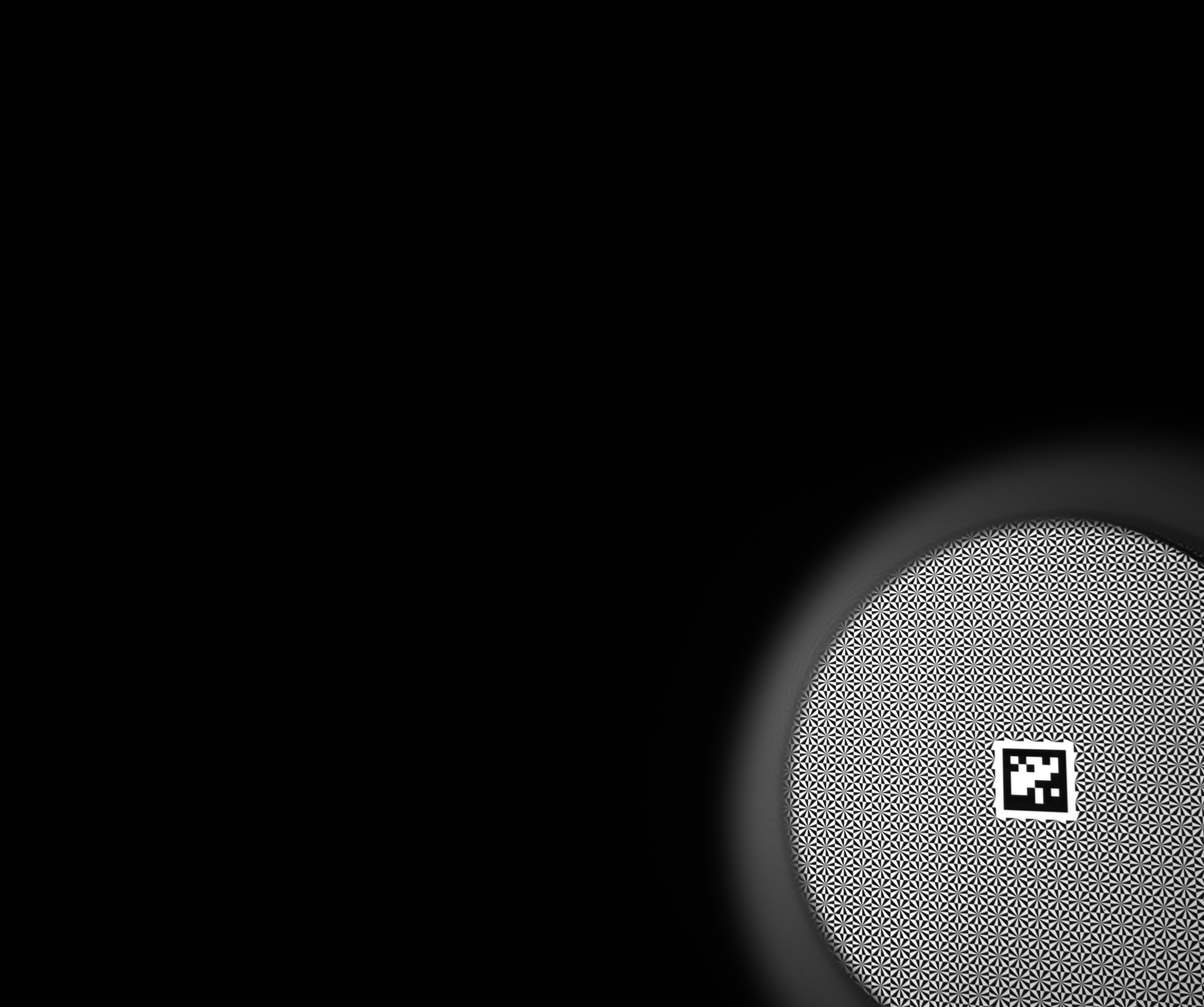}
	\caption{Samples of collimator images used for calibration. The camera observes the calibration target in our collimator system from different directions.}
	\label{fig:collimator_8mm_images}
\end{figure}

Following the experimental methodology utilized in previous studies \citep{Peng2019,Ha2017}, we design a pose estimation experiment to evaluate the accuracy of camera calibration quantitatively. The camera used in the experiment has a focal length of $8mm$ and a resolution of $2448 \times 2048$ pixels. Images are captured by the camera from both our collimator system and an actual printed pattern. Figure~\ref{fig:collimator_8mm_images} provides some samples of collimator images. The collimator images are used solely for camera calibration. A subset of the printed pattern images is used for calibration, while the remaining images are used for pose estimation to evaluate the calibration results. A total of 200 printed pattern images are used to evaluate the calibration. The number of images used for calibration is 2, 5, 10, 15, and 20.

After obtaining the camera parameters from the calibration images, the pose of each evaluation image can be determined using absolute pose estimation \citep{OpenGV}. In our setup, a quarter of the corner points on the printed pattern are used for pose estimation, and the remaining points are used for evaluation. By re-projecting the evaluation points onto the image with the camera parameters and the calculated pose, we can evaluate the calibration quality based on the re-projection error. In addition, the epipolar error between image pairs is calculated and used for evaluation. Specifically, the essential matrix between the image pairs is estimated using the camera parameters, and the epipolar error is computed based on this matrix. The epipolar error refers to the perpendicular distance from an image point to its corresponding epipolar line, which has not been optimized. {The re-projection errors of the calibration and evaluation images are denoted as $\varepsilon_{calib}$ and $\varepsilon_{eval}$, respectively.} The epipolar error is denoted as $\varepsilon_{epip}$. 

{In this experiment, we additionally compare the \texttt{BabelCalib} \citep{Lochman2021} on the collimator images, which regresses camera parameters using a back-projection model.} The comparison results of the four algorithms are shown in Table~\ref{tab:Algo}. \texttt{Hartley} method only uses the correspondence of 2D points in images, which makes it impossible to calculate the re-projection error of the calibration images directly. {From Table~\ref{tab:Algo}, we can infer that the calibration accuracy gradually improves and stabilizes as the number of images increases.} When only two calibration images are available, the spherical motion constraint provided by our collimator system is relatively weak. \texttt{Ours} shows similar accuracy to that of \texttt{Bouguet}. As the number of images increases, the spherical motion constraint helps us achieve better calibration results. Due to the limited degrees of freedom, \texttt{Ours} shows a slightly higher re-projection error on the calibration images ($\varepsilon_{calib}$) compared to other methods. However, \texttt{Ours} exhibits the lowest re-projection error ($\varepsilon_{eval}$) and epipolar error ($\varepsilon_{epip}$) in evaluation images compared to alternative methods, indicating that our method achieves more accurate camera parameters. {\texttt{BabelCalib} does not demonstrate better results and shows slightly lower accuracy than the \texttt{Bouguet}. We believe that \texttt{BabelCalib}'s advantage lies in the estimation of large distortion nonlinear parameters for cameras.} In summary, these experimental results strongly validate that our proposed algorithm offers significant advantages when using collimator images for camera calibration, providing more accurate and reliable results.

\begin{table*}[htbp]
	\centering
	\LARGE
	\caption{Comparison of calibration and pose estimation results with different numbers of collimator images.}
	\resizebox{\linewidth}{!}{
		\begin{tabular}{llllll}
			\toprule
			{}&\multicolumn{1}{c}{Zhang}&\multicolumn{1}{c}{Bouguet}&\multicolumn{1}{c}{Hartley} &\multicolumn{1}{c}{BabelCalib} &\multicolumn{1}{c}{Ours}
			\\
			\cmidrule{2-6}
			{Num}&\multicolumn{5}{c}{{$\varepsilon_{calib}$ (pixels)}, {$\varepsilon_{eval}$ (pixels)}, {$\varepsilon_{epip}$ (pixels)}}
			\\
			\midrule
			\rowcolor{myorg}
			{2}&{{0.1842},  {0.1545},  {0.0846}} &{{0.1739}, \textbf{0.1518}, \textbf{0.0770}} &{{-},  {-},  {-}} & {0.1805, 0.1533, 0.0787} &{{0.1739}, {0.1519}, \textbf{0.0770}}
			\\
			{5}&{{0.2316}, {0.1245}, {0.0809}}&{{0.2202}, \textbf{0.1236}, {0.0764}}&{{-}, {0.2577}, {0.0832 }} & {0.2221, 0.1238, 0.0768} &{{0.2203}, \textbf{0.1236}, \textbf{0.0762}}
			\\
			\rowcolor{myorg}
			{10}&{{0.1923},  {0.1231},   {0.0773}} &{{0.1912},  {0.1232},  {0.0758}}
			&{{-},  {0.1986},  {0.0854}} & {0.1934, 0.1247, 0.0784}
			&{{0.1951},  \textbf{0.1222},  \textbf{0.0757}}
			\\
			{15}&{{0.1775}, {0.1221}, {0.0782}}
			&{{0.1718}, {0.1232}, {0.0771 }}
			&{{-}, {0.1791}, {0.0837}} & {0.1811, 0.1235, 0.0780}
			&{{0.1779}, \textbf{0.1216}, \textbf{0.0756}}
			\\
			\rowcolor{myorg}
			{20}&{{0.1773},  {0.1225},  {0.0772}}
			&{{0.1722},  {0.1221},  {0.0768}}
			&{{-},  {0.1871},  {0.0818}} & {0.1803, 0.1231, 0.0778}
			&{{0.1778}, \textbf{0.1207}, \textbf{0.0756}}
			\\
			\bottomrule
			\multicolumn{5}{l}{\textbf{Bold} values indicate the best results.}
		\end{tabular}
	}
	\label{tab:Algo}
\end{table*}

\begin{table}[htbp]
	\centering
	\caption{Comparison of calibration and pose estimation results using 20 collimator images and printed pattern images.}
		\begin{tabular}{p{1cm} >{\centering\arraybackslash}p{1.3cm} >{\centering\arraybackslash}p{1.5cm} >{\centering\arraybackslash}p{1.5cm} >{\centering\arraybackslash}p{1.5cm} >{\centering\arraybackslash}p{1.5cm} >{\centering\arraybackslash}p{1.5cm}}
			\toprule
			\multirow{2}{*}{}&{Printed}&\multicolumn{5}{c}{Collimator} \\
			\cmidrule(lr){3-7}
			{}&{Bouguet}&{Zhang}&{Bouguet}&{Hartley}&{BabelCalib}&{Ours}\\
			\midrule
			\rowcolor{myorg}
			{$f_x$}& {2384.5}& {2380.8}& {2369.2}& {2644.6}& {2368.5} & {2388.3}\\
			{$f_y$}&{2386.3}&{2379.8}&{2368.9}&{2646.0} &{2368.7}&{2387.9}\\
			\rowcolor{myorg}
			{$c_x$}& {1209.8}& {1222.1}& {1221.1}& {1214.6}&{1219.7}&{1221.0}\\
			{$c_y$}&{1008.2}&{1008.4}&{1009.8}&{1012.7}&{1020.5}&{1009.6}\\
			\rowcolor{myorg}
			{$k_1$}& {-0.0843}& {-0.0899}& {-0.0908}& {-0.0754}& {-0.0892} & {-0.0900}\\
			{$k_2$}&{0.0819}&{0.0912}&{0.0892}&{0.1181} & {-0.0972} &{0.0954}\\
			\rowcolor{myorg}
			{$\varepsilon_{calib}$}& {0.1513}& {0.1773}& {0.1722}& {-}& 0.1803 & {0.1778}\\
			\midrule
			{$\varepsilon_{eval}$}&{0.1225}&{0.1225}&{0.1221}&{0.1871}&0.1231&\textbf{0.1207}\\
			\rowcolor{myorg}
			{$\varepsilon_{epip}$}& {0.0772}& {0.0772}& {0.0768}& {0.0818}& 0.0778 & \textbf{0.0756}\\
			\bottomrule
			\multicolumn{6}{l}{\textbf{Bold} values indicate the best results.}
		\end{tabular}
	\label{tab:Algo2}
\end{table}

Then, we compare the calibration results of the collimator images with those of the printed pattern images. Specifically, {the \texttt{Bouguet} toolbox is used to calculate camera parameters from printed pattern images that are not included in the evaluation. We use 20 images for calibration and 200 images for evaluation.} The estimated camera parameters and error comparison are shown in Table~\ref{tab:Algo2}. The combination of \texttt{Bouguet} with printed images is a mainstream calibration process, and its results can serve as a valuable reference. \texttt{Hartley} has significant limitations in handling image distortion, resulting in larger errors. The lower evaluation error indicates that our method outperforms other approaches. The experimental results validate the effectiveness of using collimator images for calibration, while also demonstrating the positive impact of spherical motion constraint on improving calibration accuracy. It can be expected that under good visibility conditions, the parameters obtained by different calibration methods typically do not differ significantly. Therefore, even a minor improvement holds considerable significance. 

\subsubsection{Structure from Motion Results.}
\label{sec:SfM}
\begin{figure}[tbp]
	\centering
	\subfloat[Sand table image]{
		\includegraphics[width=0.32\linewidth]{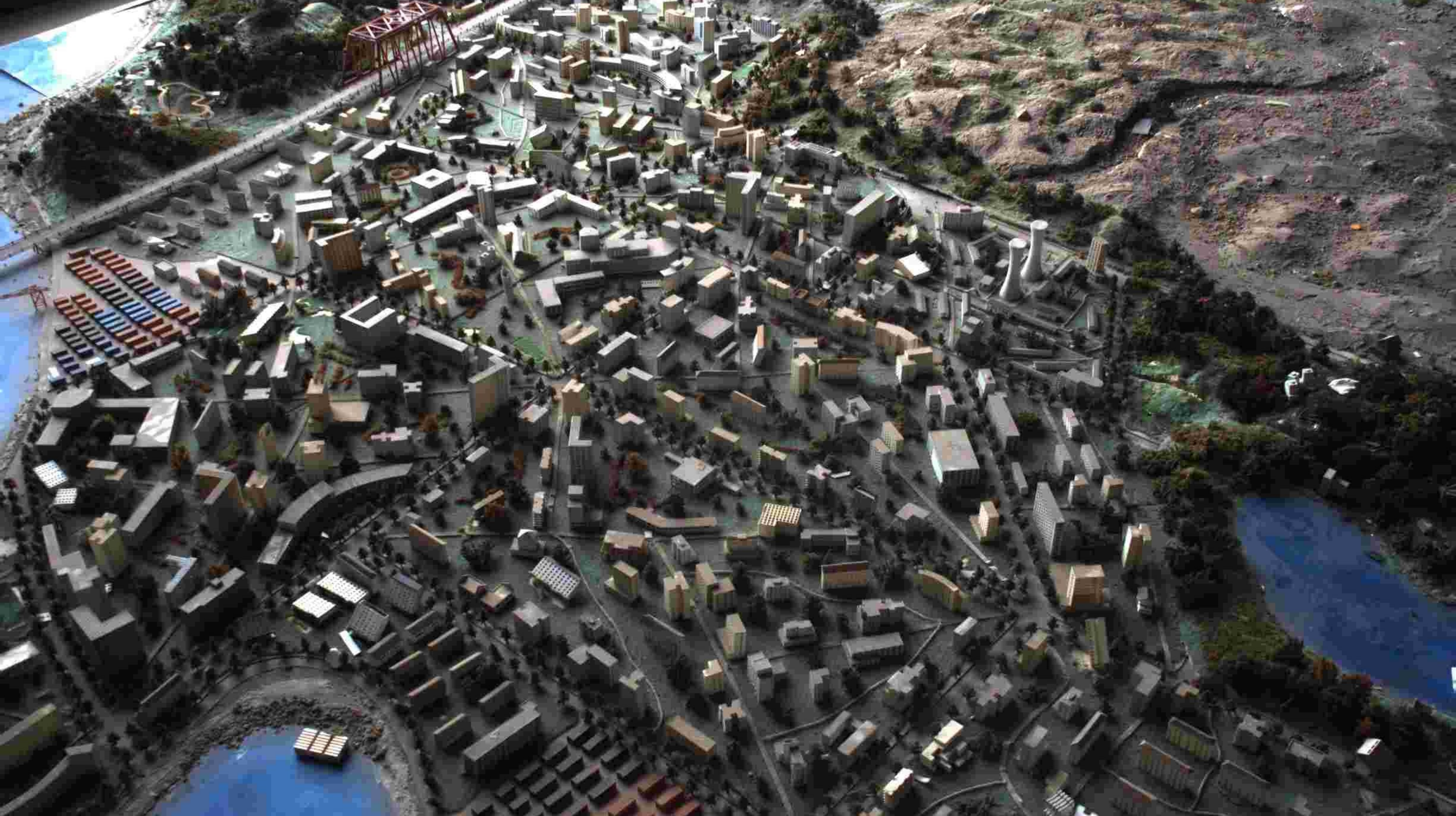} 
		\label{fig:sandTable}
	}
	\subfloat[Bouguet+Printed]{
		\includegraphics[width=0.32\linewidth]{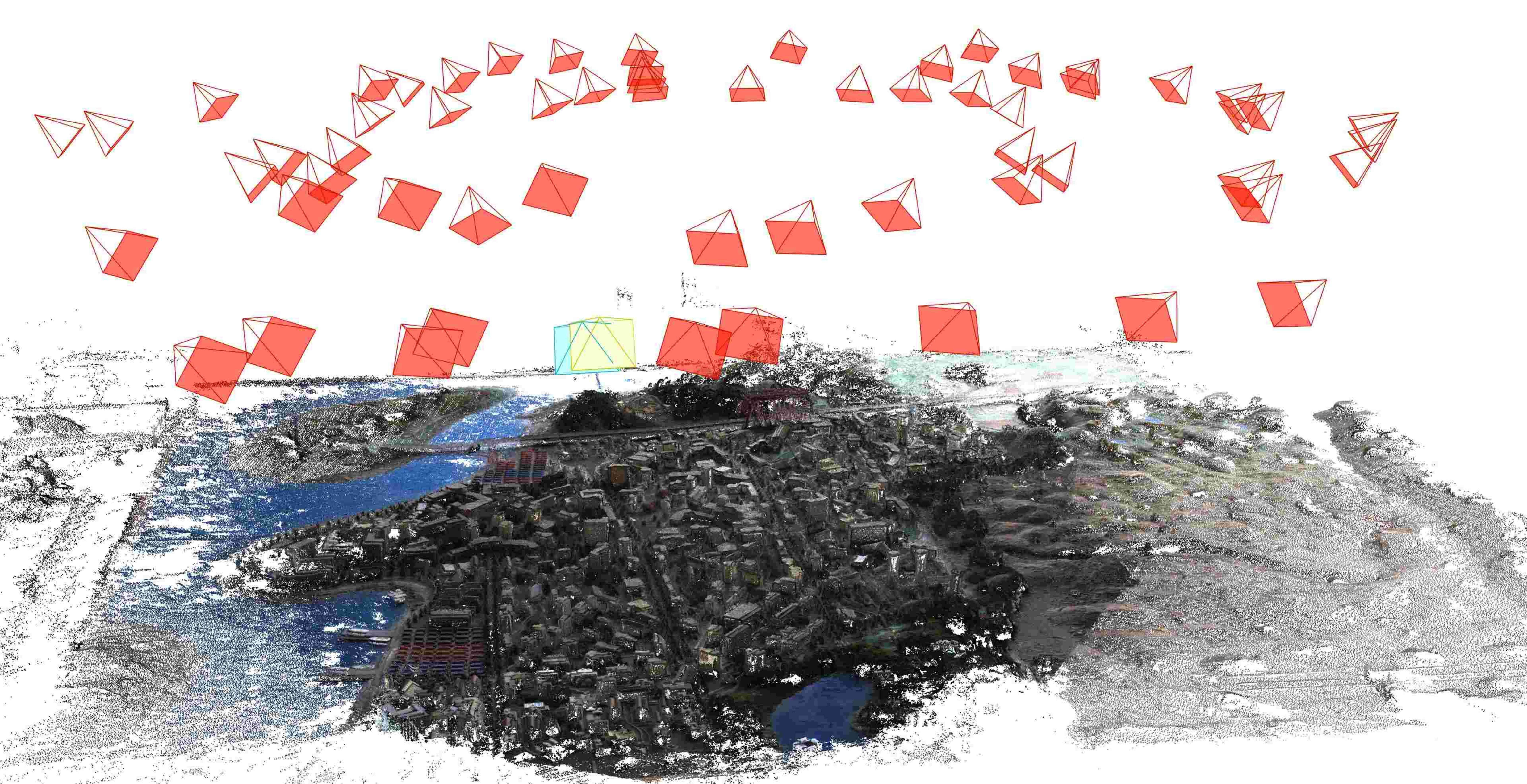}
		\label{fig:bouguet_printed}
	}
	\subfloat[Zhang+Collimator]{
		\includegraphics[width=0.32\linewidth]{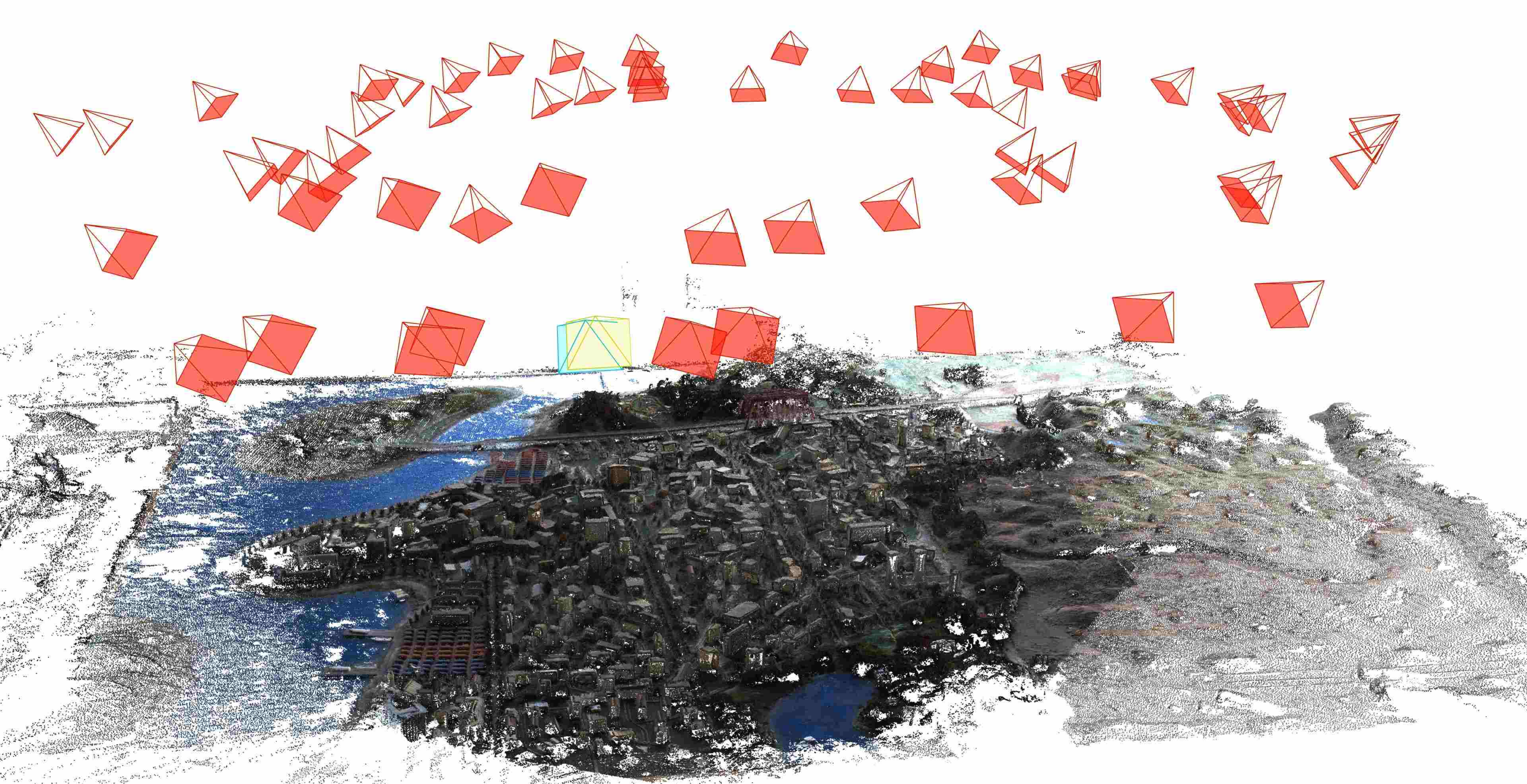}
		\label{fig:zhang_collimator}
	}	\\
	\subfloat[Bouguet+Collimator]{
		\includegraphics[width=0.32\linewidth]{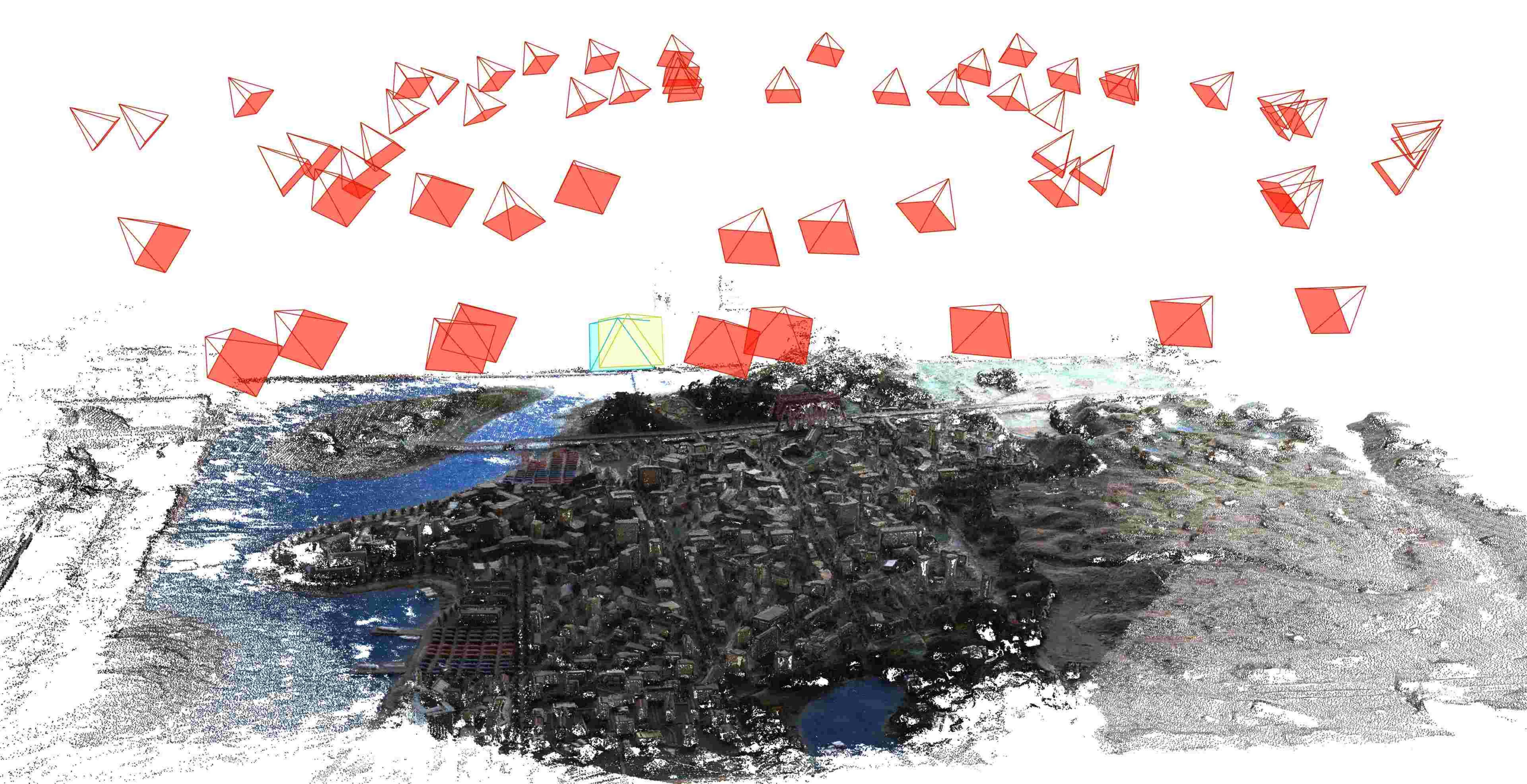}
		\label{fig:bouguet_collimator}
	}
	\subfloat[Hartley+Collimator]{
		\includegraphics[width=0.32\linewidth]{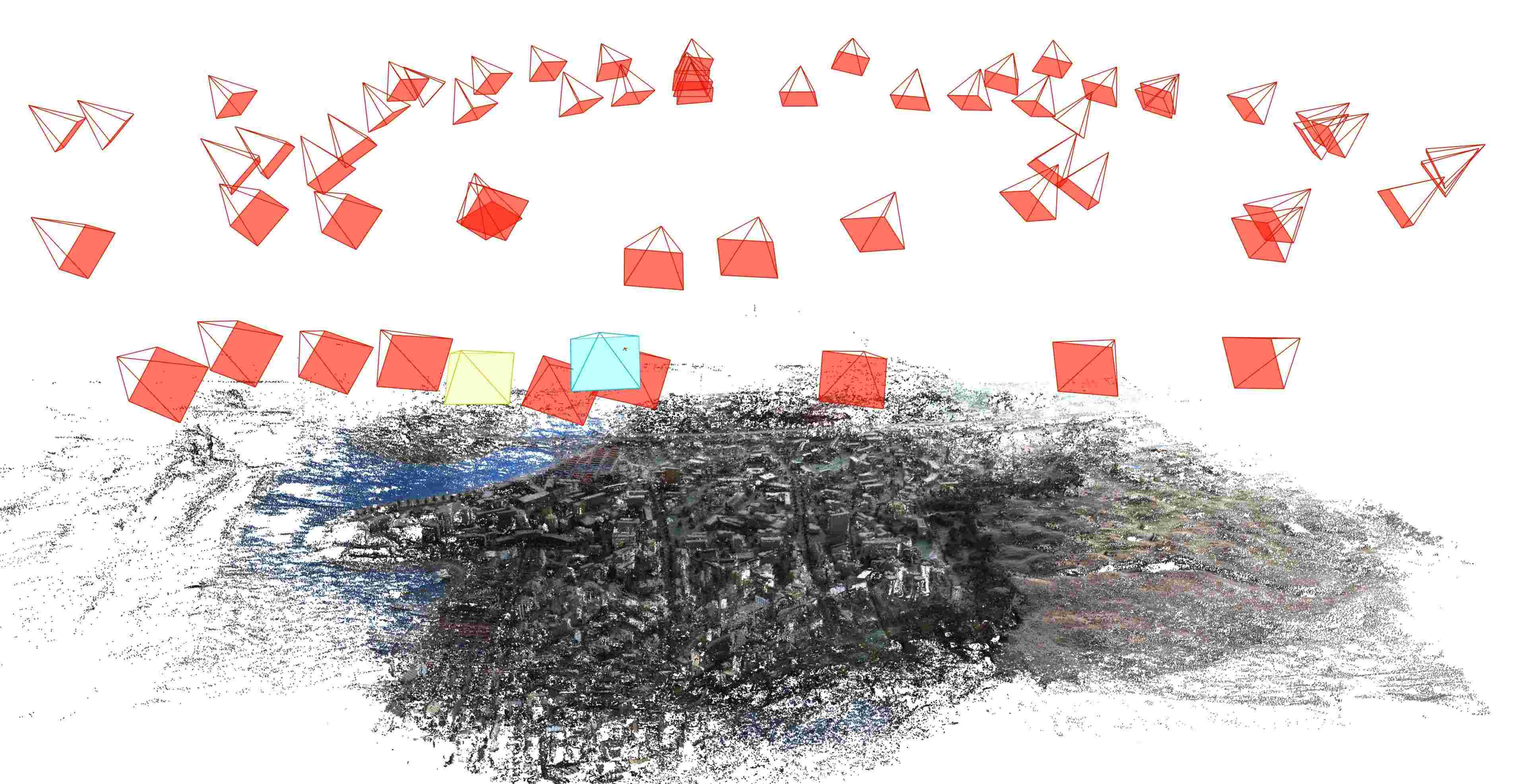}
		\label{fig:rotation_collimator}
	} 
	\subfloat[Ours+Collimator]{
		\includegraphics[width=0.32\linewidth]{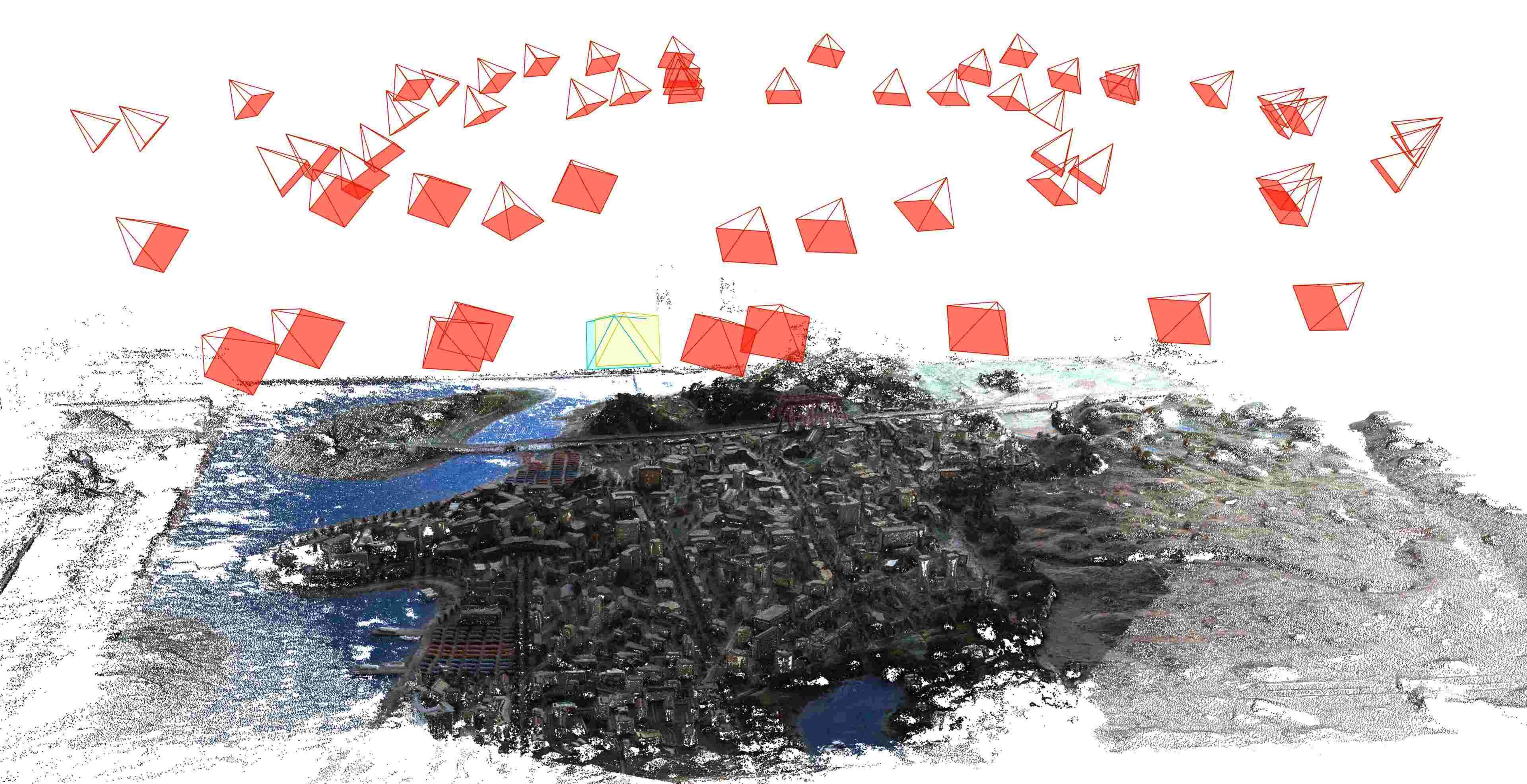}
		\label{fig:Ours_collimator}
	}
	\caption{\textbf{(a):} Samples of urban sandbox images and \textbf{(b-f):} dense reconstruction results from different calibrated camera parameters.  The poses of the images are displayed on the reconstruction results, with the first and last images specifically marked in cyan and yellow, respectively.}
	\label{fig:SfM3D}
\end{figure}

In this experiment, we conduct a structure from motion (SfM) experiment to verify the accuracy and relevance of the proposed methods. The incremental SfM method COLMAP \citep{Colmap2016} is used to reconstruct an urban sand table, as shown in Fig.~\ref{fig:sandTable}. Our SfM experiment is inspired by \citep{Peng2019}, which is an interesting experiment and allows for quantitative evaluation of calibration results. We capture a sequence of images around the sand table and duplicate the first image to the end of sequence. The camera parameters are obtained from printed pattern images and collimator images and serve as input to the COLMAP. The printed pattern images have multiple tilt angles to ensure accurate calibration. Before reconstruction, there are some specific settings. During the feature matching, we employed sequence matching instead of exhaustive matching, where only consecutive adjacent images are matched. Meanwhile, loop detection is disabled. With this matching approach, reconstruction is performed in the order of the images, and the pose estimation errors are accumulated, thereby highlighting the more accurate calibration results. During the 3D reconstruction, the initial image pair is fixed as the first two images of the sequence. The camera parameters are not optimized. In this configuration, we can evaluate the calibration results using the image poses and the 3D point cloud returned by SfM. For a more accurate calibration, the pose deviation between the first and last images returned by SfM is smaller. 

\begin{table}[tbp]
	\centering
	\caption{Comparison of SfM results for the urban sand table.}
	\begin{tabular}{p{2.5cm} p{1.6cm} p{1.6cm} p{1.6cm} p{1.6cm} p{1.6cm}}
		\toprule
		\multirow{2}{*}{}&\multicolumn{1}{l}{Printed}&\multicolumn{4}{c}{Collimator} \\
		\cmidrule(lr){3-6}
		{}&{Bouguet}&{Zhang}&{Bouguet}&{Hartley}&{Ours}\\
		\midrule
		\rowcolor{myorg}
		{$\varepsilon_\mathbf{R}$ (degree)}& {0.9821}& {0.5012}& {0.5049}& {8.2561}& {\textbf{0.3971}}\\
		{$\varepsilon_\mathbf{t}$ (\%)}&{0.5692}&{0.3260}&{0.3321}&{3.1687}&{\textbf{0.2871}}\\
		\rowcolor{myorg}
		{$\varepsilon_{proj}$ (pixels)}& {18.6869}& {11.2434}& {10.7638}& {150.5852}& {\textbf{10.1284}}\\
		{$\varepsilon_{SfM}$ (pixels)}& {0.6121}& {0.5989}& {0.6007}& {0.8536}& {\textbf{0.5776}}\\
		\rowcolor{myorg}
		{Point Count}&{{34412}}&{34395}&{34109}&{34340}&\textbf{34858}\\
		\bottomrule
		\multicolumn{6}{l}{\textbf{Bold} values indicate the best results.}
	\end{tabular}
	\label{tab:SfM}
\end{table}

The pose deviation is measured by rotation error $\varepsilon_{\mathbf{R}}$, translation error $\varepsilon_{\mathbf{t}}$. They are calculated as: $\varepsilon_{\mathbf{R}} = arccos((trace(\mathbf{R}_{first} \mathbf{R}_{last}^T)-1)/2)$ and $\varepsilon_{\mathbf{t}} = 2\|\mathbf{t}_{first} - \mathbf{t}_{last}\|/(\|\mathbf{t}_{first}\| + \|\mathbf{t}_{last}\|) \times 100\%$, {where $(\mathbf{R}_{first}, \mathbf{t}_{first})$ and $(\mathbf{R}_{last}, \mathbf{t}_{last})$ represent the rotation matrices and translation vectors of the first and last images}, respectively. Additionally, we indirectly assess the pose differences using a re-projection error $\varepsilon_{proj}$. This involves projecting the reconstructed 3D points onto the first image using the pose of the last image and the camera parameters. Then, we calculate the pixel error of the corresponding 2D points. 

Figure~\ref{fig:SfM3D} shows the urban sandbox and the dense reconstruction results obtained under different camera parameters. The poses of all images are displayed, with the first and last images specifically marked in cyan and yellow, respectively. Intuitively, {the results from our method show a higher level of alignment between the two images.} Quantitative results are shown in Table~\ref{tab:SfM}, where "Points" represents the total number of 3D reconstructed points and $\varepsilon_{SfM}$ denotes the average re-projection error of the SfM process. These two metrics for 3D reconstruction also indirectly reflect the quality of the calibration results. {Our method exhibits minimal pose deviation and maintains a minimal average re-projection error during the SfM process. The SfM outcomes indicate improved accuracy with collimator images and highlight the beneficial impact of the spherical constraint on accuracy.} Additionally, our collimator system offers a stable and controllable calibration environment, which is also a significant advantage. The printed pattern is often affected by inappropriate ambient lighting and the limitations of calibration space. As a result, collimator images yield better results even with the same \texttt{Bouguet} calibration algorithm.

\subsubsection{Single-Image Calibration Results}
In this experiment, we verify the effectiveness and accuracy of calibration using a single collimator image. Three different camera configurations are used in the experiment: $C_1$ (focal length $16mm$, resolution $1440 \times 1080$ pixels), $C_2$ (focal length $20mm$, resolution $2448 \times 2048$ pixels) and $C_3$ (focal length $25mm$, resolution $2448 \times 2048$ pixels). The purpose of selecting these configurations is to cover different focal lengths and resolutions, thereby providing a more comprehensive evaluation of our method's applicability. 

\begin{figure}[tbp]
	\centering
	\subfloat[ Reference image]{
		\includegraphics[width=0.23\linewidth]{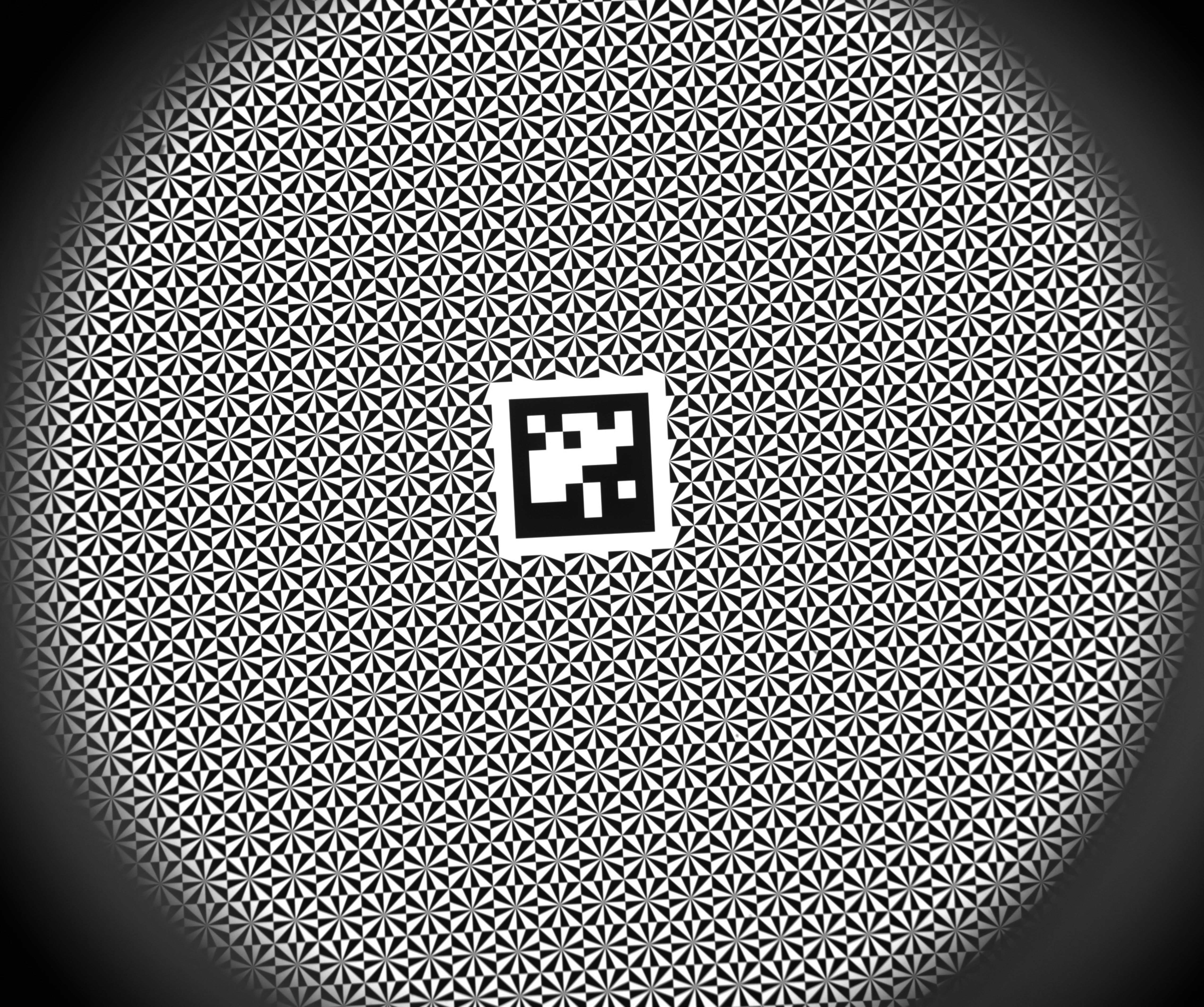} 
		\label{fig:R20}
	}
	\subfloat[ Calibration image $C_1$]{
		\includegraphics[width=0.255\linewidth]{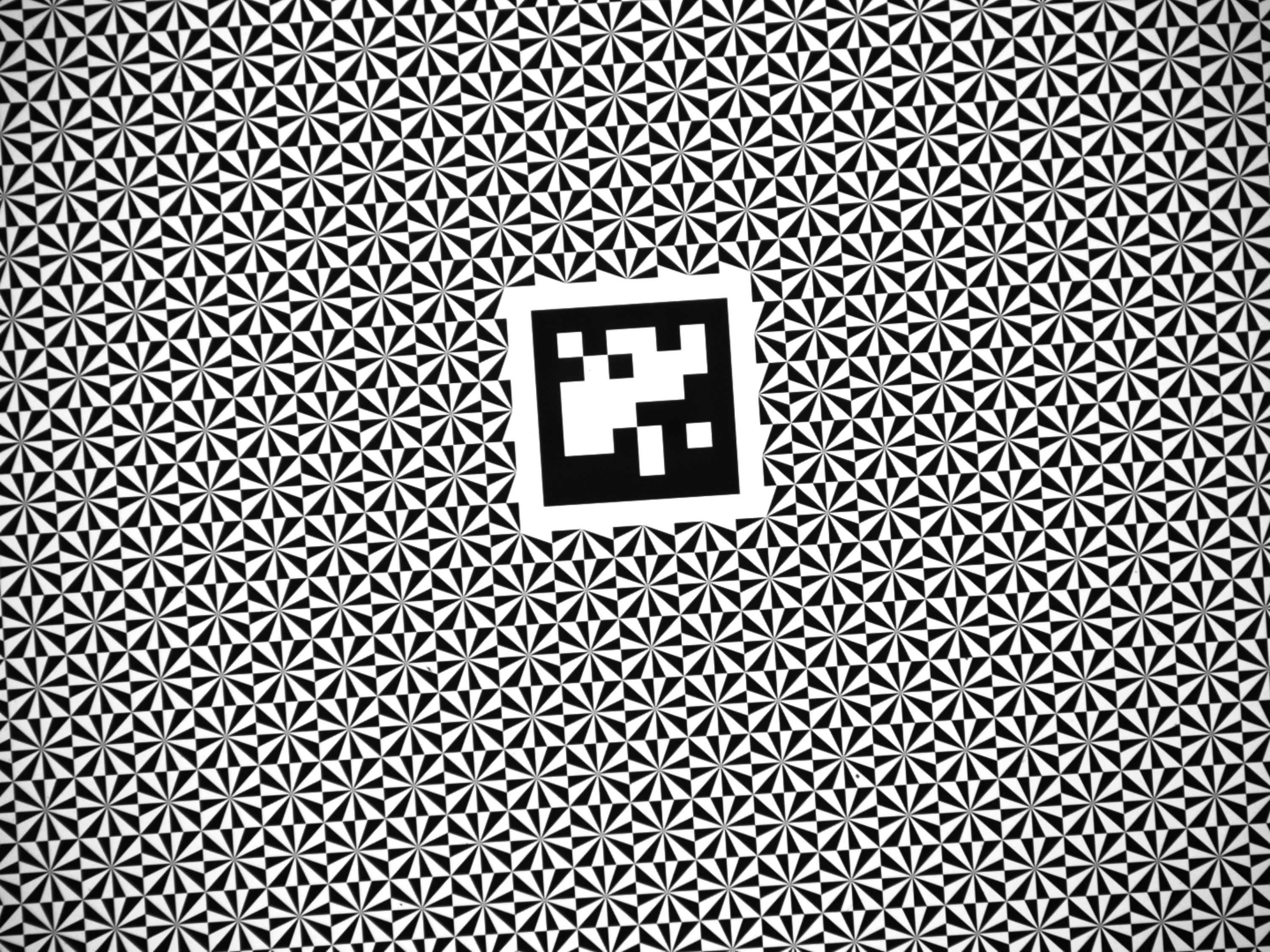}
		\label{fig:C16}
	}
	\subfloat[ Calibration image $C_2$]{
		\includegraphics[width=0.23\linewidth]{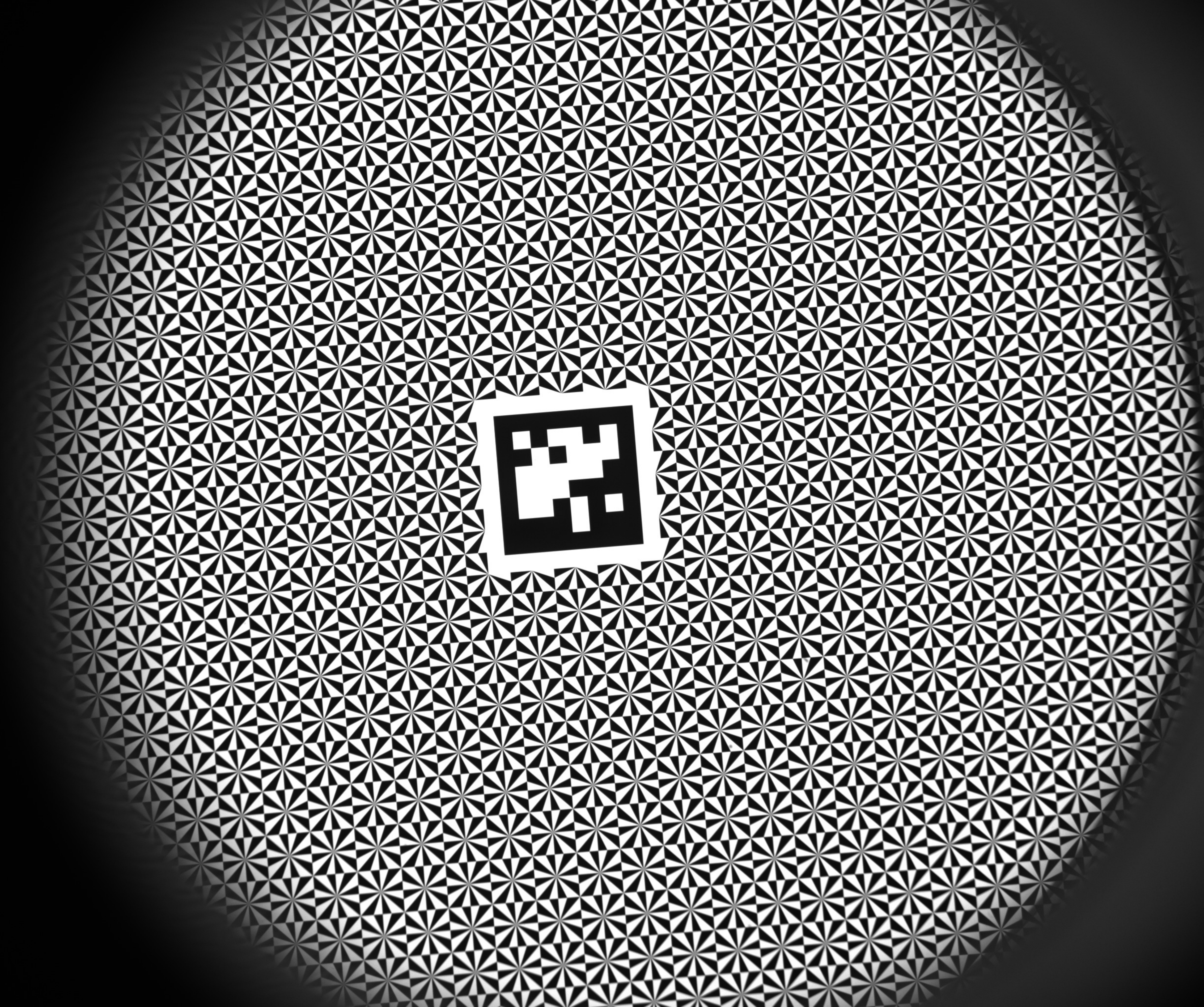}
		\label{fig:C20}
	}
	\subfloat[ Calibration image $C_3$]{
		\includegraphics[width=0.23\linewidth]{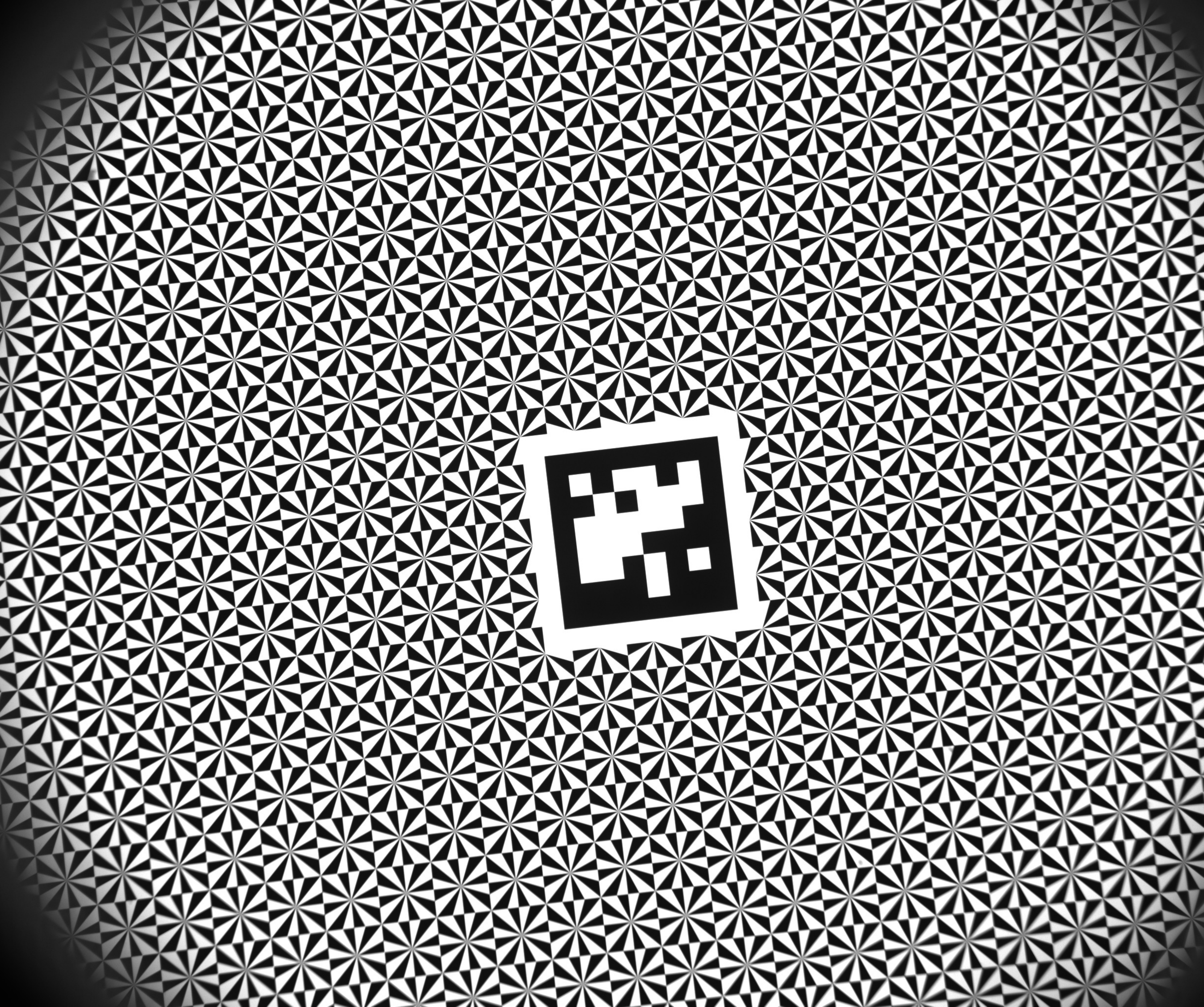}
		\label{fig:C25}
	}
	\caption{Reference image and calibration images of three cameras. The parameters of the reference image are known. The proposed method completes camera calibration using only one collimator image.}
	\label{fig:single}
\end{figure}

The experimental procedure is described in detail below. First, we capture multiple images to calibrate camera $C_2$ and record the corresponding camera parameters. Then, camera $C_2$ re-captures a collimator image as a reference for subsequent calibration, as shown in Fig.~\ref{fig:R20}. Next, all cameras capture one collimator image as the calibration image and simultaneously obtain a set of printed pattern images. The calibration images of $C_1$, $C_2$, and $C_3$ are shown in Fig.~\ref{fig:C16}-\ref{fig:C25}, respectively. The proposed method does not impose any specific requirements on camera's pose when capturing images from the collimator. It only requires that the captured target covers as much of the image area as possible. This flexibility increases the practicality of the method in real-world applications. This experiment aims to validate the effectiveness and accuracy of single-image collimator calibration by comparing its results with printed pattern calibration results. 
\begin{table*}[tbp]
	\centering
	\caption{Comparison results between calibration using a single collimator image and calibration using ten printed pattern images.}
	\resizebox{\linewidth}{!}{
		\begin{tabular}{clcccccccc}
			\toprule
			\multicolumn{2}{c}{}& $f_x$ & $f_y$ & $c_x$ & $c_y$ & $k_1$ & $k_2$ & $\varepsilon_{eval}$ & $\varepsilon_{epip}$\\
			\midrule
			{\multirow{2}{*}{$C_1$}} &\cellcolor{myorg}Collimator &\cellcolor{myorg}4646.3 &\cellcolor{myorg}4648.4 &\cellcolor{myorg}725.4 &\cellcolor{myorg}544.5 &\cellcolor{myorg}0.0204 &\cellcolor{myorg}0.3451 &\cellcolor{myorg}\textbf{0.0627} &\cellcolor{myorg}\textbf{0.0373} \\
			& Printed & 4617.5 & 4615.2 & 740.5& 535.8 & 0.0008 & 0.5815 & 0.0649 & 0.0382 \\
			\midrule
			{\multirow{2}{*}{$C_2$}} & \cellcolor{myorg}Collimator & \cellcolor{myorg}5907.2 & \cellcolor{myorg}5909.1 & \cellcolor{myorg}1252.1 & \cellcolor{myorg}1024.9 & \cellcolor{myorg}0.2491 & \cellcolor{myorg}1.5880 & \cellcolor{myorg}\textbf{0.1535} & \cellcolor{myorg}\textbf{0.0755} \\
			& Printed & 5871.2 & 5879.1 & 1262.8& 1028.1 & 0.2479 & 1.5699 & 0.1542 & 0.0763 \\
			\midrule
			{\multirow{2}{*}{$C_3$}} & \cellcolor{myorg}Collimator & \cellcolor{myorg}7525.5 & \cellcolor{myorg}7529.3 & \cellcolor{myorg}1237.4 & \cellcolor{myorg}1040.2 & \cellcolor{myorg}0.5772 & \cellcolor{myorg}1.4808 & \cellcolor{myorg}\textbf{0.0796} & \cellcolor{myorg}0.1206 \\
			& Printed & 7530.0 & 7529.2 & 1229.4& 1018.1 & 0.5070 & 3.4461 & 0.1004 & \textbf{0.1049} \\
			\bottomrule
			\multicolumn{5}{l}{\textbf{Bold} values indicate the best results.}
		\end{tabular}
	}
	\normalsize
	\label{tab:C1C2C3}
\end{table*}	

For comparative analysis, we use mainstream Bouguet's toolbox \citep{Bouguet2004} to estimate the camera parameters from 10 printed pattern images. The remaining printed pattern images are used for pose estimation to quantify the accuracy of the calibration results. The calibrated camera parameters and the computed pose estimation errors are presented in Table~\ref{tab:C1C2C3}. {The results indicate that the proposed method requires only one collimator image to achieve more competitive results than those obtained using 10 printed pattern images.} Although our method relies on a reference image with known parameters, it is important to note that this reference image can be reused across multiple calibration processes, reducing long-term calibration costs in practical applications. The results from camera $C_1$ demonstrate that the proposed method remains reliable even when calibrating a camera with completely different focal lengths and resolutions. The advantage of our method lies in its ability to perform camera calibration using just one collimator image, which significantly simplifies the calibration process. At the same time, the performance of our method is comparable to traditional methods that utilize multiple printed pattern images.

\subsubsection{Multi-focus Cameras Calibration Results}
\begin{figure}[tbp]
	\centering
	\subfloat[4.02$m$]{
		\includegraphics[width=0.15\linewidth]{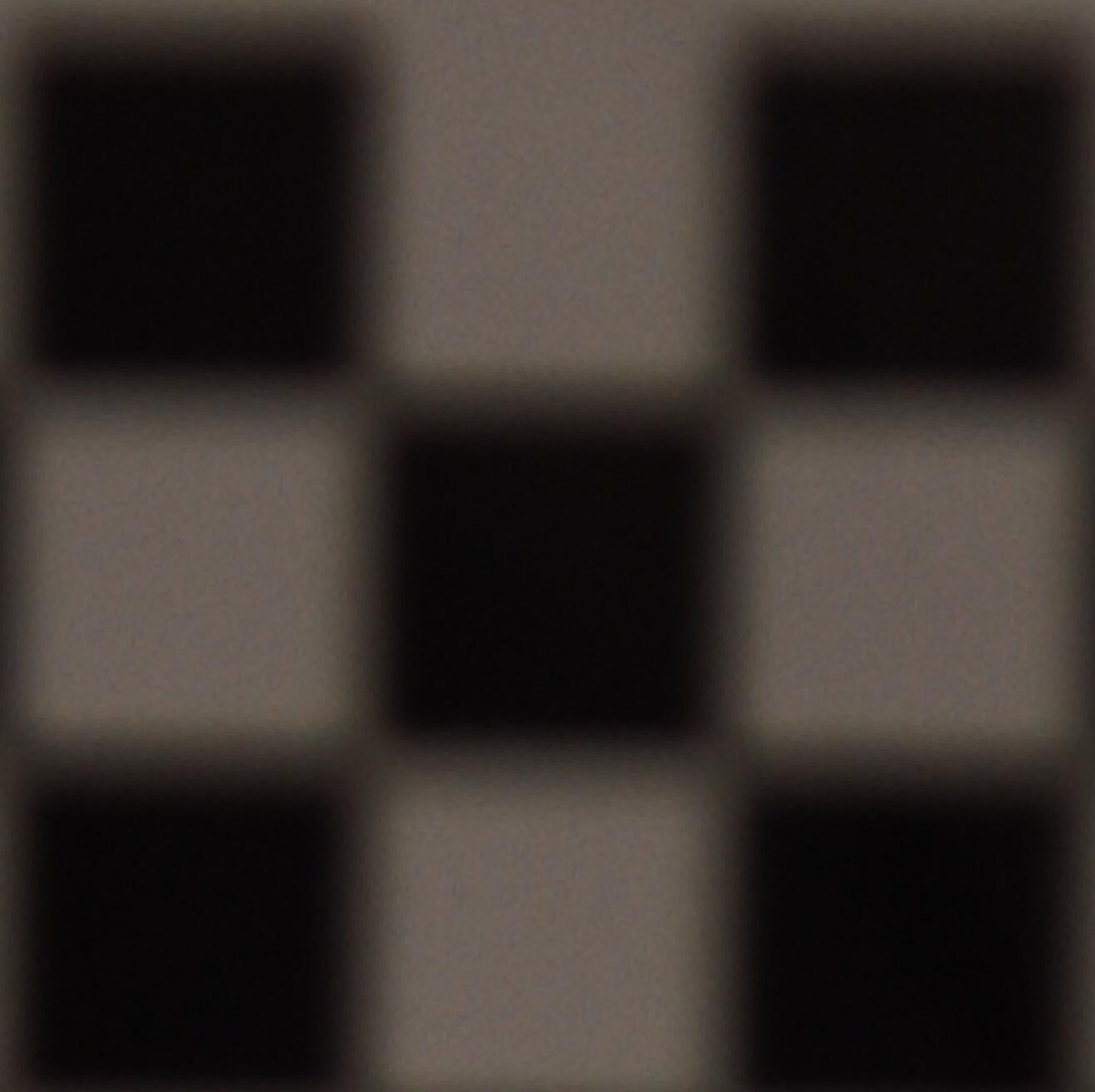} 
		\label{fig:4023}
	}
	\subfloat[6.02$m$]{
		\includegraphics[width=0.15\linewidth]{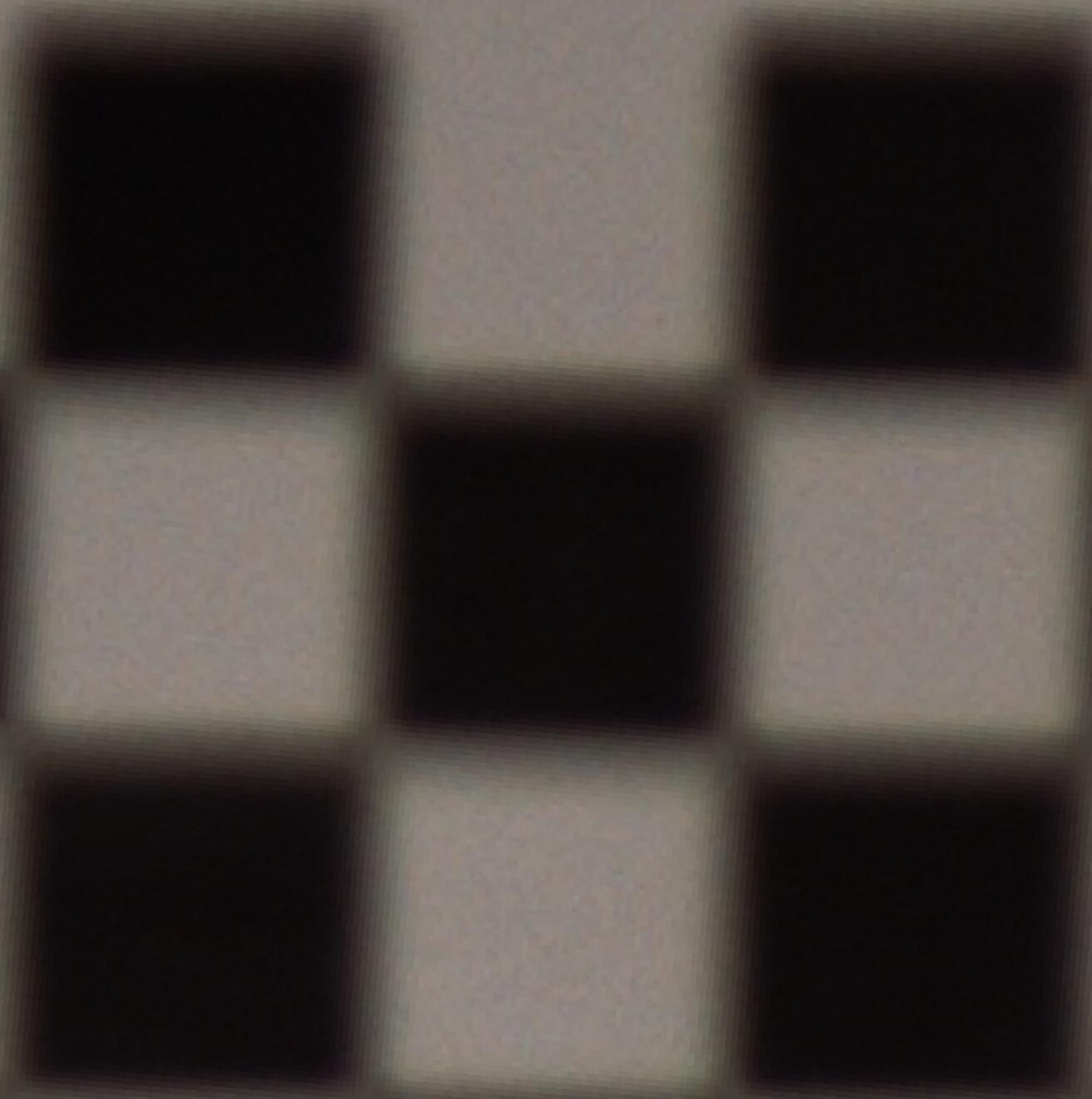}
		\label{fig:6017}
	}
	\subfloat[7.99$m$]{
		\includegraphics[width=0.15\linewidth]{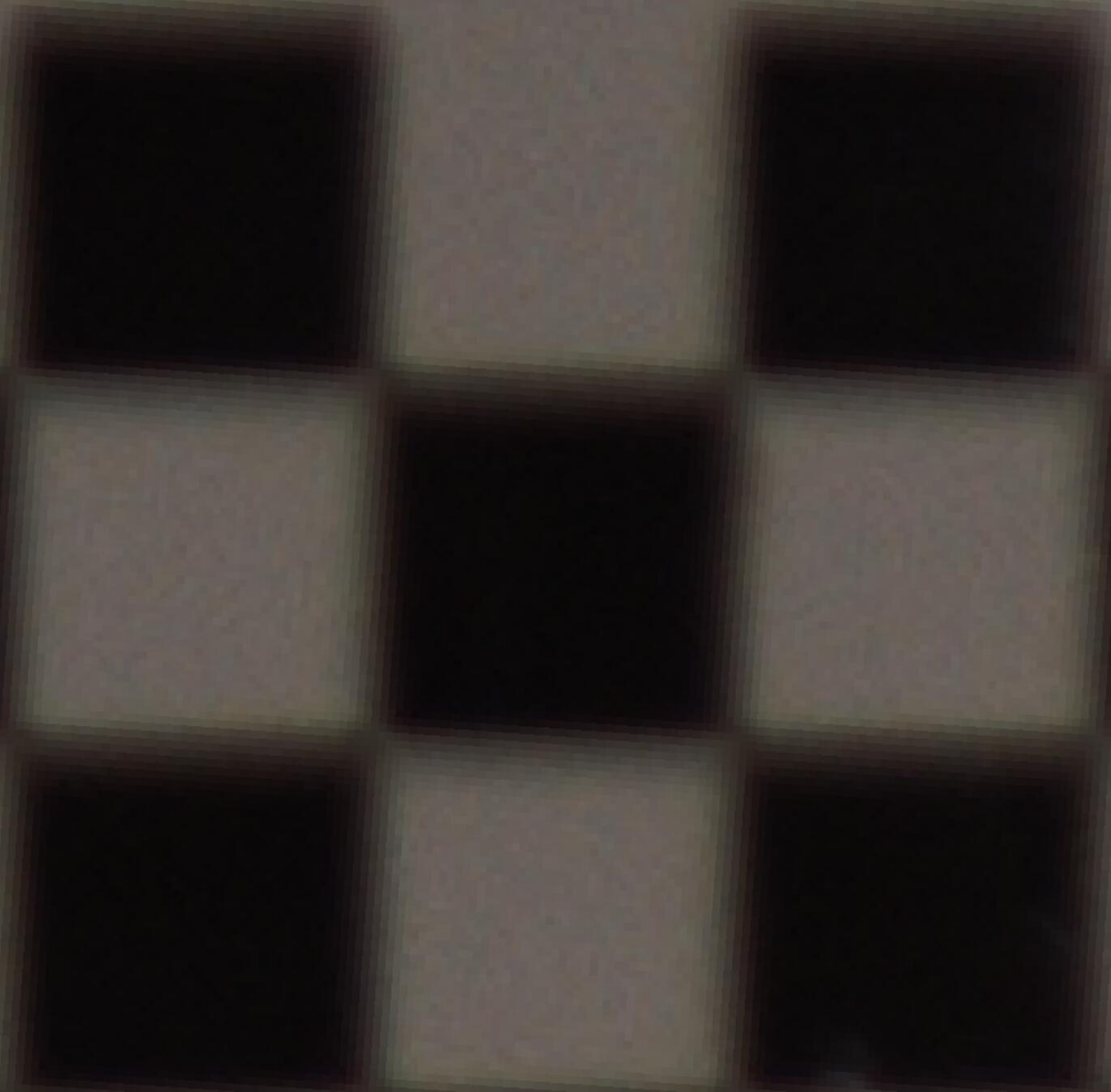}
		\label{fig:7992}
	}	
	\subfloat[10.00$m$]{
		\includegraphics[width=0.15\linewidth]{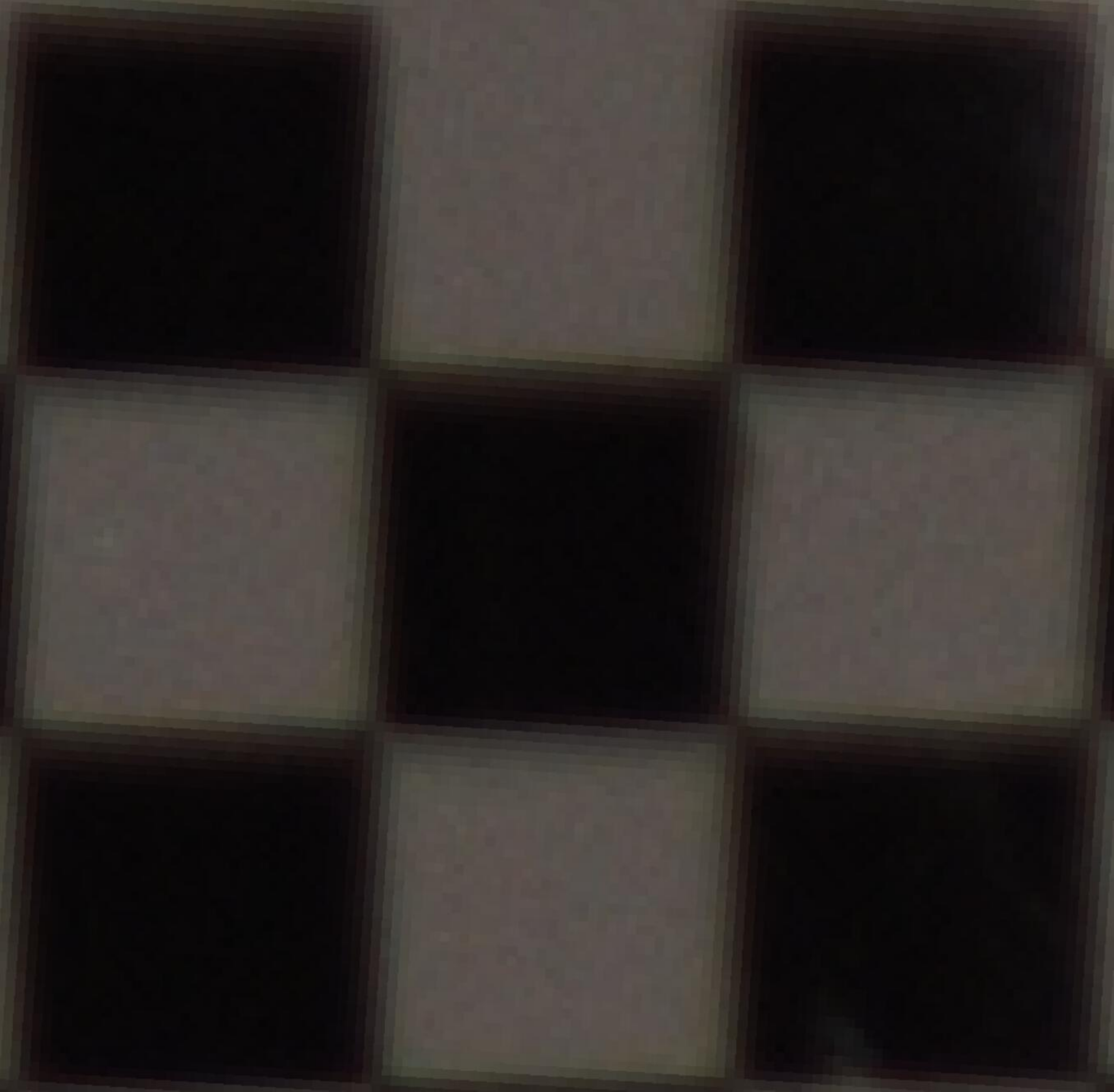}
		\label{fig:9995}
	}
	\subfloat[12.02$m$]{
		\includegraphics[width=0.15\linewidth]{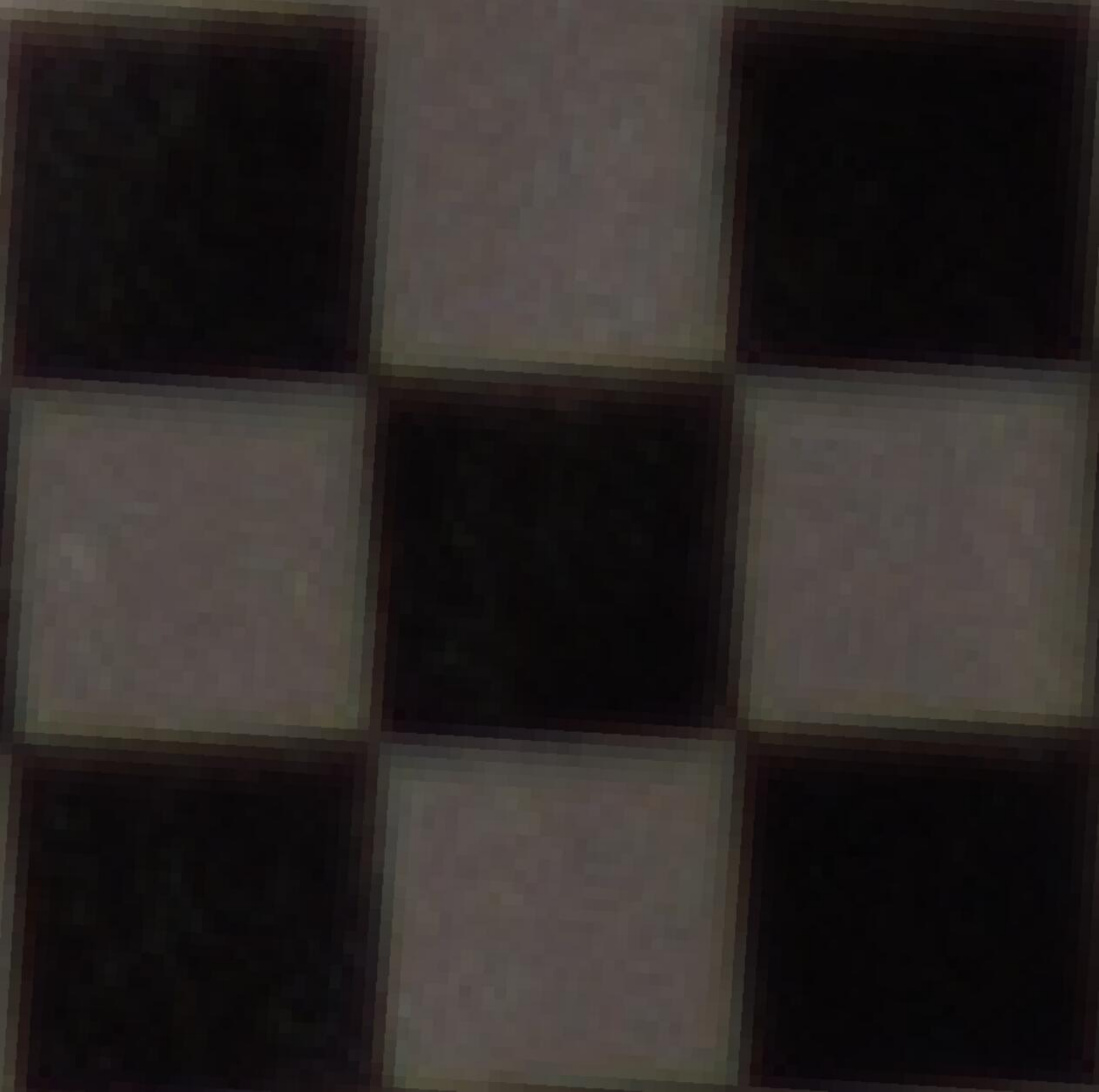}
		\label{fig:12023}
	} 
	\subfloat[15.03$m$]{
		\includegraphics[width=0.15\linewidth]{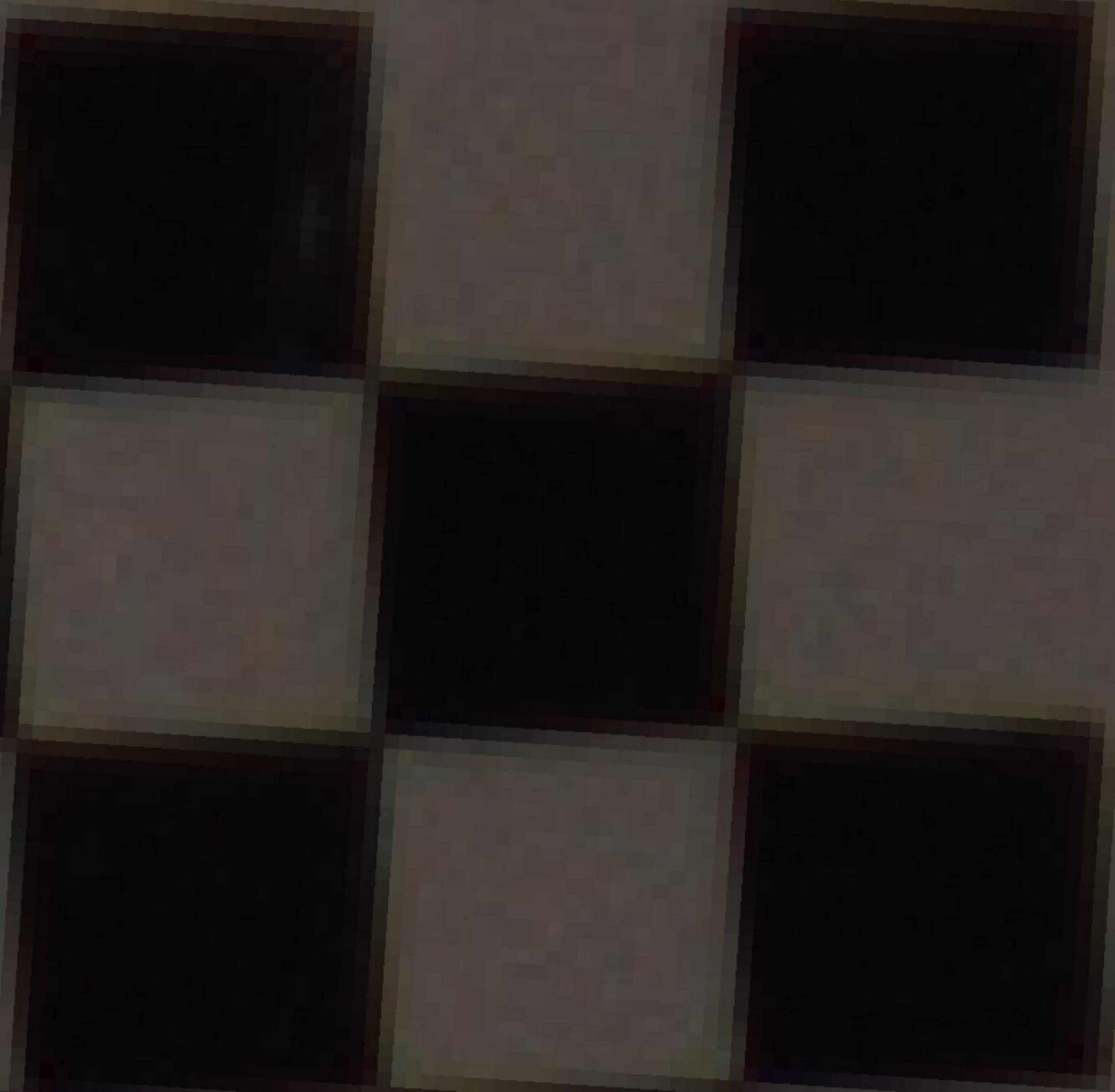}
		\label{fig:15032}
	}
	\setlength{\abovecaptionskip}{0.1cm}
	\caption{The local enlarged images of the checkerboard observed by the $35mm$ camera at different distances. The corners can be clearly distinguished at a minimum distance of $12m$. At distances less than $12m$, the corners appear blurred.}
	\label{fig:35mm}
\end{figure}

In this experiment, we verify the effectiveness of our collimator system for calibrating cameras with different focal lengths. We selected four lenses with focal lengths of 12$mm$, 16$mm$, 25$mm$, and 35$mm$ for calibration. The resolution of the camera is $2448 \times 2048$ pixels. {All images are obtained from a same collimator system.} Before the formal calibration experiments, we test the required placement distance for calibrating the $35mm$ lens using the traditional checkerboard calibration method. The camera is configured as a long working distance model, which means we used the maximum aperture and the farthest focusing distance. This setup ensures the camera can clearly observe distant buildings. The size of the checkerboard used for testing is $740mm\times518mm$ with $10\times7$ squares. Figure~\ref{fig:35mm} shows the local enlarged images of chessboard corners observed by a $35mm$ camera at different distances. Intuitively, the camera can only clearly resolve the corner points on the chessboard at observation distances of 12 meters or more. Quantitatively, we collected multiple images at different distances for calibration and reported the re-projection error. As shown in Table~\ref{tab:distance}, the recommended calibration distance for the $35mm$ camera is 12 meters. Such a calibration distance is inconvenient in practical operations. In contrast, our method requires only a $0.3m$-collimator to achieve accurate calibration, offering significant advantages.

\begin{table}[tbp]
	\centering
	\caption{Re-projection error for the calibration of the $35mm$ focus length camera at different distances.}
	\begin{tabular}{p{4cm} >{\centering\arraybackslash}p{1.0cm} >{\centering\arraybackslash}p{1.0cm} >{\centering\arraybackslash}p{1.0cm} >{\centering\arraybackslash}p{1.0cm} >{\centering\arraybackslash}p{1.0cm} >{\centering\arraybackslash}p{1.0cm}}
		\toprule
		Distance ($m$) & 4.02 & 6.02 & 7.99 & 10.00 & 12.02 & 15.03 \\
		\midrule
		\rowcolor{myorg}
		re-projection error (pixels)& {0.94} & {0.33} & {0.21} & {0.17} & {0.13} & {0.11}\\
		\bottomrule	
	\end{tabular}
	\label{tab:distance}
\end{table} 
\begin{figure}[tbp]
	\centering
	\subfloat[12$mm$]{
		\includegraphics[width=0.23\linewidth]{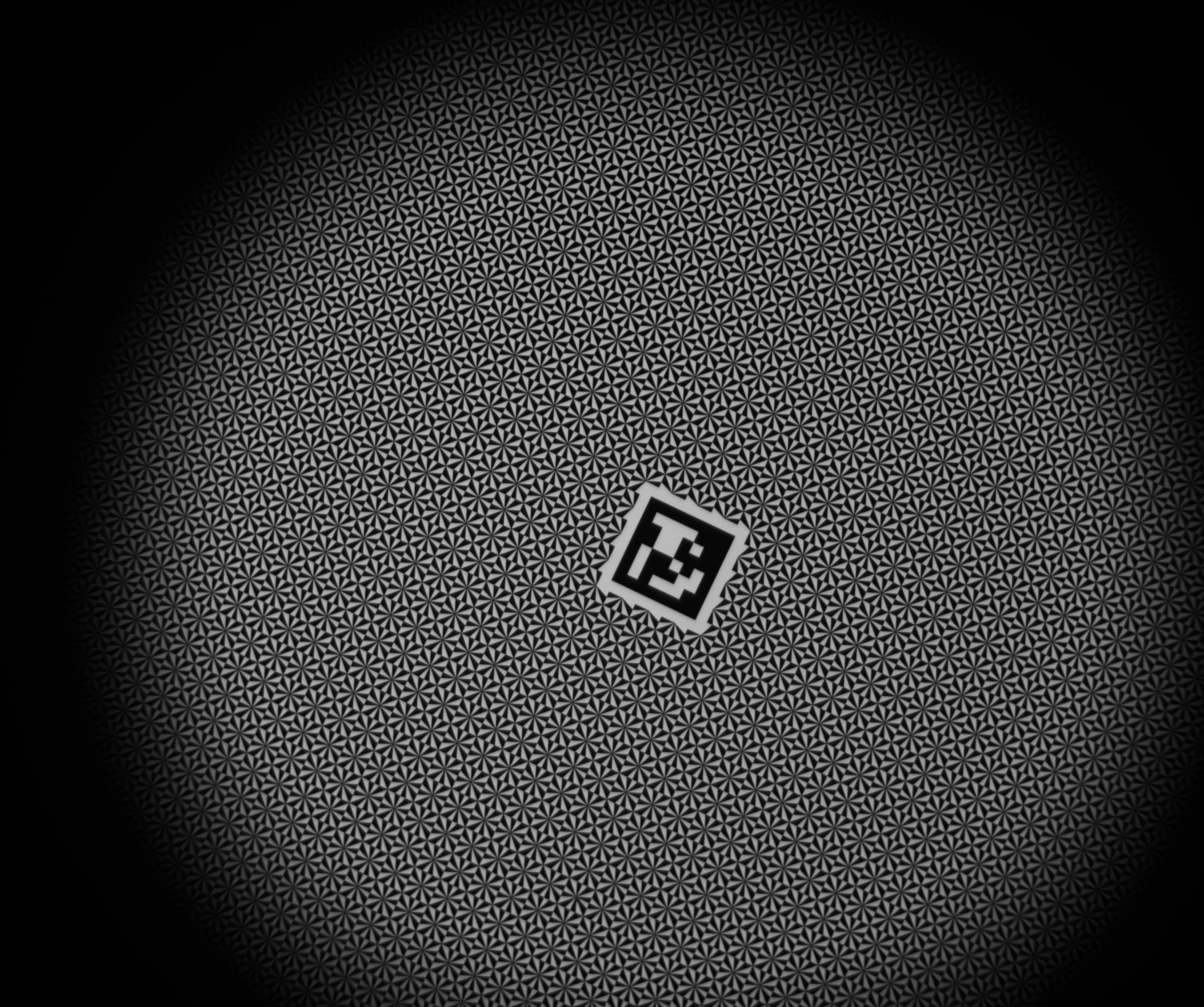} 
	}
	\subfloat[16$mm$]{
		\includegraphics[width=0.23\linewidth]{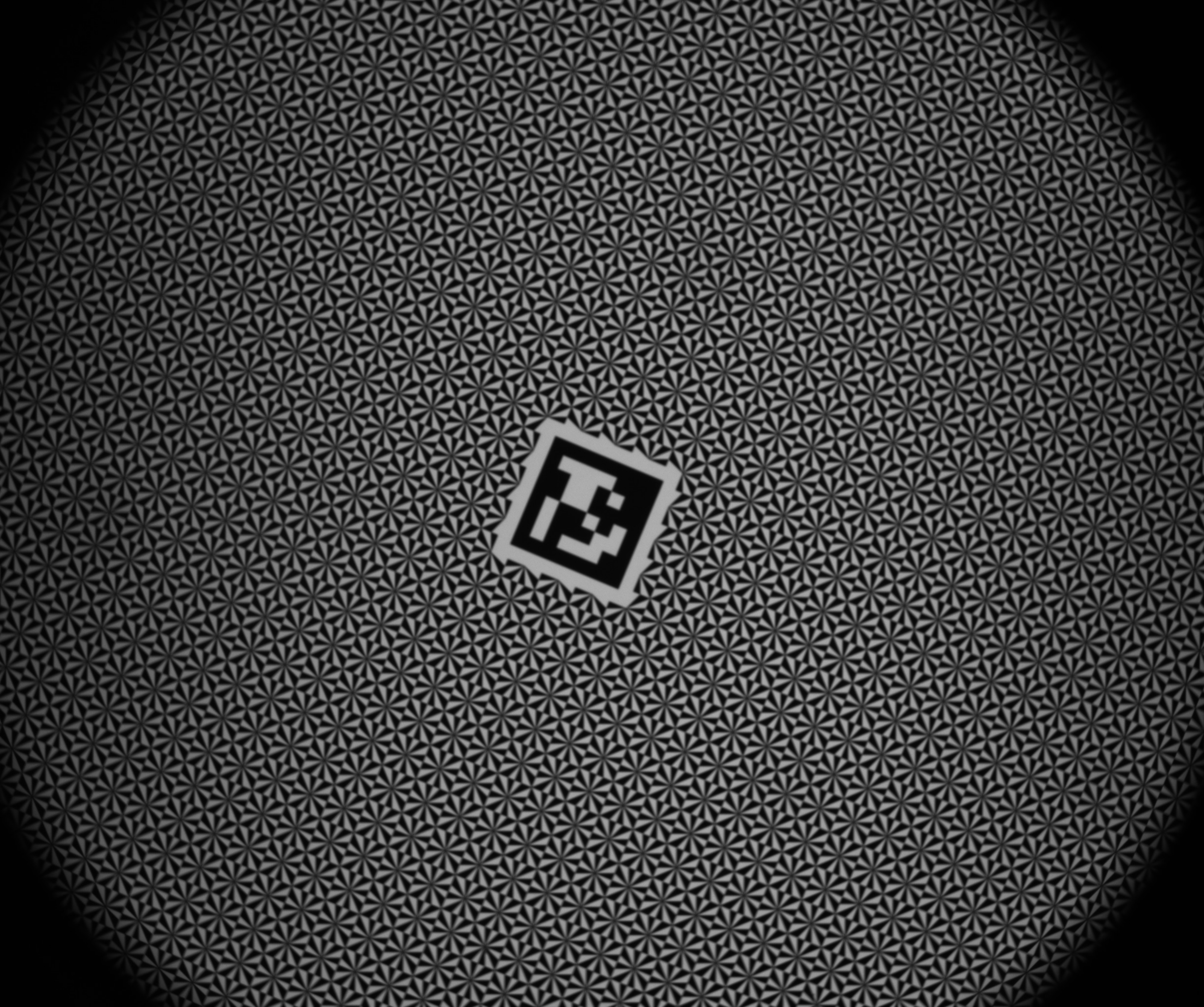}
	}
	\subfloat[25$mm$]{
		\includegraphics[width=0.23\linewidth]{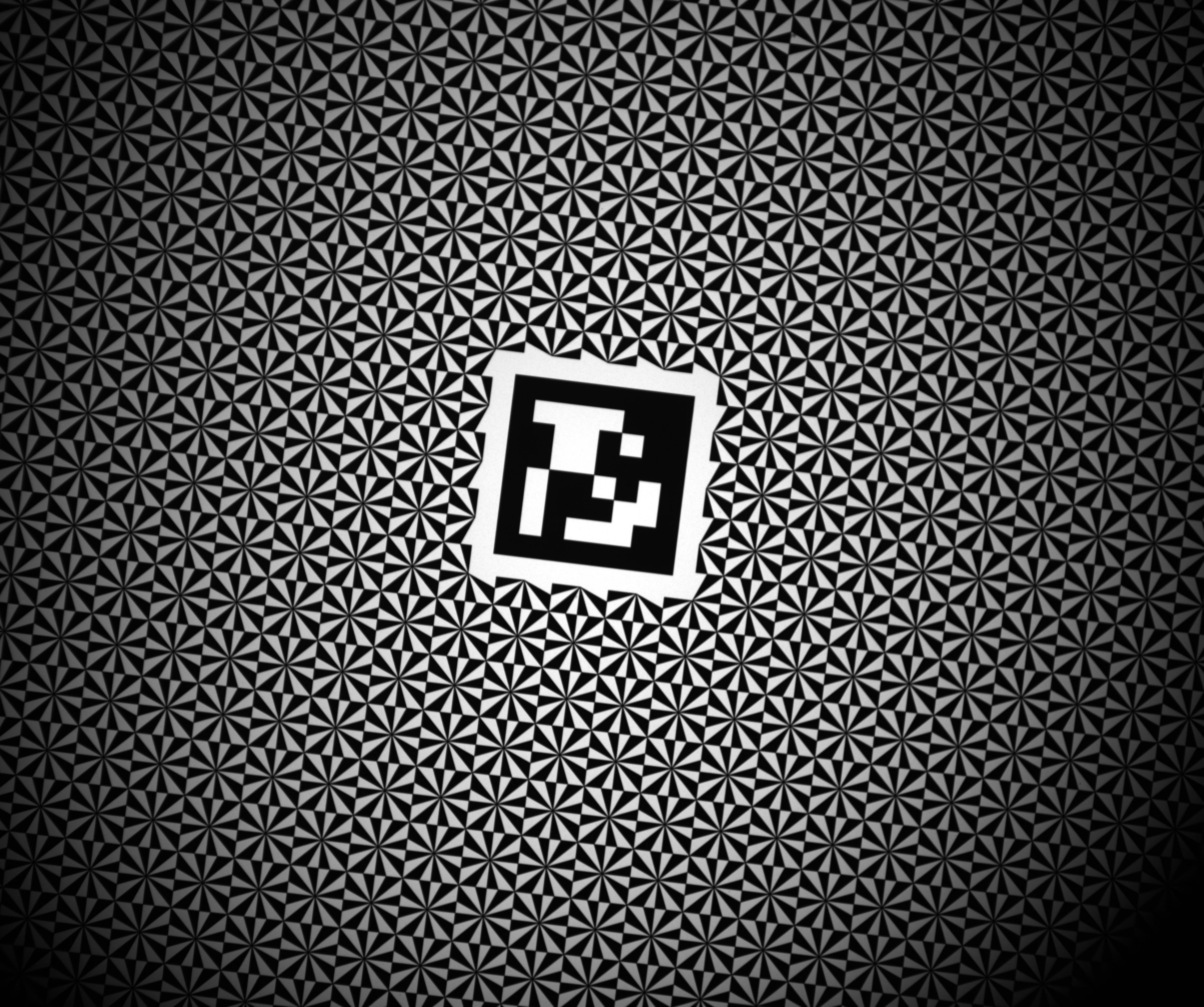}
	}	
	\subfloat[35$mm$]{
		\includegraphics[width=0.23\linewidth]{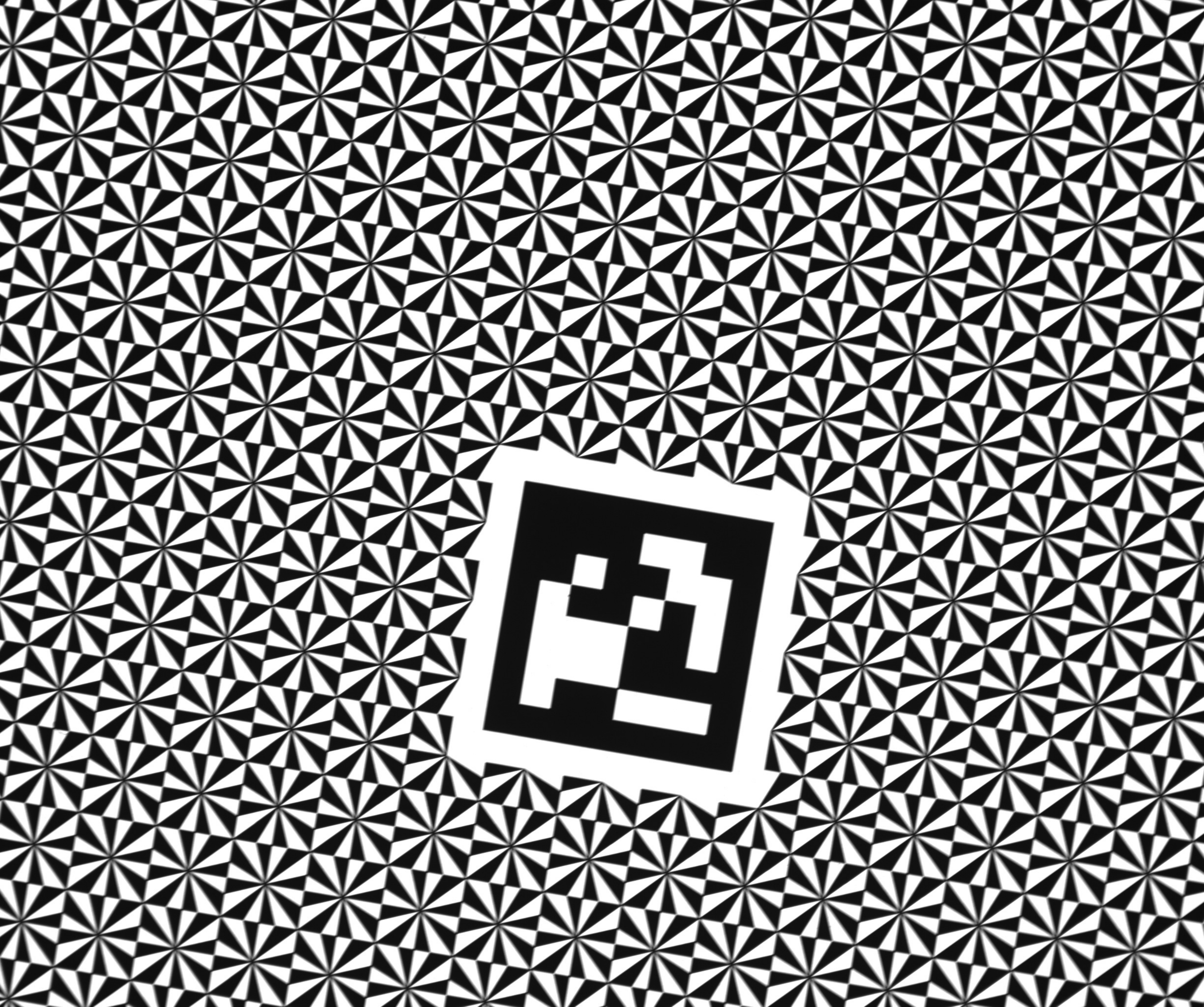}
	}
	\caption{{Sample images of the collimator captured by cameras with different focal lengths. Cameras with different focal lengths are capable of clearly observing the calibration target from the collimator system}}
	\label{fig:SampleImages}
\end{figure}
\begin{figure}[tbp]
	\centering
	\includegraphics[width=0.244\linewidth]{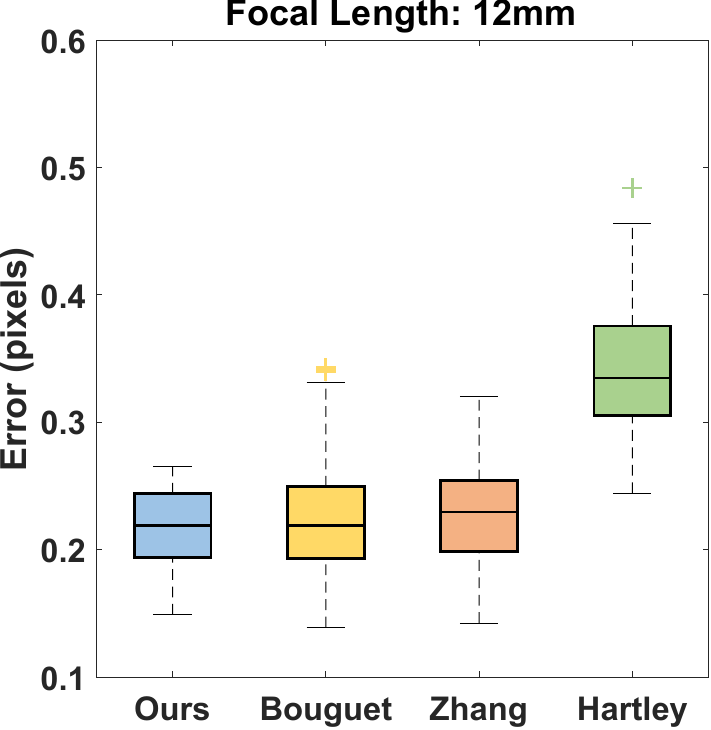} 
	\includegraphics[width=0.244\linewidth]{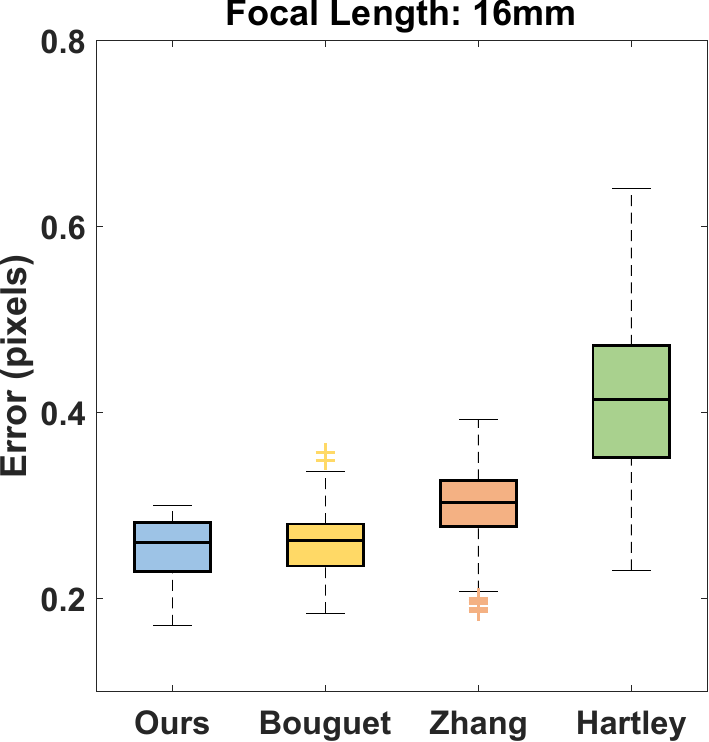}
	\includegraphics[width=0.244\linewidth]{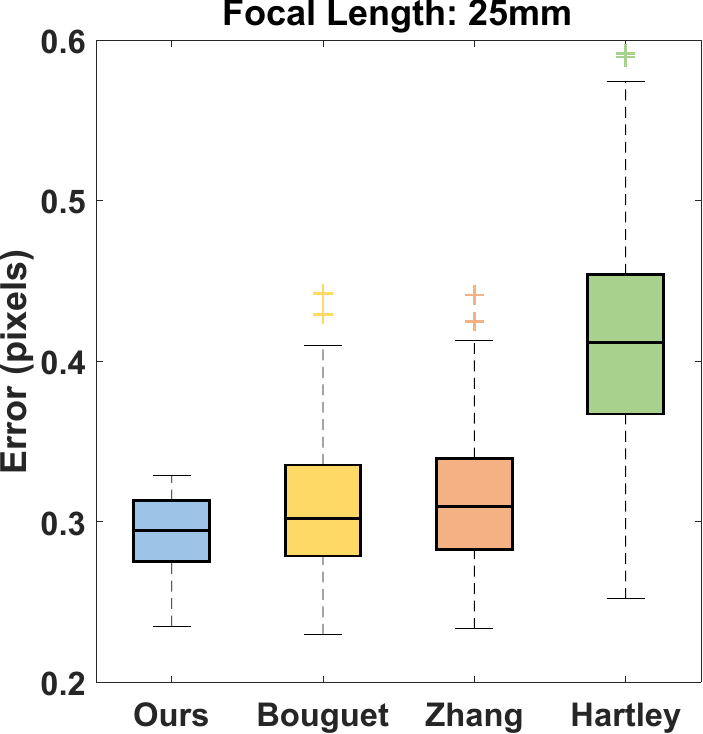}
	\includegraphics[width=0.247\linewidth]{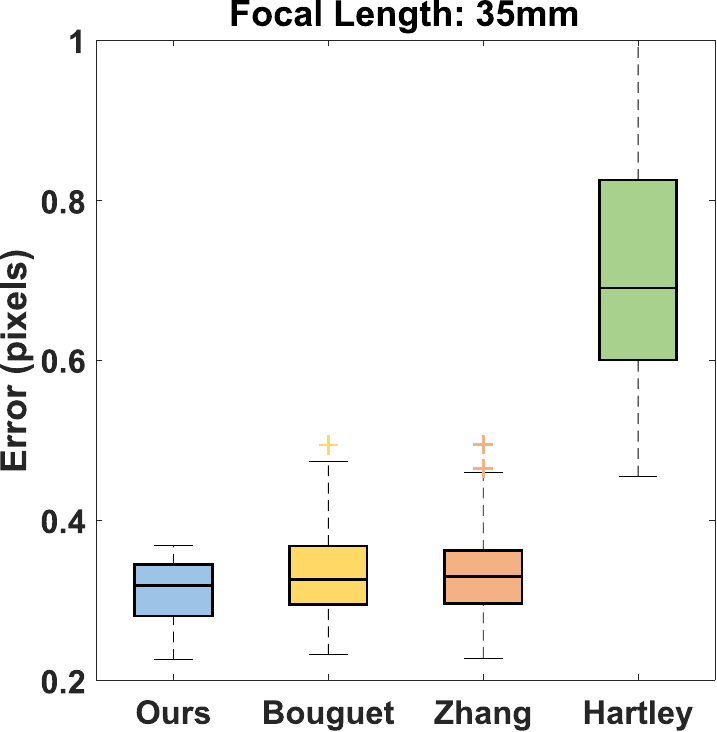}
	\caption{Evaluation error distribution of different focal length cameras. Our designed collimator system is compatible with lenses of varying focal lengths, and in conjunction with the proposed calibration algorithm, achieves optimal accuracy.}
	\label{fig:F12-35}
\end{figure}	
\begin{table*}[tbp]
	\centering
	\caption{Comparison of calibration results with different focal length lense (pixels).}
	\begin{tabular}{>{\centering\arraybackslash}p{2.2cm} >{\centering\arraybackslash}p{2.2cm} >{\centering\arraybackslash}p{2.2cm} >{\centering\arraybackslash}p{2.2cm} >{\centering\arraybackslash}p{2.2cm}}
		\toprule
		{Focal length}&{Zhang}&{Bouguet}&{Hartley}&{Ours}\\
		\midrule
		\rowcolor{myorg}
		{12$mm$}& {0.2273}& {0.2225}& {0.3404}& \textbf{0.2174}	\\
		{16$mm$}&{0.2994}&{0.2591}&{0.4182}&\textbf{0.2536}	\\
		\rowcolor{myorg}
		{25$mm$}& {0.3133}& {0.3078}& {0.4195}& \textbf{0.2923}	\\
		{35$mm$}&{0.3289}&{0.3348}&{0.7272}&\textbf{0.3118}	\\
		\bottomrule
		\multicolumn{5}{l}{\textbf{Bold} values indicate the best results.}
	\end{tabular}
	\label{tab:Lenses}
\end{table*}

We capture images using four cameras with different focal lengths from the collimator system. {The sample images are shown in Fig.~\ref{fig:SampleImages}.} Cameras with different focal lengths can observe the calibration target through our collimator system in a limited space. As the focal length increases, the more pixels a single square occupies. Similar to the evaluation approach in the previous experiment, we segregate all collimator images into calibration and evaluation sets. For each lens, {we use 20 images for calibration and 200 images for evaluation.} Figure~\ref{fig:F12-35} shows the distribution of 200 evaluation errors. The results indicate that the proposed algorithm performs best, with a more concentrated error distribution and lower overall error levels. Quantitative results are listed in Table~\ref{tab:Lenses}. The proposed algorithm demonstrates the lowest re-projection error for lenses with varying focal lengths among the evaluation images. The re-projection error tends to increase with the increase of focal length. The main reason is that the longer the lens's focal length, the fewer points observed. More importantly, this experiment confirms the effectiveness of our collimator system and the proposed algorithm. Compared to the printed pattern, our collimator system reduces space requirements, making it more flexible and efficient for practical applications.

\section{Conclusion}
This paper presents a novel camera calibration method using a designed collimator system. Based on the optical geometry of the collimator system, {we prove that the relative motion between the target and the camera conforms to the spherical motion, and provide experimental validation using real collimator images.} This significant finding allows us to simplify the original general 6 DoF motion to a 3 DoF pure rotational motion, thereby providing new constraints for camera calibration. We then develop several camera calibration algorithms using collimator images. For multiple images, we propose a closed-form linear solution method. For two images, we introduce a minimal solution approach. Furthermore, we innovatively propose an algorithm that calibrates using only a single collimator image, fully leveraging the angle invariance provided by the collimator. Synthetic and real data experiments demonstrate that the calibration accuracy of the proposed method is superior to the baseline methods. Additionally, our method offers greater flexibility due to the stable calibration environment created by the collimator and its reduced spatial requirements.

\section*{Acknowledgments}
This research has been supported by the Hunan Provincial Natural Science Foundation for Excellent Young Scholars under Grant 2023JJ20045, the National Natural Science Foundation of China under Grant 12372189, {the Science and Technology Innovation Program of Hunan Province under Grant 2022RC1196}. We thank Zijie Wang for useful discussion on the writing of the paper.

\section*{Data Availability Statements}
The code and example data used in this study is publicly available at \href{https://github.com/LiangSK98/CollimatorCalibration}{https://github.com/LiangSK98/CollimatorCalibration}. The complete data that support the findings of this study are available from the corresponding author on reasonable request.

\bibliography{sn-bibliography}
\end{document}